\documentclass[letterpaper]{article} 
\usepackage{arxiv}  
\usepackage{times}  
\usepackage{helvet}  
\usepackage{courier}  
\usepackage[hyphens]{url}  
\usepackage{graphicx} 
\usepackage{natbib}  
\usepackage{caption} 
\usepackage{algorithm}
\usepackage{algorithmic}
\usepackage[T1]{fontenc}

\usepackage{newfloat}
\usepackage{listings}
\DeclareCaptionStyle{ruled}{labelfont=normalfont,labelsep=colon,strut=off} 
\floatstyle{ruled}
\newfloat{listing}{tb}{lst}{}
\floatname{listing}{Listing}
\usepackage{amsmath}
\usepackage{amsfonts}
\usepackage{xcolor}
\usepackage{tabularx}
\usepackage{booktabs}
\usepackage{multirow}

\DeclareMathOperator*{\mean}{mean}

\title{Fragment-Wise Interpretability in Graph Neural Networks \\ via Molecule Decomposition and Contribution Analysis}
\author{
    Sebastian Musia\l{}\textsuperscript{\rm 1}, Bartosz Zieli\'{n}ski\textsuperscript{\rm 1}, Tomasz Danel\textsuperscript{\rm 1,2,$\ast$}
}
\affiliations{
    \textsuperscript{\rm 1} Faculty of Mathematics and Computer Science, Jagiellonian University, Krak\'{o}w 30-348, Poland\\
    \textsuperscript{\rm 2} Faculty of Chemistry, Jagiellonian University, Krak\'{o}w 30-387, Poland\\
    \textsuperscript{$\ast$} tomasz.danel@uj.edu.pl
}

\begin{document}

\maketitle

\begin{abstract}
Graph neural networks have demonstrated remarkable success in predicting molecular properties by leveraging the rich structural information encoded in molecular graphs. However, their black-box nature reduces interpretability, which limits trust in their predictions for important applications such as drug discovery and materials design. Furthermore, existing explanation techniques often fail to reliably quantify the contribution of individual atoms or substructures due to the entangled message-passing dynamics. We introduce SEAL (Substructure Explanation via Attribution Learning), a new interpretable graph neural network that attributes model predictions to meaningful molecular subgraphs. SEAL decomposes input graphs into chemically relevant fragments and estimates their causal influence on the output. The strong alignment between fragment contributions and model predictions is achieved by explicitly reducing inter-fragment message passing in our proposed model architecture. Extensive evaluations on synthetic benchmarks and real-world molecular datasets demonstrate that SEAL outperforms other explainability methods in both quantitative attribution metrics and human-aligned interpretability. A user study further confirms that SEAL provides more intuitive and trustworthy explanations to domain experts. By bridging the gap between predictive performance and interpretability, SEAL offers a promising direction for more transparent and actionable molecular modeling.
\end{abstract}

\section{Introduction}

Predicting molecular properties is a cornerstone task in computational chemistry, with applications spanning drug discovery, materials science, and chemical synthesis~\citep{elton2019deep}. Graph Neural Networks (GNNs) have emerged as state-of-the-art models for this task, thanks to their ability to naturally represent molecular structures as graphs of atoms and bonds~\citep{wieder2020compact}. However, despite their impressive predictive performance, GNNs suffer from a critical limitation: their decision-making processes are largely opaque. Moreover, explaining the predictions of GNNs through external methods is challenging, primarily due to the message aggregation mechanism, which is invariant to node permutation. This results in a disorganized exchange of information among nodes, making explanations increasingly difficult with each graph layer. This often leads to the well-known issue of oversmoothing, where the atomic representation becomes independent of its localization within the graph~\citep{zhang2023comprehensive}. This challenge creates a significant problem when applying GNNs in domains where interpretability is crucial for advancing scientific discovery.

We hypothesize that GNNs operating on molecular graphs decomposed into fragments are more interpretable while preserving sufficient expressivity. Many chemical prediction tasks inherently depend more on the presence and identity of specific fragments than on complex interactions between them. Functional groups and other substructures often dominate the behavior of molecules with respect to properties such as solubility, toxicity, and reactivity~\citep{von2006toxicity, tanaka2010construction, tuttle2023predicting}. For instance, in drug discovery, the binding affinity of a molecule to a target protein is frequently the cumulative result of individual fragment-level interactions, such as hydrogen bonds, hydrophobic interactions, or $\pi$-$\pi$ stacking from aromatic rings, each contributing additively to the overall binding energy~\citep{murcko1995computational}. Similarly, properties like lipophilicity or metabolic stability are often strongly associated with the presence of particular chemical motifs, irrespective of subtle global dependencies. While GNNs are capable of modeling complex dependencies between atoms, in many real-world tasks, fragment-level contributions offer sufficient explanatory power for guiding chemical reasoning and decision-making. By focusing on these contributions, the fragment-based model can produce explanations that are chemically intuitive and well-suited for application in molecular design.

\begin{figure*}
    \centering
    \includegraphics[width=0.9\linewidth]{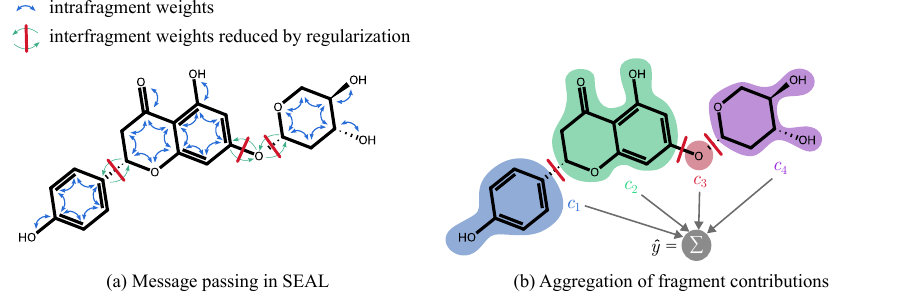}
    \caption{Overview of SEAL comprising a message passing stage \textbf{(a)}, in which a molecular graph is divided into fragments and different weights are applied within and between fragments, and a fragment aggregation stage \textbf{(b)}, in which fragment representations are computed and used to produce fragment contributions; the model prediction is a sum of all fragment contributions.}
    \label{fig:seal}
\end{figure*}

Recent efforts have sought to interpret GNN predictions by attributing importance to individual nodes, edges, or subgraphs~\citep{ponzoni2023explainable}. While these methods provide some insight, they often fail to produce explanations that align with the way chemists conceptualize molecular behavior. Chemists typically reason about molecules not as collections of individual atoms or bonds but in terms of fragments, such as functional groups, aromatic rings, or side chains, that are strongly associated with specific properties or behaviors~\citep{talanquer2022complexity}. Existing explainability methods fail to capture this crucial level of abstraction, limiting their practical relevance for molecular modeling.

In this paper, we introduce \textbf{SEAL} (\textbf{S}ubstructure \textbf{E}xplanation via \textbf{A}ttribution \textbf{L}earning), a novel interpretable GNN that generates fragment-wise explanations of model predictions for molecular property prediction (Figure~\ref{fig:seal}). SEAL decomposes molecular graphs into chemically meaningful fragments and quantifies the individual contribution of each fragment to the model output. By aligning interpretability with chemical intuition, our approach enables domain experts to leverage machine learning predictions for better hypothesis generation and experimental design. Our contributions are threefold:
\begin{enumerate}
    \item We propose a novel approach for fragment-based interpretability in GNNs for molecular property prediction.
    \item We define a new graph convolutional layer and regularization terms that facilitate information retention in the fragments, resulting in more intuitive explanations.
    \item We conduct extensive experiments that demonstrate that our model produces more coherent and actionable explanations compared to existing methods while maintaining competitive predictive performance.
\end{enumerate}

The code is available at \url{https://github.com/gmum/SEAL}.

\section{Related Work}

\paragraph{Graph neural networks.} GNNs have become a standard method for analyzing molecular data, often using either a message-passing mechanism~\citep{gilmer2020message} or a transformer-based architecture~\citep{rong2020self,maziarka2024relative}. Some of these networks work on fragment graphs where atom groups serve as nodes instead of individual atoms. For example, \citet{cao2024group} proposed a GNN that uses fragment-level message passing for better explainability but still relies on external explainers to determine fragment contributions. \citet{wangfragformer} recently introduced FragFormer, a transformer that operates on fragments and employs a variant of the CAM method~\citep{zhou2016learning} to explain its predictions. In both models, fragments can contain significant signals coming from other parts of the molecule, potentially reducing local interpretability.

\paragraph{Graph-based explainers.} Many explainable AI (XAI) techniques have been proposed to elucidate the predictions of GNNs. Some identify important subgraphs by perturbing the input graph~\citep{ying2019gnnexplainer,vu2020pgm,yuan2021explainability}, while other methods analyze the message-passing mechanism in each layer~\citep{fengdegree,gui2023flowx}. Often, explaining GNNs is difficult due to the large number of subgraphs and the complex message-passing process. Therefore, \citet{henderson2021improving} proposed regularization techniques that disentangle node representations, aiding in generating better explanations. Another approach involves presenting explanations at the fragment level. For instance, \citet{wu2023chemistry} employed BRICS~\citep{degen2008art} to break down molecules into chemically plausible segments and elucidate predictions by masking entire molecular fragments. Inherently interpretable methods are also being developed, including prototype-based graph neural networks~\citep{zhang2022protgnn,rymarczyk2023progrest} and attention-based models~\citep{xiong2019pushing,lee2023multi}. However, prototypical parts and attention maps on graphs can still be difficult for humans to interpret because of the multitude of explanation patterns that need to be analyzed.

\section{Methods}

\subsection{Fragment Contributions}

The interpretability of our model is achieved by redesigning the prediction head in graph-based models. Typically, a readout function in GNNs is used to create a graph-level representation, and then an MLP is applied to make predictions. However, the graph readout aggregates information from all atoms in the graph, hindering the ability to attribute predictions to specific atoms or functional groups.

Our model first aggregates information within graph fragments, which can be predefined or learned through a differentiable graph pooling layer. In our method, we use sum pooling followed by a LayerNorm~\citep{ba2016layer} to create the fragment representation from the fragment atom representations. Then, the contribution for each fragment is computed with an MLP, and the final prediction is the sum of all fragment contributions. 

Let us define a molecular graph $\mathcal{G}=(\mathcal{V}, \mathcal{E}, X)$, where $\mathcal{V}=\{v_i\}_{i=1}^N$ is a set of nodes corresponding to atoms, $\mathcal{E}\subseteq \mathcal{V}\times\mathcal{V}$ is a set of edges corresponding to chemical bonds, $X\in \mathbb{R}^{N\times D}$ is an atom feature matrix, and $D$ is the number of node features. After passing this graph through a sequence of GNN layers, a matrix of atom representations $H\in\mathbb{R}^{N\times M}$ is obtained. Each atom is assigned to exactly one of the $K$ fragments $\mathcal{F}_1,\dots,\mathcal{F}_K$. Then, the model output is computed as follows:
\begin{gather}
    \bar{\mathbf{h}}_i = \sum_{v_j\in \mathcal{F}_i} \mathbf{h}_j, \quad
    c_i = \operatorname{MLP}\left(\bar{\mathbf{h}}_i\right), \\
    \hat{y} = \sum_{i=1}^K c_i + b,
\end{gather}
where $\bar{\mathbf{h}_i}$ is the representation of $i$-th fragment, $c_i$ is the contribution of this fragment, $b$ is a trainable bias term, and $\hat{y}$ is the model prediction. The fragment contributions can be interpreted as the importance of these fragments. The bias is crucial to calibrate the model predictions, particularly when the label distribution is not centered at zero. In that case, the lack of bias would cause an equal distribution of contributions across all fragments, diminishing the interpretability of the model.

In all of our experiments, we use a variant of the BRICS algorithm~\citep{degen2008art} similar to that proposed by \citet{zhang2021motif}. Briefly, we isolate side chains from rings, even if the side chain contains only one atom, and non-ring atoms with four or more neighbors are also treated as separate fragments. Additionally, we cut all non-ring bonds that connect two rings and all halogen groups.

\subsection{SEAL-GCN}

The aggregation of messages from neighboring nodes in GNNs is invariant to node permutations. While this mechanism is effective in extracting meaningful information from molecular graphs needed for making correct predictions, the information from each node is easily diffused in the graph, hurting the model's ability to localize crucial atoms and leading to the known problem of oversmoothing.

To mitigate the problem of leaking unnecessary information to neighboring nodes, we propose SEAL-GCN, a new graph neural network variant that operates on pre-fragmented graphs, controlling the information exchanged between fragments. In our implementation, graph layers have separate weights for interfragment and intrafragment edges. This enables the network to extract relevant information within molecular fragments and block information leaks to neighboring fragments. The SEAL-GCN layer is defined as follows:
\begin{equation}
    \mathbf{h}_i'= W \mathbf{h}_i + W_\text{intra} 
    \mean_{j \in \mathcal{N}_\text{in}(i)}\mathbf{h}_j + W_\text{inter} \mean_{j \in \mathcal{N}_\text{out}(i)}\mathbf{h}_j,
\end{equation}
where $\mathcal{N}_\text{in}(i)$ is a set of neighbors of the $i$-th node within the same fragment, and $\mathcal{N}_\text{out}(i)$ is the set of its neighbors outside the fragment. If any of these sets is empty, the corresponding term is removed from the formula.

To avoid leakage of information that is not crucial for prediction, we introduce a regularization term to the loss function, which is the $L_1$ norm of the interfragment weights $W_\text{inter}$. This term is controlled by a hyperparameter $\lambda$ that should be chosen on a case-by-case basis, but typically higher values lead to more interpretable results. The loss function in our model is:
\begin{equation}
    \mathcal{L} = \mathcal{L}_\text{pred} + \lambda \sum_{l=1}^L\left\Vert W_\text{inter}^{(l)} \right\Vert_1 
\end{equation}
where $\mathcal{L}_\text{pred}$ is the prediction error loss function (mean squared error for regression and cross entropy for classification), $W_\text{inter}^{(l)}$ are the interfragment weights in the $l$-th layer.

To balance the trade-off between model performance and interpretability, we perform a cross-validation testing multiple values of $\lambda$. The selected model is the one with the highest $\lambda$ values that is not significantly worse than the best model in terms of the target metric (e.g. RMSE or ROC AUC) according to the Wilcoxon signed-rank test.

\section{Results}

All experiments were conducted on an NVIDIA Grace Hopper GH200 (see Appendix~A for training details and implementation).

\subsection{Synthetic dataset benchmark}

Real-world molecular datasets only offer graph-level labels without assigning importance to specific atoms. Therefore, we chose to first use a synthetic dataset that allows for controlled and reliable assessment of Explainable AI (XAI) methods by providing direct ground-truth explanations.
We evaluate our method on the B-XAIC benchmark~\citep{proszewska2025b}, which is designed to compare GNN-based XAI methods in the molecular domain. The dataset includes various tasks focused on identifying different substructures: boron atoms (B), phosphorus atoms (P), halogens (X), indole rings, and pan-assay interference compounds (PAINS). The remaining two tasks focus on counting rings or atoms within rings. Each task has a known ground truth explanation, enabling a precise evaluation of the model's explanation quality.

\subsubsection{Metrics.}
To evaluate both model performance and explanation faithfulness, we use a two-part evaluation strategy. For classification, we report standard metrics such as AUROC, F1 Score, and Accuracy, where the last two metrics are available in the Appendix~B. For interpretability of explanations, we use the same as proposed by ~\citet{proszewska2025b}. 
\textbf{Subgraph Explanations (SE)} is the AUROC metric computed between model explanation and the ground-truth explanation for the positive examples. \textbf{Null Explanations (NE)} is the percentage of outliers in explained node attributions computed using the interquartile range method for the negative examples.

\begin{table*}[h!]
    \centering
    \small
    \setlength{\tabcolsep}{9.4pt} 
    \caption{AUROC score of various graph neural network architectures on the B-XAIC benchmark, including the GIN model used for explanations in the original benchmark. The last row shows the regularization value selected for SEAL on the validation set.}
    \label{tab:results_synth}
        \begin{tabular}{p{1.8cm} c c c  c c c c}
        \toprule
        Model & rings-count & rings-max & X & P & B & Indole & PAINS \\
        \midrule
            GIN  & \textbf{1.00} $\pm$ 0.00 & \underline{0.93} $\pm$ 0.02 & \textbf{1.00} $\pm$ 0.00 & \textbf{1.00} $\pm$ 0.00 & \textbf{1.00} $\pm$ 0.00 & \textbf{1.00} $\pm$ 0.00 & \textbf{0.99} $\pm$ 0.00 \\
            GCN  & \textbf{1.00} $\pm$ 0.00 & 0.82 $\pm$ 0.01 &  \textbf{1.00} $\pm$ 0.00 & \textbf{1.00} $\pm$ 0.00 &  \textbf{1.00} $\pm$ 0.00 & \underline{0.99} $\pm$ 0.00 & \underline{0.97} $\pm$ 0.00 \\
            GAT  & 0.88 $\pm$ 0.01 & 0.75 $\pm$ 0.02 &  \textbf{1.00} $\pm$ 0.00 &  \textbf{1.00} $\pm$ 0.00 & \textbf{1.00} $\pm$ 0.00 & 0.97 $\pm$ 0.00 & 0.92 $\pm$ 0.01 \\
            SEAL  & \underline{0.98} $\pm$ 0.01 & \textbf{0.99} $\pm$ 0.01 & 
            \textbf{1.00} $\pm$ 0.00 & \textbf{1.00} $\pm$ 0.00 & 
            \textbf{1.00} $\pm$ 0.00 & \textbf{1.00} $\pm$ 0.00 & \textbf{0.99} $\pm$ 0.00 \\
                        
        \midrule
        & $\lambda=10^{-3}$& $\lambda=2 $& $\lambda=2$& $\lambda=2$& $\lambda=2$& $\lambda=10^{-4}$& $\lambda=0$ \\
        \bottomrule
    \end{tabular}
\end{table*}

\begin{table*}[h!]
    \centering
    \small
    \setlength{\tabcolsep}{9pt} 
    \caption{
    Performance of model explanations on the B-XAIC benchmark. The subgraph explanation (SE) metric is employed for positive examples containing the relevant pattern. The last row shows the regularization value selected for SEAL on the validation set.
    }
    \label{tab:results_expl_synth}

    \begin{tabular}{ p{2cm} c c c c c c c
    }
        \toprule
        Model & rings-count & rings-max & X & P & B & Indole & PAINS \\
        \midrule
        Deconvolution  & 0.55 $\pm$ 0.24 & 0.38 $\pm$ 0.18 & 0.07 $\pm$ 0.00 & 0.90 $\pm$ 0.00 & 0.72 $\pm$ 0.01 & 0.36 $\pm$ 0.21 & 0.33 $\pm$ 0.01 \\
        GuidedBackprop  & \underline{0.69} $\pm$ 0.05 & \textbf{0.67} $\pm$ 0.02 & \underline{0.94} $\pm$ 0.01 & 0.85 $\pm$ 0.11 & \textbf{1.00} $\pm$ 0.00 & 0.85 $\pm$ 0.03 & 0.78 $\pm$ 0.02 \\
        IntegratedGradients  & 0.36 $\pm$ 0.00 & 0.34 $\pm$ 0.03 & \textbf{1.00} $\pm$ 0.00 & \textbf{1.00} $\pm$ 0.00 & \textbf{1.00} $\pm$ 0.00 & 0.84 $\pm$ 0.06 & 0.76 $\pm$ 0.02 \\
        Saliency  & 0.51 $\pm$ 0.04 & \underline{0.55} $\pm$ 0.03 & 0.92 $\pm$ 0.02 & \textbf{1.00} $\pm$ 0.00 & \textbf{1.00} $\pm$ 0.00 & \underline{0.87} $\pm$ 0.02 & \underline{0.81} $\pm$ 0.01 \\
        InputXGradient  & 0.49 $\pm$ 0.03 & 0.47 $\pm$ 0.05 & \textbf{1.00} $\pm$ 0.00 & \textbf{1.00} $\pm$ 0.00 & \textbf{1.00} $\pm$ 0.00 & 0.74 $\pm$ 0.05 & 0.54 $\pm$ 0.03 \\
        GNNExplainer  & 0.49 $\pm$ 0.01 & 0.50 $\pm$ 0.00 & 0.50 $\pm$ 0.00 & 0.51 $\pm$ 0.01 & 0.53 $\pm$ 0.05 & 0.53 $\pm$ 0.03 & 0.54 $\pm$ 0.06 \\
        SEAL  & \textbf{0.98} $\pm$ 0.01 & 0.34 $\pm$ 0.03 & \textbf{1.00} $\pm$ 0.00 & \underline{0.99} $\pm$ 0.00 & \underline{0.88} $\pm$ 0.01 & \textbf{0.96} $\pm$ 0.00 & \textbf{0.83} $\pm$ 0.01 \\
        \midrule
        & $\lambda=10^{-3}$& $\lambda=2 $& $\lambda=2$& $\lambda=2$& $\lambda=2$& $\lambda=10^{-4}$& $\lambda=0$ \\
        \bottomrule
    \end{tabular}
\end{table*}

\subsubsection{Models and baselines.}
We benchmark our method against a diverse set of GNN explanation techniques, spanning both mask-based and gradient-based approaches: GNNExplainer~\citep{ying2019gnnexplainer}, Saliency Maps~\citep{simonyan2014deepinsideconvolutionalnetworks}, InputXGradients~\citep{shrikumar2016not}, Integrated Gradients~\citep{sundararajan2017axiomatic}, Deconvolution~\citep{mahendran2016salient}, \citep{shrikumar2016not}, and Guided Backpropagation~\citep{springenberg2014striving}.

The evaluation of model performance is conducted for three GNN backbones: GCN~\citep{kipf2016semi}, GAT~\citep{velivckovic2017graph}, and GIN~\citep{xu2018powerful}. However, the explanation results are reported only for the GIN model due to its strongest performance across tasks, following the recommendations of the B-XAIC benchmark. Hyperparameters for all models, including SEAL, were optimized through random search. The search space and the optimal hyperparameters found are listed in Appendix~A.

\begin{figure*}
    \centering
    \includegraphics[width=\linewidth]{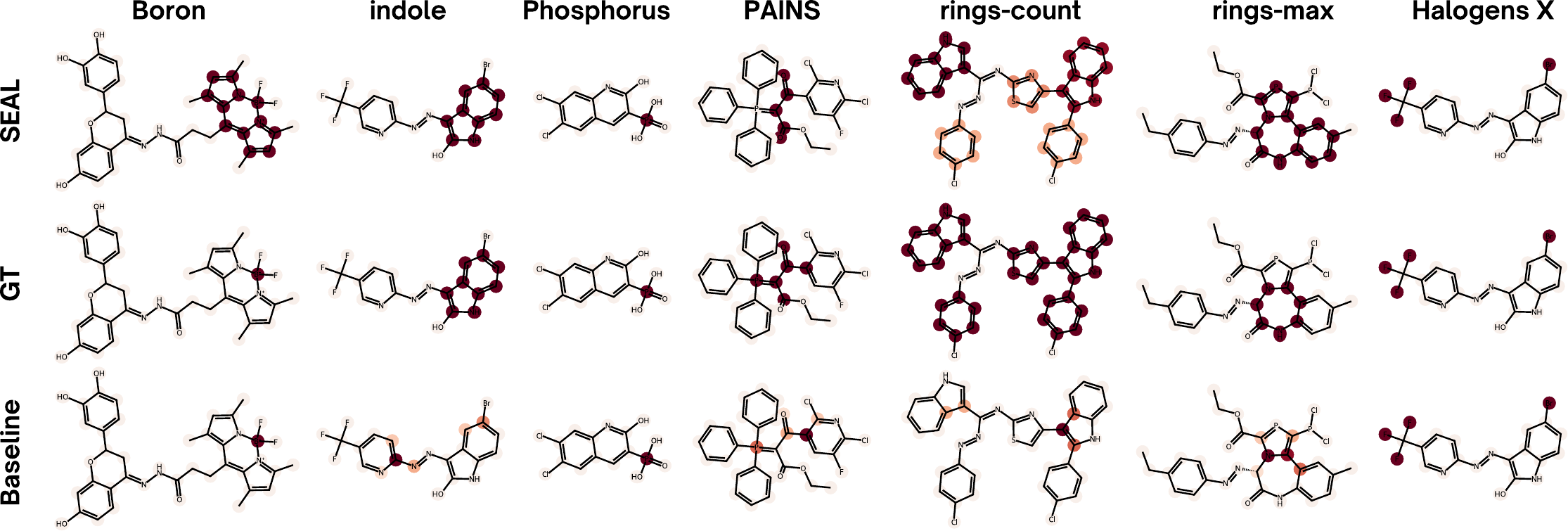}
    \caption{Node-level explanation examples for selected synthetic compounds from the B-XAIC dataset.
    Each column corresponds to a compound drawn from one of the tasks.
    From top to bottom: Our method (SEAL), the ground-truth explanation, and the other best explanation technique, denoted as Baseline.
    Highlighted nodes indicate substructures or atoms contributing most to the prediction. Red denotes a strong positive contribution to the predicted class.}
    \label{fig:mol_synths}
\end{figure*}

\begin{figure}[tb]
    \centering
    \includegraphics[width=0.9\columnwidth]{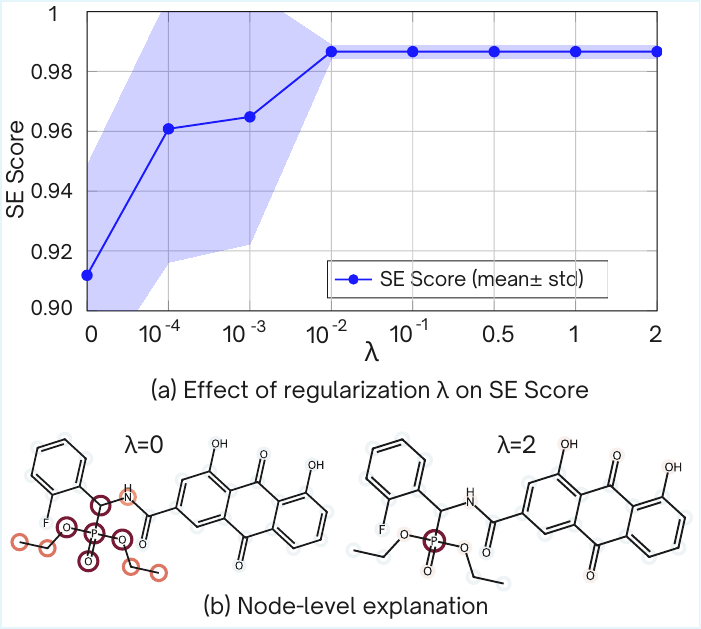}
    \caption{Effect of regularization on explanation quality in the phosphorus detection task. \textbf{(a)} Plot showing the relationship between $\lambda$ and SE. \textbf{(b)} Visual comparison of explanations generated with two different values of $\lambda$. High $\lambda$ values prevent the attribution of high contribution to neighboring fragments.  }
    \label{fig:mol_reg}
\end{figure}

\subsubsection{Results.}
Table~\ref{tab:results_synth} presents the results of the model performance on the synthetic dataset. Our model achieves competitive classification performance with an AUROC score comparable or even exceeding those of existing GNN architectures, while adding the capability of explaining its predictions. 
Table~\ref{tab:results_expl_synth} shows the results of the explanation evaluation, where our method yields significantly higher SE scores than other explainers on challenging tasks such as PAINS, rings-count, and indole. In the halogens (X) and phosphorus (P) tasks, our performance is on par with that of the strongest baselines, reflecting the relative ease of localizing single-atom substructures. 
A significant decline in performance appears in the boron (B) task due to its frequent appearance in complex substructures that our extended BRICS decomposition cannot efficiently segment (Figure~\ref{fig:mol_synths}). The largest ring pattern, like boron, mostly occurs in larger substructures, but it also presents an additional challenge due to its highly imbalanced nature, with a low percentage of positive samples across the dataset. The evaluation of the negative examples is provided in the Appendix~B.

Figure~\ref{fig:mol_synths} presents an example explanation generated by our model for randomly selected molecules. It includes both correct explanations and failure cases where larger fragments are highlighted. The figure also compares the ground-truth annotations with the outputs of the best-performing baseline method for each task. SEAL effectively highlights chemically meaningful subgraphs, whereas other approaches tend to assign the prediction to only a few atoms, distributing smaller weights across the entire graph. More examples can be found in Appendix E.

\subsubsection{Regularization}
SEAL dynamically adapts its $\lambda$ parameter to maximize interpretability without sacrificing performance. We perform an ablation study by varying the regularization parameter $\lambda$, which determines how much message passing is restricted in our model. We discover that the optimal value of $\lambda$ depends on the specific task. In some cases, limiting message propagation improves explanation by preventing information from leaking across irrelevant parts of the graph. For example, in the phosphorus (P) task, increasing \(\lambda\) leads to a notable improvement in subgraph explanation quality, as shown in Figure~\ref{fig:mol_reg}. This indicates that stronger regularization helps the model concentrate on localized substructures without causing over-smoothing. In contrast, other tasks, such as PAINS detection, require the information to flow across distant parts of the graph. In these cases, we find that the best explanation performance occurs when $\lambda = 0$. Notably, low $\lambda$ values cause information leakage into adjacent fragments, whereas higher $\lambda$ values provide more focused and faithful explanations.

\subsection{Evaluation on real-world datasets}

Evaluating explanation performance on real-world molecular datasets remains a challenging task. Unlike synthetic benchmarks, these datasets do not provide ground-truth explanations that identify which atoms or substructures are responsible for the prediction. Additionally, most of the molecular properties relevant for real-world applications are much more complex, often involving long-range interactions between fragments or features based on the spatial distribution of atoms. To benchmark our method with real-world compounds, we follow the same setup as used for the synthetic dataset.

\begin{table}[tb]
    \centering
    \small
    \setlength{\tabcolsep}{10.5pt}
    \caption{Scores of various GNN architectures on the real-world datasets (Solubility, hERG, and CYP), using MAE and AUROC as metrics. The last row shows the regularization value selected for SEAL on the validation set.}
    \label{tab:real_world_scores}

    \begin{tabular}{ l c c c}
        \toprule
        & AqSol & hERG & CYP2C9 \\
        Model & MAE $\downarrow$ & AUROC $\uparrow$ & AUROC $\uparrow$ \\
         \midrule
        GIN  & \textbf{0.41} $\pm$ 0.01 & \textbf{0.86} $\pm$ 0.01 & \textbf{0.86} $\pm$ 0.01 \\
        GAT  & 0.57 $\pm$ 0.01 & 0.70 $\pm$ 0.01 & 0.68 $\pm$ 0.01 \\
        GCN  & 0.49 $\pm$ 0.02 & 0.81 $\pm$ 0.03 & \underline{0.84} $\pm$ 0.01 \\
        SEAL & \underline{0.47} $\pm$ 0.01 & \underline{0.85} $\pm$ 0.01 & 0.81 $\pm$ 0.01 \\      
        \midrule
        & $\lambda=10^{-4}$ & $\lambda=10^{-4}$ & $\lambda=2$ \\

     \bottomrule
    \end{tabular}
\end{table}

\begin{table*}[h!]
    \centering
    \small
    \setlength{\tabcolsep}{7.8pt} 
        \caption{Performance of model explanations on real-world datasets (hERG and CYP2C9). Explanations are evaluated using Fidelity metrics at 10\%, 20\%, and 30\% masking thresholds, representing the proportion of most important atoms (nodes) either removed or retained during the evaluation.}
    \label{tab:results_expl_real}

    \begin{tabular}{ c l c  c c c c c
    }
        \toprule
        & Model & $\text{Fidelity}_{10}+ \uparrow$ & $\text{Fidelity}_{10}- \downarrow$ & $\text{Fidelity}_{20}+ \uparrow$ & $\text{Fidelity}_{20}- \downarrow$ & $\text{Fidelity}_{30}+ \uparrow$ & $\text{Fidelity}_{30}- \downarrow$ \\
        \midrule
        \parbox[t]{2mm}{\multirow{7}{*}{\rotatebox[origin=c]{90}{hERG}}} & Deconvolution  & 0.33 $\pm$ 0.06 & \underline{0.49} $\pm$ 0.02 & 0.45 $\pm$ 0.04 & 0.49 $\pm$ 0.02 & 0.48 $\pm$ 0.02 & 0.49 $\pm$ 0.02 \\
        & GuidedBackprop  & 0.39 $\pm$ 0.03 & \underline{0.49} $\pm$ 0.02 & 0.44 $\pm$ 0.04 & 0.49 $\pm$ 0.02 & 0.47 $\pm$ 0.02 & 0.49 $\pm$ 0.02 \\
        & IntegratedGradients  & \underline{0.47} $\pm$ 0.07 & \underline{0.49} $\pm$ 0.02 & \underline{0.54} $\pm$ 0.10 & \underline{0.47} $\pm$ 0.03 & \underline{0.58} $\pm$ 0.19 & \underline{0.41} $\pm$ 0.16 \\
        & Saliency  & 0.35 $\pm$ 0.03 & \underline{0.49} $\pm$ 0.02 & 0.42 $\pm$ 0.03 & 0.49 $\pm$ 0.02 & 0.46 $\pm$ 0.02 & 0.49 $\pm$ 0.02 \\
        & InputXGradient  & 0.33 $\pm$ 0.05 & \underline{0.49} $\pm$ 0.02 & 0.40 $\pm$ 0.04 & 0.49 $\pm$ 0.02 & 0.44 $\pm$ 0.04 & 0.50 $\pm$ 0.02 \\
        & GNNExplainer  & 0.42 $\pm$ 0.15 & \underline{0.49} $\pm$ 0.02 & 0.46 $\pm$ 0.03 & 0.49 $\pm$ 0.02 & 0.50 $\pm$ 0.05 & 0.47 $\pm$ 0.04 \\
        & SEAL ($\lambda = 10^{-4}$) & \textbf{0.63} $\pm$ 0.01 & \textbf{0.09} $\pm$ 0.03 & \textbf{0.71} $\pm$ 0.01 & \textbf{0.07} $\pm$ 0.02 & \textbf{0.78} $\pm$ 0.01 & \textbf{0.05} $\pm$ 0.02 \\
        \midrule
        \parbox[t]{2mm}{\multirow{7}{*}{\rotatebox[origin=c]{90}{CYP2C9}}} & Deconvolution  & 0.24 $\pm$ 0.02 & 0.40 $\pm$ 0.09 & 0.31 $\pm$ 0.03 & 0.37 $\pm$ 0.03 & 0.34 $\pm$ 0.03 & 0.37 $\pm$ 0.03 \\
        & GuidedBackprop  & 0.33 $\pm$ 0.03 & 0.40 $\pm$ 0.10 & 0.34 $\pm$ 0.04 & 0.39 $\pm$ 0.07 & 0.35 $\pm$ 0.03 & 0.38 $\pm$ 0.05 \\
        & IntegratedGradients  & \underline{0.45} $\pm$ 0.08 & \underline{0.29} $\pm$ 0.16 & \textbf{0.59} $\pm$ 0.14 & \underline{0.29} $\pm$ 0.16 & \textbf{0.68} $\pm$ 0.21 & \underline{0.29} $\pm$ 0.16 \\
        & Saliency  & 0.32 $\pm$ 0.04 & 0.38 $\pm$ 0.06 & 0.37 $\pm$ 0.03 & 0.36 $\pm$ 0.03 & 0.38 $\pm$ 0.04 & 0.36 $\pm$ 0.03 \\
        & InputXGradient  & 0.31 $\pm$ 0.02 & 0.41 $\pm$ 0.12 & 0.34 $\pm$ 0.02 & 0.41 $\pm$ 0.11 & 0.35 $\pm$ 0.03 & 0.41 $\pm$ 0.11 \\
        & GNNExplainer  & 0.30 $\pm$ 0.02 & 0.37 $\pm$ 0.03 & 0.43 $\pm$ 0.10 & 0.43 $\pm$ 0.16 & 0.42 $\pm$ 0.08 & 0.33 $\pm$ 0.08 \\        
        & SEAL ($\lambda=2$) & \textbf{0.52} $\pm$ 0.02 & \textbf{0.04} $\pm$ 0.05 & \underline{0.57} $\pm$ 0.03 & \textbf{0.03} $\pm$ 0.04 & \underline{0.66} $\pm$ 0.03 & \textbf{0.01} $\pm$ 0.01 \\ 
        \bottomrule
    \end{tabular}
\end{table*}

\begin{figure}[tb]
    \centering
    \includegraphics[width=\linewidth]
    {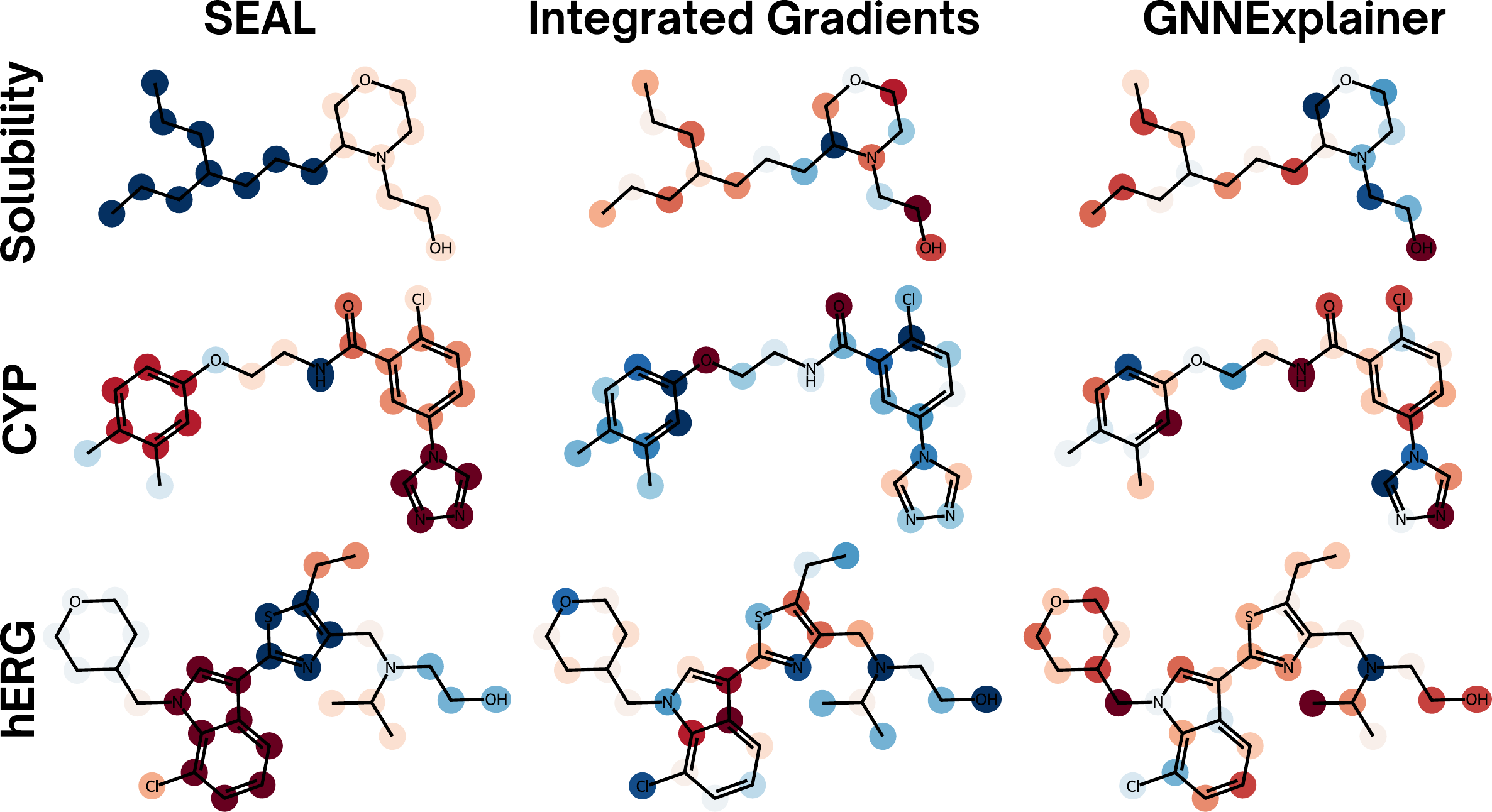}
    \caption{Node-level explanation examples for selected compounds from the Aqueous Solubility, CYP 2C9, and hERG datasets.
    Each row corresponds to a compound from one of the datasets.
    From left to right: Our method (SEAL), a gradient-based method (Integrated Gradients), a perturbation-based method (GNNExplainer).
    Highlighted nodes indicate substructures or atoms contributing most to the prediction. Red denotes a strong positive contribution, while blue indicates a strong negative influence on the predicted property.}
    \label{fig:results_real_vis}
\end{figure}

\subsubsection{Datasets.}
We evaluate our method on three real-world molecular property prediction datasets from TDC~\citep{huang2021therapeutics}. hERG inhibition~\citep{karim2021cardiotox} is a binary classification task that includes molecular structures labeled as hERG blockers or non-blockers, a property critical for cardiac safety assessment in drug development. CYP450 2C9 inhibition \citep{veith2009comprehensive} is a binary classification task that focuses on inhibition of the cytochrome P450 2C9 enzyme, which is central to drug metabolism. Aqueous Solubility (AqSol)~\citep{sorkun2019aqsoldb} is a regression task that contains compounds with measured solubility in water.

\subsubsection{Metrics.}
To evaluate explanations across different methods and fairly compare them with our model, we decided to evaluate on standard Positive and Negative fidelity. For all models, we mask node features at the input level, ensuring a fair comparison. Fidelity definition that we decided to use:
\begin{itemize}
    \item \textbf{Positive Fidelity}: we mask the most important nodes and check how the prediction has changed.
    \item \textbf{Negative Fidelity}: we retain the most important nodes and check if the prediction is the same.
\end{itemize}

For classification tasks, fidelity is measured by the proportion of times the predicted class changes after masking. We evaluate masking at thresholds of $ 10\%, 20\%$, and $30\% $ of nodes, ensuring that the most relevant atoms are included in explanations without exceeding the specified percentage. The advantage of our model is that the prediction is a sum of contributions, so we can directly mask contributions instead of masking the input graph nodes and features, which leads to out-of-distribution samples. An ablation study on different masking strategies in SEAL is presented in Appendix~C.

\subsubsection{Results.}
Table~\ref{tab:real_world_scores} shows our model's performance on real-world molecular datasets. In terms of predictive AUROC, our model matches or exceeds the GCN and GAT baselines, while also providing the advantage of interpretability. Compared to the more robust GIN baseline, our model remains competitive without a significant loss in performance. F1 score, Accuracy, and RMSE can be found in Appendix~B.

For the explanation evaluation, our model consistently achieves strong fidelity scores. On the hERG dataset, it outperforms all other methods in both positive and negative fidelity with a significant margin. On the CYP dataset, our approach again exceeds existing techniques; only Integrated Gradients shows competitive results but experiences a noticeable drop in negative fidelity, demonstrating the strength of our method in capturing relevant structures.

For regression tasks like Solubility, evaluating explanation quality is more difficult, and not all explainers are well-defined in this context. Nevertheless, our method attains reasonable fidelity values compared to other explanation methods. These results are detailed in Appendix~B.

\subsubsection{Qualitative examples.}
In Figure~\ref{fig:results_real_vis}, we present qualitative visualizations of explanations generated by our model compared to the top-performing baselines for the AqSolDB, CYP2C9, and hERG datasets. While other methods tend to produce scattered or noisy explanations, our model yields more compact and interpretable substructures. These results show that our approach captures chemically plausible explanations that are easier to interpret and often more localized, especially in tasks like solubility, where polarity and solubility driving fragments are correctly emphasized. More examples can be found in Appendix E.

\subsubsection{Discussion.}
Across all evaluated tasks, our model consistently demonstrates strong performance, both in terms of the prediction performance and explanation faithfulness, while providing an added benefit of interpretability. We got strong and comparative results compared to the GNN baselines. Furthermore, we also outperform the other explainer techniques, in terms of positive and negative fidelity. 

By combining strong quantitative results with interpretability aligned with chemical intuition, SEAL proves to be a reliable tool for understanding model decisions across both real and synthetic molecular data. However, fidelity is not a perfect metric because it compares model predictions for the real molecule and its masked counterpart, which has some nodes or their features removed. This artificial reference point is an out-of-distribution sample for the model, so its prediction should be approached with caution. To further support these findings and assess the practical usefulness of the explanations, also keeping in mind that fidelity is not the most informative metric, we conducted a follow-up user study with expert chemists. This enables us to determine whether the generated explanations are not only mathematically accurate but also chemically meaningful and trustworthy in real-world applications.

\begin{figure}[tb]
    \centering
    \includegraphics[width=\linewidth]{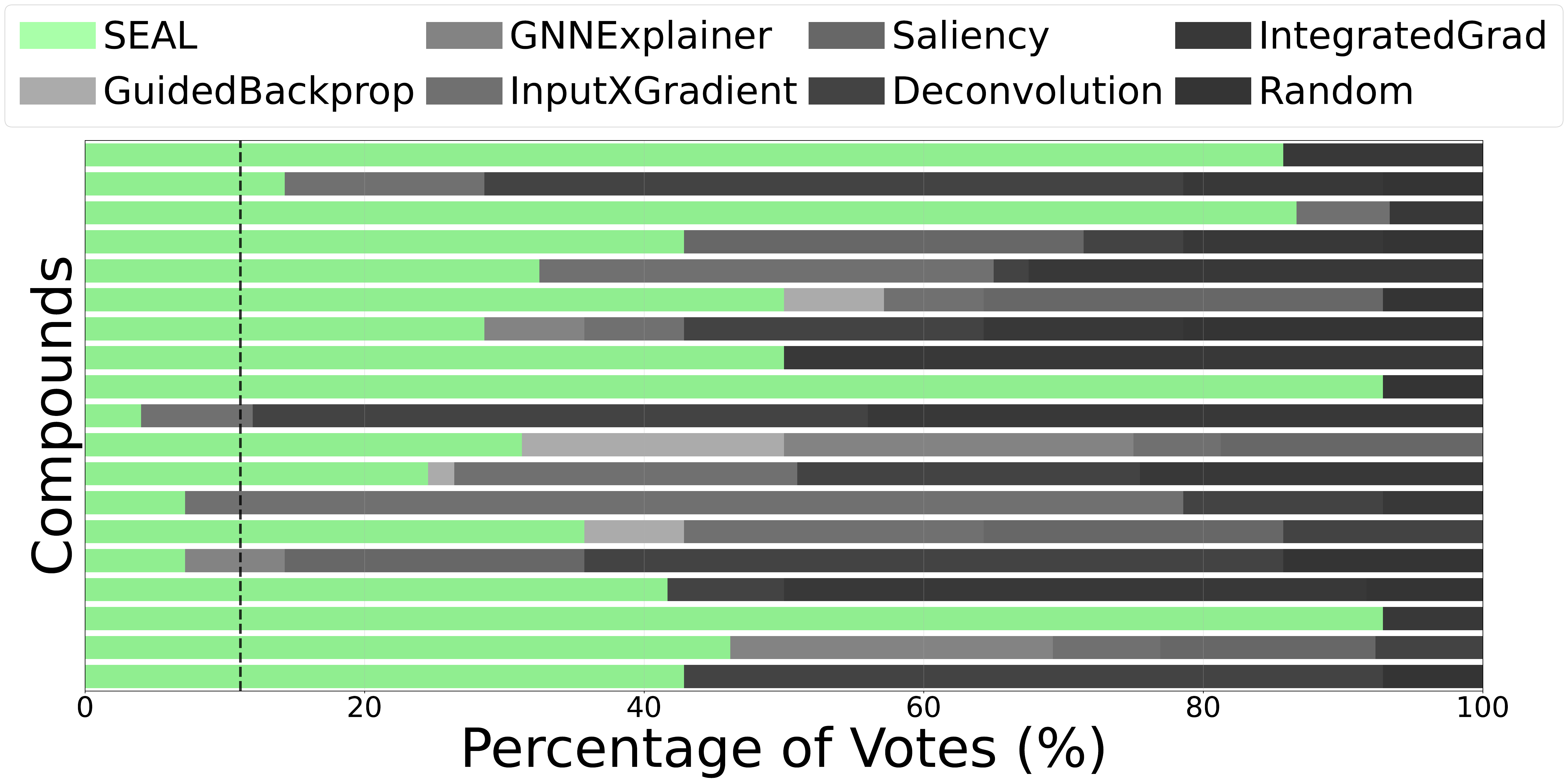}
\caption{Distribution of votes per explanation method across all 19 compound-based questions. Each bar represents a compound on the y-axis, and the x-axis shows the percentage of user votes received by each method. Different colors correspond to different explanation techniques. The black dashed line denotes the expected value for random guessing.}
    \label{fig:vote_distribution_by_method}
\end{figure}

\subsection{User study}

While the results from synthetic and real-world benchmarks appear promising, the evaluation metrics used often fail to fully capture the human-aligned interpretability of explanations generated by XAI techniques. These quantitative metrics measure how explanations influence model predictions but do not ensure that the highlighted atoms or substructures are chemically meaningful or align with expert understanding. Explanations that perform well on such metrics can still be useless and poorly understandable to chemistry experts. To assess whether SEAL explanations are interpretable and to evaluate if the explanations produced by different methods align with chemistry knowledge and intuition, we decided to conduct an examination across chemistry experts using various methods and randomly selected compounds.

\subsubsection{Experimental design.}
Our user study consisted of 19 questions featuring various randomly selected compounds. Each question included nine different explanations generated by SEAL and seven explainability methods (the same methods used in previous experiments). In addition, two random explanations were added as separate options, and all answers were shuffled. Each explanation highlights atoms that were predicted to increase the compound's solubility in water (aqueous solubility). Participants had to choose the explanation they believed best highlights the atoms most responsible for increasing solubility. All participants were experts with at least a master's degree in chemistry, ensuring a representative sample with relevant knowledge to evaluate factors influencing solubility. Each question included:\\
\begin{itemize}
    \item a question saying: "Which explanation best highlights the atoms that contribute most to increasing the solubility of this compound?",
    \item molecular structure of a compound,
    \item nine explanation visualizations with important atoms marked in red, as predicted by various explainers.
\end{itemize}

The study included two randomized control conditions. In the first one, randomly selected atoms are assigned importance. The second one assigns importance to random BRICS structures. To ensure that all methods have the same chances, we fixed the number of highlighted atoms to approximately half of all atoms in the compounds. In case there were two explainers that highlight the same atoms, we replaced redundant ones with randomly generated alternatives. The participants were blinded to the source of each explanation and instructed to focus solely on the plausibility of the provided explanation.

\subsubsection{Discussion.}
Among the methods tested, SEAL was chosen most frequently in 14 out of 19 questions, significantly outperforming all other methods. The remaining votes were distributed as follows: Deconvolution and IntegratedGradients  were chosen 5 times, and InputXGradients was selected 3 times. Other methods (Saliency, GNNExplainer, and Guided Backprop) did not win in any of the questions. The distribution of votes between methods in each question is shown in Figure~\ref{fig:vote_distribution_by_method}. All compounds and visualizations that were used for this user study are listed in the Appendix~D.
The user study confirms that our method, SEAL, provides explanations that align more closely with human intuition and chemical understanding. It was favored over other techniques, emphasizing its ability to produce meaningful and understandable atom-level attributions.

\section{Conclusions}

In this study, we presented SEAL, an interpretable GNN model for molecular property prediction. The method breaks down an input molecular graph into fragments that are used to calculate their contributions to the final model prediction, which is a sum of these contributions and the bias term. A regularization term is employed to regulate the exchange of information between fragments in our new SEAL-GCN architecture. Experiments on both synthetic and real-world datasets show that SEAL can accurately identify important molecular fragments. The practical usefulness of our method is further validated by a user study, where SEAL is the most frequently chosen method for explaining solubility in water.

\section{Acknowledgments}

This work of B.Z. and T.D. was funded by the "Interpretable and Interactive Multimodal Retrieval in Drug Discovery" project. The "Interpretable and Interactive Multimodal Retrieval in Drug Discovery” project (FENG.02.02-IP.05-0040/23) is carried out within the First Team programme of the Foundation for Polish Science co-financed by the European Union under the European Funds for Smart Economy 2021-2027 (FENG). The work of S.M. was supported by the National Science Centre, Poland (grant no. 2022/47/B/ST6/03397). We gratefully acknowledge Polish high-performance computing infrastructure PLGrid (HPC Center: ACK Cyfronet AGH) for providing computer facilities and support within computational grant no. PLG/2025/018272. Some experiments were performed on servers purchased with funds from the flagship project entitled “Artificial Intelligence Computing Core Facility” from the DigiWorld Priority Research Area within the Excellence Initiative – Research University program at Jagiellonian University in Krakow.

\bibliography{references}

\begin{thebibliography}{42}
\providecommand{\natexlab}[1]{#1}

\bibitem[{Ba, Kiros, and Hinton(2016)}]{ba2016layer}
Ba, J.~L.; Kiros, J.~R.; and Hinton, G.~E. 2016.
\newblock Layer normalization.
\newblock \emph{arXiv preprint arXiv:1607.06450}.

\bibitem[{Cao et~al.(2024)Cao, He, Cui, Zhang, Zhang, and Zhang}]{cao2024group}
Cao, P.-Y.; He, Y.; Cui, M.-Y.; Zhang, X.-M.; Zhang, Q.; and Zhang, H.-Y. 2024.
\newblock Group graph: a molecular graph representation with enhanced performance, efficiency and interpretability.
\newblock \emph{Journal of Cheminformatics}, 16(1): 133.

\bibitem[{Degen et~al.(2008)Degen, Wegscheid-Gerlach, Zaliani, and Rarey}]{degen2008art}
Degen, J.; Wegscheid-Gerlach, C.; Zaliani, A.; and Rarey, M. 2008.
\newblock On the art of compiling and using'drug-like'chemical fragment spaces.
\newblock \emph{ChemMedChem}, 3(10): 1503.

\bibitem[{Elton et~al.(2019)Elton, Boukouvalas, Fuge, and Chung}]{elton2019deep}
Elton, D.~C.; Boukouvalas, Z.; Fuge, M.~D.; and Chung, P.~W. 2019.
\newblock Deep learning for molecular design—a review of the state of the art.
\newblock \emph{Molecular Systems Design \& Engineering}, 4(4): 828--849.

\bibitem[{Feng et~al.(2022)Feng, Liu, Yang, Tang, Du, and Hu}]{fengdegree}
Feng, Q.; Liu, N.; Yang, F.; Tang, R.; Du, M.; and Hu, X. 2022.
\newblock {DEGREE}: Decomposition Based Explanation for Graph Neural Networks.
\newblock In \emph{International Conference on Learning Representations}.

\bibitem[{Gilmer et~al.(2020)Gilmer, Schoenholz, Riley, Vinyals, and Dahl}]{gilmer2020message}
Gilmer, J.; Schoenholz, S.~S.; Riley, P.~F.; Vinyals, O.; and Dahl, G.~E. 2020.
\newblock Message passing neural networks.
\newblock In \emph{Machine learning meets quantum physics}, 199--214. Springer.

\bibitem[{Gui et~al.(2023)Gui, Yuan, Wang, Lao, Li, and Ji}]{gui2023flowx}
Gui, S.; Yuan, H.; Wang, J.; Lao, Q.; Li, K.; and Ji, S. 2023.
\newblock Flowx: Towards explainable graph neural networks via message flows.
\newblock \emph{IEEE Transactions on Pattern Analysis and Machine Intelligence}, 46(7): 4567--4578.

\bibitem[{Henderson, Clevert, and Montanari(2021)}]{henderson2021improving}
Henderson, R.; Clevert, D.-A.; and Montanari, F. 2021.
\newblock Improving molecular graph neural network explainability with orthonormalization and induced sparsity.
\newblock In \emph{International Conference on Machine Learning}, 4203--4213. PMLR.

\bibitem[{Huang et~al.(2021)Huang, Fu, Gao, Zhao, Roohani, Leskovec, Coley, Xiao, Sun, and Zitnik}]{huang2021therapeutics}
Huang, K.; Fu, T.; Gao, W.; Zhao, Y.; Roohani, Y.; Leskovec, J.; Coley, C.~W.; Xiao, C.; Sun, J.; and Zitnik, M. 2021.
\newblock Therapeutics data commons: Machine learning datasets and tasks for drug discovery and development.
\newblock \emph{arXiv preprint arXiv:2102.09548}.

\bibitem[{Karim et~al.(2021)Karim, Lee, Balle, and Sattar}]{karim2021cardiotox}
Karim, A.; Lee, M.; Balle, T.; and Sattar, A. 2021.
\newblock CardioTox net: a robust predictor for hERG channel blockade based on deep learning meta-feature ensembles.
\newblock \emph{Journal of Cheminformatics}, 13(1): 60.

\bibitem[{Kipf(2016)}]{kipf2016semi}
Kipf, T. 2016.
\newblock Semi-Supervised Classification with Graph Convolutional Networks.
\newblock \emph{arXiv preprint arXiv:1609.02907}.

\bibitem[{Lee et~al.(2023)Lee, Park, Choi, Kim, Kim, Han, Kang, Kang, and Son}]{lee2023multi}
Lee, S.; Park, H.; Choi, C.; Kim, W.; Kim, K.~K.; Han, Y.-K.; Kang, J.; Kang, C.-J.; and Son, Y. 2023.
\newblock Multi-order graph attention network for water solubility prediction and interpretation.
\newblock \emph{Scientific Reports}, 13(1): 957.

\bibitem[{Mahendran and Vedaldi(2016)}]{mahendran2016salient}
Mahendran, A.; and Vedaldi, A. 2016.
\newblock Salient deconvolutional networks.
\newblock In \emph{European conference on computer vision}, 120--135. Springer.

\bibitem[{Maziarka et~al.(2024)Maziarka, Majchrowski, Danel, Gai{\'n}ski, Tabor, Podolak, Morkisz, and Jastrz{\k{e}}bski}]{maziarka2024relative}
Maziarka, {\L}.; Majchrowski, D.; Danel, T.; Gai{\'n}ski, P.; Tabor, J.; Podolak, I.; Morkisz, P.; and Jastrz{\k{e}}bski, S. 2024.
\newblock Relative molecule self-attention transformer.
\newblock \emph{Journal of Cheminformatics}, 16(1): 3.

\bibitem[{Murcko(1995)}]{murcko1995computational}
Murcko, M.~A. 1995.
\newblock Computational methods to predict binding free energy in ligand-receptor complexes.
\newblock \emph{Journal of medicinal chemistry}, 38(26): 4953--4967.

\bibitem[{Ponzoni, P{\'a}ez~Prosper, and Campillo(2023)}]{ponzoni2023explainable}
Ponzoni, I.; P{\'a}ez~Prosper, J.~A.; and Campillo, N.~E. 2023.
\newblock Explainable artificial intelligence: A taxonomy and guidelines for its application to drug discovery.
\newblock \emph{Wiley Interdisciplinary Reviews: Computational Molecular Science}, 13(6): e1681.

\bibitem[{Proszewska, Danel, and Rymarczyk(2025)}]{proszewska2025b}
Proszewska, M.; Danel, T.; and Rymarczyk, D. 2025.
\newblock B-XAIC Dataset: Benchmarking Explainable AI for Graph Neural Networks Using Chemical Data.
\newblock \emph{arXiv preprint arXiv:2505.22252}.

\bibitem[{Rong et~al.(2020)Rong, Bian, Xu, Xie, Wei, Huang, and Huang}]{rong2020self}
Rong, Y.; Bian, Y.; Xu, T.; Xie, W.; Wei, Y.; Huang, W.; and Huang, J. 2020.
\newblock Self-supervised graph transformer on large-scale molecular data.
\newblock \emph{Advances in neural information processing systems}, 33: 12559--12571.

\bibitem[{Rymarczyk, Dobrowolski, and Danel(2023)}]{rymarczyk2023progrest}
Rymarczyk, D.; Dobrowolski, D.; and Danel, T. 2023.
\newblock {ProGReST}: Prototypical graph regression soft trees for molecular property prediction.
\newblock In \emph{Proceedings of the 2023 SIAM International Conference on Data Mining (SDM)}, 379--387. SIAM.

\bibitem[{Shrikumar et~al.(2016)Shrikumar, Greenside, Shcherbina, and Kundaje}]{shrikumar2016not}
Shrikumar, A.; Greenside, P.; Shcherbina, A.; and Kundaje, A. 2016.
\newblock Not just a black box: Learning important features through propagating activation differences.
\newblock \emph{arXiv preprint arXiv:1605.01713}.

\bibitem[{Simonyan, Vedaldi, and Zisserman(2014)}]{simonyan2014deepinsideconvolutionalnetworks}
Simonyan, K.; Vedaldi, A.; and Zisserman, A. 2014.
\newblock Deep Inside Convolutional Networks: Visualising Image Classification Models and Saliency Maps.

\bibitem[{Sorkun, Khetan, and Er(2019)}]{sorkun2019aqsoldb}
Sorkun, M.~C.; Khetan, A.; and Er, S. 2019.
\newblock AqSolDB, a curated reference set of aqueous solubility and 2D descriptors for a diverse set of compounds.
\newblock \emph{Scientific data}, 6(1): 143.

\bibitem[{Springenberg et~al.(2014)Springenberg, Dosovitskiy, Brox, and Riedmiller}]{springenberg2014striving}
Springenberg, J.~T.; Dosovitskiy, A.; Brox, T.; and Riedmiller, M. 2014.
\newblock Striving for simplicity: The all convolutional net.
\newblock \emph{arXiv preprint arXiv:1412.6806}.

\bibitem[{Sundararajan, Taly, and Yan(2017)}]{sundararajan2017axiomatic}
Sundararajan, M.; Taly, A.; and Yan, Q. 2017.
\newblock Axiomatic attribution for deep networks.
\newblock In \emph{International conference on machine learning}, 3319--3328. PMLR.

\bibitem[{Talanquer(2022)}]{talanquer2022complexity}
Talanquer, V. 2022.
\newblock The complexity of reasoning about and with chemical representations.
\newblock \emph{Jacs Au}, 2(12): 2658--2669.

\bibitem[{Tanaka, Okamoto, and Bersohn(2010)}]{tanaka2010construction}
Tanaka, A.; Okamoto, H.; and Bersohn, M. 2010.
\newblock Construction of functional group reactivity database under various reaction conditions automatically extracted from reaction database in a synthesis design system.
\newblock \emph{Journal of chemical information and modeling}, 50(3): 327--338.

\bibitem[{Tuttle et~al.(2023)Tuttle, Brackman, Sorourifar, Paulson, and Zhang}]{tuttle2023predicting}
Tuttle, M.~R.; Brackman, E.~M.; Sorourifar, F.; Paulson, J.; and Zhang, S. 2023.
\newblock Predicting the solubility of organic energy storage materials based on functional group identity and substitution pattern.
\newblock \emph{The Journal of Physical Chemistry Letters}, 14(5): 1318--1325.

\bibitem[{Veith et~al.(2009)Veith, Southall, Huang, James, Fayne, Artemenko, Shen, Inglese, Austin, Lloyd et~al.}]{veith2009comprehensive}
Veith, H.; Southall, N.; Huang, R.; James, T.; Fayne, D.; Artemenko, N.; Shen, M.; Inglese, J.; Austin, C.~P.; Lloyd, D.~G.; et~al. 2009.
\newblock Comprehensive characterization of cytochrome P450 isozyme selectivity across chemical libraries.
\newblock \emph{Nature biotechnology}, 27(11): 1050--1055.

\bibitem[{Veli{\v{c}}kovi{\'c} et~al.(2017)Veli{\v{c}}kovi{\'c}, Cucurull, Casanova, Romero, Lio, and Bengio}]{velivckovic2017graph}
Veli{\v{c}}kovi{\'c}, P.; Cucurull, G.; Casanova, A.; Romero, A.; Lio, P.; and Bengio, Y. 2017.
\newblock Graph attention networks.
\newblock \emph{arXiv preprint arXiv:1710.10903}.

\bibitem[{von Korff and Sander(2006)}]{von2006toxicity}
von Korff, M.; and Sander, T. 2006.
\newblock Toxicity-indicating structural patterns.
\newblock \emph{Journal of chemical information and modeling}, 46(2): 536--544.

\bibitem[{Vu and Thai(2020)}]{vu2020pgm}
Vu, M.; and Thai, M.~T. 2020.
\newblock Pgm-explainer: Probabilistic graphical model explanations for graph neural networks.
\newblock \emph{Advances in neural information processing systems}, 33: 12225--12235.

\bibitem[{Wang et~al.(2025)Wang, Min, Li, and Wu}]{wangfragformer}
Wang, J.; Min, Y.; Li, M.; and Wu, J. 2025.
\newblock FragFormer: A Fragment-based Representation Learning Framework for Molecular Property Prediction.
\newblock \emph{Transactions on Machine Learning Research}.

\bibitem[{Wieder et~al.(2020)Wieder, Kohlbacher, Kuenemann, Garon, Ducrot, Seidel, and Langer}]{wieder2020compact}
Wieder, O.; Kohlbacher, S.; Kuenemann, M.; Garon, A.; Ducrot, P.; Seidel, T.; and Langer, T. 2020.
\newblock A compact review of molecular property prediction with graph neural networks.
\newblock \emph{Drug Discovery Today: Technologies}, 37: 1--12.

\bibitem[{Wu et~al.(2023)Wu, Wang, Du, Jiang, Kang, Li, Pan, Deng, Cao, Hsieh et~al.}]{wu2023chemistry}
Wu, Z.; Wang, J.; Du, H.; Jiang, D.; Kang, Y.; Li, D.; Pan, P.; Deng, Y.; Cao, D.; Hsieh, C.-Y.; et~al. 2023.
\newblock Chemistry-intuitive explanation of graph neural networks for molecular property prediction with substructure masking.
\newblock \emph{Nature communications}, 14(1): 2585.

\bibitem[{Xiong et~al.(2019)Xiong, Wang, Liu, Zhong, Wan, Li, Li, Luo, Chen, Jiang et~al.}]{xiong2019pushing}
Xiong, Z.; Wang, D.; Liu, X.; Zhong, F.; Wan, X.; Li, X.; Li, Z.; Luo, X.; Chen, K.; Jiang, H.; et~al. 2019.
\newblock Pushing the boundaries of molecular representation for drug discovery with the graph attention mechanism.
\newblock \emph{Journal of medicinal chemistry}, 63(16): 8749--8760.

\bibitem[{Xu et~al.(2018)Xu, Hu, Leskovec, and Jegelka}]{xu2018powerful}
Xu, K.; Hu, W.; Leskovec, J.; and Jegelka, S. 2018.
\newblock How powerful are graph neural networks?
\newblock \emph{arXiv preprint arXiv:1810.00826}.

\bibitem[{Ying et~al.(2019)Ying, Bourgeois, You, Zitnik, and Leskovec}]{ying2019gnnexplainer}
Ying, Z.; Bourgeois, D.; You, J.; Zitnik, M.; and Leskovec, J. 2019.
\newblock Gnnexplainer: Generating explanations for graph neural networks.
\newblock \emph{Advances in neural information processing systems}, 32.

\bibitem[{Yuan et~al.(2021)Yuan, Yu, Wang, Li, and Ji}]{yuan2021explainability}
Yuan, H.; Yu, H.; Wang, J.; Li, K.; and Ji, S. 2021.
\newblock On explainability of graph neural networks via subgraph explorations.
\newblock In \emph{International conference on machine learning}, 12241--12252. PMLR.

\bibitem[{Zhang et~al.(2023)Zhang, Xu, He, Guo, and Cui}]{zhang2023comprehensive}
Zhang, X.; Xu, Y.; He, W.; Guo, W.; and Cui, L. 2023.
\newblock A comprehensive review of the oversmoothing in graph neural networks.
\newblock In \emph{CCF Conference on Computer Supported Cooperative Work and Social Computing}, 451--465. Springer.

\bibitem[{Zhang et~al.(2022)Zhang, Liu, Wang, Lu, and Lee}]{zhang2022protgnn}
Zhang, Z.; Liu, Q.; Wang, H.; Lu, C.; and Lee, C. 2022.
\newblock Protgnn: Towards self-explaining graph neural networks.
\newblock In \emph{Proceedings of the AAAI conference on artificial intelligence}, volume~36, 9127--9135.

\bibitem[{Zhang et~al.(2021)Zhang, Liu, Wang, Lu, and Lee}]{zhang2021motif}
Zhang, Z.; Liu, Q.; Wang, H.; Lu, C.; and Lee, C.-K. 2021.
\newblock Motif-based graph self-supervised learning for molecular property prediction.
\newblock \emph{Advances in Neural Information Processing Systems}, 34: 15870--15882.

\bibitem[{Zhou et~al.(2016)Zhou, Khosla, Lapedriza, Oliva, and Torralba}]{zhou2016learning}
Zhou, B.; Khosla, A.; Lapedriza, A.; Oliva, A.; and Torralba, A. 2016.
\newblock Learning deep features for discriminative localization.
\newblock In \emph{Proceedings of the IEEE conference on computer vision and pattern recognition}, 2921--2929.

\end{thebibliography}

\appendix

\section{A. Training details}
\subsection{Hardware and software details}

\subsubsection{Hardware.} All of our experiments were conducted on NVIDIA Grace Hopper GH200 (96 GB memory), NVIDIA Grace CPU 72-Core @ 3.1 GHz, 16GB RAM, CUDA toolkit 12.4. 
\subsubsection{Python environment.}
Our experiments were carried out in Python 3.11, with Pytorch 2.5.1, Pytorch Geometric 2.6.1 for training, and RDKit (2024.9.6) for preprocessing molecules.

\subsection{Experimental details}
We trained the networks with a batch size of 256, using the AdamW optimizer, early stopping after 30 epochs, and a warm-up for the first 50 epochs (10 epochs for tasks that required fewer epochs, like atom-specific tasks from the synthetic dataset). For our model, we used 10-fold cross-validation to select the optimal $\lambda$ using the Wilcoxon signed-rank test. We used MAE and AUROC as target evaluation metrics for hyperparameter searching and the Wilcoxon test. A weight decay of 0.0001 was applied to all models and tasks. Seed was set to 0 during training, while for explanation extraction and evaluation, it was set to 123. All experiment results were obtained using a 5-fold split approach. The B-XAIC benchmark proposed a fixed train-test set, and we followed this recommendation. For the datasets from TDC, we sampled five testing sets using seeds from 0 to 4, following the benchmark recommendation. The ranges of hyperparameters are shown in Table~\ref{tab:hyperparams}.

\begin{table}[h!]
\centering
\caption{Hyperparameter search space used during model optimization.}
\small
\begin{tabular}{ll}
\toprule
\textbf{Hyperparameter} & \textbf{Values} \\
\midrule
Hidden dimensions & [64, 128, 256, 512, 1024] \\
GNN layers   & [2, 3, 4] \\
Learning rate          & [0.001, 0.003, 0.0001, 0.0003] \\
Dropout rate           & [0.0, 0.1, 0.2, 0.3, 0.4, 0.5] \\
$\lambda$ & $[2, 1, 0.5, 10^{-1}, 10^{-2}, 10^{-3}, 10^{-4}, 0]$ \\
\bottomrule
\end{tabular}
\label{tab:hyperparams}
\end{table}

The hyperparameters selected for the synthetic datasets are listed in Table~\ref{tab:hyperparameters_synth}, whereas those for the real-world datasets are presented in Table~\ref{tab:hyperparameters_real}.

\subsection{Data preprocessing}
In our experiments, we standardize target values in our regression task (Solubility), but we do not perform any preprocessing in classification tasks. The atom features used for training include one-hot encoded atom types [C, N, O, F, Cl, Br, P, S, B, I, Other]; we do not use any bond features.

\section{B. Extended results}
The performance of SEAL with different regularization values $\lambda$ for the synthetic benchmark is presented in Table \ref{tab:extended_results_synth}. Detailed results for the subgraph explanation metric are shown in Table \ref{tab:extended_SE_synth}, and for the null explanation metric in Table \ref{tab:extended_NE_synth}. For real-world datasets, the evaluation of classification tasks is presented in Table \ref{tab:extended_results_real_class}, while for the regression task in Table \ref{tab:extended_results_real_regr}. The values of the fidelity metric for these datasets are presented in Table \ref{tab:extended_fidelity_real}.

\begin{table*}[tb]
    \centering
    \setlength{\tabcolsep}{7.8pt} 
    \caption{Hyperparameters found for SEAL, GAT, GCN, and GIN in synthetic dataset evaluation.}
    \label{tab:hyperparameters_synth}
    \begin{tabular}{ p{3.2cm} c c c c c c c  }
        \toprule
        Model & B & P & PAINS & X & indole & rings-count & rings-max \\
        \midrule
        \multicolumn{8}{c}{SEAL} \\
        \midrule
        Hidden dimensions & 1024 & 1024 & 512 & 1024 & 512 & 1024 & 256 \\
        GNN layers & 4 & 4 & 3 & 2 & 4 & 2 & 4 \\
        Learning rate & 0.0001 & 0.003 & 0.003 & 0.003 & 0.0003 & 0.003 & 0.003 \\
        Dropout & 0.4 & 0.1 & 0.1 & 0.1 & 0.1 & 0.1 & 0.2 \\
        $\lambda$ & $2$ & $2$ & $0$ & $2$ & $10^{-4}$ & $10^{-3}$ & $2$ \\
        \midrule
        \multicolumn{8}{c}{GAT} \\
        \midrule
        Hidden dimensions & 256 & 1024 & 256 & 1024 & 256 & 1024 & 256 \\
        GNN layers & 3 & 4 & 3 & 4 & 3 & 4 & 3 \\
        Learning rate & 0.0003 & 0.0001 & 0.0001 & 0.0001 & 0.0001 & 0.003 & 0.0001 \\
        Dropout & 0.4 & 0.4 & 0 & 0.4 & 0 & 0.1 & 0 \\

        \midrule
        \multicolumn{8}{c}{GCN} \\
        \midrule
        Hidden dimensions  & 1024 & 1024 & 512 & 1024 & 512 & 512 & 1024 \\
        GNN layers & 4 & 4 & 4 & 4 & 4 & 4 & 4 \\
        Learning rate & 0.0001 & 0.0001 & 0.0003 & 0.0001 & 0.0003 & 0.0003 & 0.0003 \\
        Dropout & 0.4 & 0.4 & 0.1 & 0.4 & 0.1 & 0.1 & 0.1 \\

        \midrule
        \multicolumn{8}{c}{GIN} \\
        \midrule
        Hidden dimensions & 1024 & 1024 & 1024 & 1024 & 512 & 256 & 1024 \\
        GNN layers & 4 & 4 & 4 & 4 & 4 & 3 & 4 \\
        Learning rate & 0.0001 & 0.0001 & 0.0003 & 0.0001 & 0.0003 & 0.001 & 0.0003 \\
        Dropout & 0.4 & 0.4 & 0.1 & 0.4 & 0.1 & 0.5 & 0.1 \\
        \bottomrule
    \end{tabular}
\end{table*}
\begin{table*}[tb]
    \centering
    \setlength{\tabcolsep}{7.8pt}     
    \caption{Hyperparameters found for SEAL, GAT, GCN, and GIN in real-world dataset evaluation.}
    \label{tab:hyperparameters_real}

    \begin{tabular}{ p{3.2cm} c c c  }
        \toprule
        Model & CYP & hERG & Solubility \\
        \midrule
        \multicolumn{4}{c}{SEAL} \\
        \midrule
        Hidden dimensions & 512 & 512 & 1024 \\
        GNN layers & 4 & 4 & 4 \\
        Learning rate & 0.0003 & 0.0003 & 0.003 \\
        Dropout & 0.1 & 0.1 & 0.1 \\
        $\lambda$  & $2.0$ & $0.0001$ & $0.0001$ \\
        \midrule
        \multicolumn{4}{c}{GAT} \\
        \midrule
        Hidden dimensions & 256 & 256 & 128 \\
        GNN layers & 3 & 3 & 3 \\
        Learning rate & 0.0001 & 0.0001 & 0.0003 \\
        Dropout & 0 & 0 & 0.3 \\
        \midrule
        \multicolumn{4}{c}{GCN} \\
        \midrule
        Hidden dimensions & 256 & 1024 & 1024 \\
        GNN layers & 4 & 4 & 4 \\
        Learning rate & 0.003 & 0.003 & 0.003 \\
        Dropout & 0.2 & 0.1 & 0.1 \\
        \midrule
        \multicolumn{4}{c}{GIN} \\
        \midrule
        Hidden dimensions & 512 & 512 & 1024 \\
        GNN layers & 4 & 4 & 4 \\
        Learning rate & 0.0003 & 0.0003 & 0.0003 \\
        Dropout & 0.1 & 0.1 & 0.1 \\
        \bottomrule
    \end{tabular}
    
\end{table*}

\begin{table*}[tb]
    \centering
    \small
    \setlength{\tabcolsep}{9.4pt} 
    \caption{AUROC, F1, and Accuracy score of various graph neural network architectures on the B-XAIC benchmark.}
    \label{tab:extended_results_synth}
        \begin{tabular}{p{2.4cm} c c c  c c c c}
        \toprule
        Model & rings-count & rings-max & X & P & B & Indole & PAINS \\
        \midrule
        \multicolumn{8}{c}{AUROC $\uparrow$} \\
        \midrule
            GIN  & 1.00 $\pm$ 0.00 & 0.93 $\pm$ 0.02 & 1.00 $\pm$ 0.00 & 1.00 $\pm$ 0.00 & 1.00 $\pm$ 0.00 & 1.00 $\pm$ 0.00 & 0.99 $\pm$ 0.00 \\
            GCN  & 1.00 $\pm$ 0.00 & 0.82 $\pm$ 0.01 & 1.00 $\pm$ 0.00 & 1.00 $\pm$ 0.00 & 1.00 $\pm$ 0.00 & 0.99 $\pm$ 0.00 & 0.97 $\pm$ 0.00 \\
            GAT  & 0.88 $\pm$ 0.01 & 0.75 $\pm$ 0.02 & 1.00 $\pm$ 0.00 & 1.00 $\pm$ 0.00 & 1.00 $\pm$ 0.00 & 0.97 $\pm$ 0.00 & 0.92 $\pm$ 0.01 \\
            SEAL ($\lambda = 2$)  & 0.97 $\pm$ 0.01 & 0.99 $\pm$ 0.01 & 1.00 $\pm$ 0.00 & 1.00 $\pm$ 0.00 & 1.00 $\pm$ 0.00 & 1.00 $\pm$ 0.00 & 0.95 $\pm$ 0.00 \\
            SEAL ($\lambda = 1$)  & 0.97 $\pm$ 0.00 & 0.99 $\pm$ 0.01 & 1.00 $\pm$ 0.00 & 1.00 $\pm$ 0.00 & 1.00 $\pm$ 0.00 & 1.00 $\pm$ 0.00 & 0.96 $\pm$ 0.01 \\
            SEAL ($\lambda = 0.5$)  & 0.97 $\pm$ 0.00 & 0.99 $\pm$ 0.00 & 1.00 $\pm$ 0.00 & 1.00 $\pm$ 0.00 & 1.00 $\pm$ 0.00 & 1.00 $\pm$ 0.00 & 0.96 $\pm$ 0.00 \\
            SEAL ($\lambda = 10^{-1}$)  & 0.98 $\pm$ 0.00 & 0.99 $\pm$ 0.00 & 1.00 $\pm$ 0.00 & 1.00 $\pm$ 0.00 & 1.00 $\pm$ 0.00 & 1.00 $\pm$ 0.00 & 0.96 $\pm$ 0.00 \\
            SEAL ($\lambda = 10^{-2}$)  & 0.98 $\pm$ 0.00 & 0.99 $\pm$ 0.01 & 1.00 $\pm$ 0.00 & 1.00 $\pm$ 0.00 & 1.00 $\pm$ 0.00 & 1.00 $\pm$ 0.00 & 0.96 $\pm$ 0.01 \\
            SEAL ($\lambda = 10^{-3}$)  & 0.98 $\pm$ 0.01 & 0.99 $\pm$ 0.00 & 1.00 $\pm$ 0.00 & 1.00 $\pm$ 0.00 & 1.00 $\pm$ 0.00 & 1.00 $\pm$ 0.00 & 0.99 $\pm$ 0.00 \\
            SEAL ($\lambda = 10^{-4}$)  & 0.99 $\pm$ 0.00 & 0.99 $\pm$ 0.01 & 1.00 $\pm$ 0.00 & 1.00 $\pm$ 0.00 & 1.00 $\pm$ 0.00 & 1.00 $\pm$ 0.00 & 0.99 $\pm$ 0.00 \\
            SEAL ($\lambda = 0$)  & 0.99 $\pm$ 0.00 & 0.98 $\pm$ 0.00 & 1.00 $\pm$ 0.00 & 1.00 $\pm$ 0.00 & 1.00 $\pm$ 0.00 & 1.00 $\pm$ 0.00 & 0.99 $\pm$ 0.00 \\
            \midrule
            \multicolumn{8}{c}{F1 Score $\uparrow$} \\
            \midrule
            GIN  & 1.00 $\pm$ 0.00 & 0.96 $\pm$ 0.00 & 1.00 $\pm$ 0.00 & 1.00 $\pm$ 0.00 & 1.00 $\pm$ 0.00 & 0.99 $\pm$ 0.00 & 0.97 $\pm$ 0.00 \\
            GCN  & 0.98 $\pm$ 0.00 & 0.93 $\pm$ 0.01 & 1.00 $\pm$ 0.00 & 1.00 $\pm$ 0.00 & 1.00 $\pm$ 0.00 & 0.97 $\pm$ 0.00 & 0.93 $\pm$ 0.00 \\
            GAT  & 0.79 $\pm$ 0.03 & 0.92 $\pm$ 0.01 & 1.00 $\pm$ 0.00 & 1.00 $\pm$ 0.00 & 1.00 $\pm$ 0.00 & 0.92 $\pm$ 0.01 & 0.85 $\pm$ 0.01 \\
            SEAL ($\lambda = 2$)  & 0.90 $\pm$ 0.03 & 0.85 $\pm$ 0.05 & 1.00 $\pm$ 0.00 & 1.00 $\pm$ 0.00 & 0.99 $\pm$ 0.01 & 0.98 $\pm$ 0.00 & 0.86 $\pm$ 0.00 \\
            SEAL ($\lambda = 1$)  & 0.86 $\pm$ 0.02 & 0.87 $\pm$ 0.01 & 1.00 $\pm$ 0.00 & 1.00 $\pm$ 0.00 & 0.99 $\pm$ 0.01 & 0.98 $\pm$ 0.00 & 0.86 $\pm$ 0.01 \\
            SEAL ($\lambda = 0.5$)  & 0.87 $\pm$ 0.03 & 0.87 $\pm$ 0.02 & 1.00 $\pm$ 0.00 & 1.00 $\pm$ 0.00 & 0.99 $\pm$ 0.01 & 0.98 $\pm$ 0.00 & 0.86 $\pm$ 0.01 \\
            SEAL ($\lambda = 10^{-1}$)  & 0.88 $\pm$ 0.04 & 0.89 $\pm$ 0.01 & 1.00 $\pm$ 0.00 & 0.99 $\pm$ 0.01 & 0.99 $\pm$ 0.01 & 0.98 $\pm$ 0.00 & 0.86 $\pm$ 0.01 \\
            SEAL ($\lambda = 10^{-2}$)  & 0.92 $\pm$ 0.01 & 0.90 $\pm$ 0.02 & 1.00 $\pm$ 0.00 & 1.00 $\pm$ 0.00 & 0.98 $\pm$ 0.02 & 0.99 $\pm$ 0.00 & 0.86 $\pm$ 0.02 \\
            SEAL ($\lambda = 10^{-3}$)  & 0.93 $\pm$ 0.02 & 0.91 $\pm$ 0.02 & 1.00 $\pm$ 0.00 & 1.00 $\pm$ 0.00 & 0.99 $\pm$ 0.01 & 0.99 $\pm$ 0.00 & 0.93 $\pm$ 0.01 \\
            SEAL ($\lambda = 10^{-4}$)  & 0.94 $\pm$ 0.01 & 0.91 $\pm$ 0.02 & 1.00 $\pm$ 0.00 & 1.00 $\pm$ 0.00 & 0.98 $\pm$ 0.01 & 0.99 $\pm$ 0.00 & 0.95 $\pm$ 0.00 \\
            SEAL ($\lambda = 0$)  & 0.93 $\pm$ 0.01 & 0.88 $\pm$ 0.01 & 1.00 $\pm$ 0.00 & 1.00 $\pm$ 0.00 & 1.00 $\pm$ 0.00 & 0.99 $\pm$ 0.00 & 0.96 $\pm$ 0.01 \\
            \midrule
            \multicolumn{8}{c}{Accuracy $\uparrow$} \\
            \midrule
            GIN  & 1.00 $\pm$ 0.00 & 0.96 $\pm$ 0.00 & 1.00 $\pm$ 0.00 & 1.00 $\pm$ 0.00 & 1.00 $\pm$ 0.00 & 0.99 $\pm$ 0.00 & 0.97 $\pm$ 0.00 \\
            GCN  & 0.98 $\pm$ 0.00 & 0.93 $\pm$ 0.01 & 1.00 $\pm$ 0.00 & 1.00 $\pm$ 0.00 & 1.00 $\pm$ 0.00 & 0.97 $\pm$ 0.00 & 0.93 $\pm$ 0.00 \\
            GAT  & 0.81 $\pm$ 0.02 & 0.91 $\pm$ 0.02 & 1.00 $\pm$ 0.00 & 1.00 $\pm$ 0.00 & 1.00 $\pm$ 0.00 & 0.92 $\pm$ 0.01 & 0.86 $\pm$ 0.01 \\    
            SEAL ($\lambda = 2$)  & 0.93 $\pm$ 0.02 & 0.98 $\pm$ 0.01 & 1.00 $\pm$ 0.00 & 1.00 $\pm$ 0.00 & 1.00 $\pm$ 0.00 & 0.99 $\pm$ 0.00 & 0.91 $\pm$ 0.00 \\
            SEAL ($\lambda = 1$)  & 0.91 $\pm$ 0.02 & 0.98 $\pm$ 0.00 & 1.00 $\pm$ 0.00 & 1.00 $\pm$ 0.00 & 1.00 $\pm$ 0.00 & 0.99 $\pm$ 0.00 & 0.91 $\pm$ 0.01 \\
            SEAL ($\lambda = 0.5$)  & 0.92 $\pm$ 0.02 & 0.98 $\pm$ 0.00 & 1.00 $\pm$ 0.00 & 1.00 $\pm$ 0.00 & 1.00 $\pm$ 0.00 & 0.99 $\pm$ 0.00 & 0.91 $\pm$ 0.01 \\
            SEAL ($\lambda = 10^{-1}$)  & 0.92 $\pm$ 0.03 & 0.99 $\pm$ 0.00 & 1.00 $\pm$ 0.00 & 1.00 $\pm$ 0.00 & 1.00 $\pm$ 0.00 & 0.99 $\pm$ 0.00 & 0.91 $\pm$ 0.00 \\
            SEAL ($\lambda = 10^{-2}$)  & 0.95 $\pm$ 0.01 & 0.99 $\pm$ 0.00 & 1.00 $\pm$ 0.00 & 1.00 $\pm$ 0.00 & 1.00 $\pm$ 0.00 & 0.99 $\pm$ 0.00 & 0.91 $\pm$ 0.01 \\
            SEAL ($\lambda = 10^{-3}$)  & 0.96 $\pm$ 0.01 & 0.99 $\pm$ 0.00 & 1.00 $\pm$ 0.00 & 1.00 $\pm$ 0.00 & 1.00 $\pm$ 0.00 & 0.99 $\pm$ 0.00 & 0.95 $\pm$ 0.00 \\
            SEAL ($\lambda = 10^{-4}$)  & 0.97 $\pm$ 0.00 & 0.99 $\pm$ 0.00 & 1.00 $\pm$ 0.00 & 1.00 $\pm$ 0.00 & 1.00 $\pm$ 0.00 & 0.99 $\pm$ 0.00 & 0.97 $\pm$ 0.00 \\
            SEAL ($\lambda = 0$)  & 0.96 $\pm$ 0.00 & 0.99 $\pm$ 0.00 & 1.00 $\pm$ 0.00 & 1.00 $\pm$ 0.00 & 1.00 $\pm$ 0.00 & 0.99 $\pm$ 0.00 & 0.97 $\pm$ 0.00 \\

            \bottomrule
    \end{tabular}
\end{table*}

\begin{table*}
    \centering
    \small
    \setlength{\tabcolsep}{9pt} 
    \caption{
    Performance of various model explanations on the B-XAIC benchmark. The subgraph explanation (SE) metric is employed for positive examples containing the relevant pattern.
    }
    \label{tab:extended_SE_synth}

    \begin{tabular}{ p{2.4cm} c c c c c c c
    }
        \toprule
        Model & rings-count & rings-max & X & P & B & Indole & PAINS \\
        \midrule
        \multicolumn{8}{c}{SE $\uparrow$} \\
        \midrule
        Deconvolution  & 0.55 $\pm$ 0.24 & 0.38 $\pm$ 0.18 & 0.07 $\pm$ 0.00 & 0.90 $\pm$ 0.00 & 0.72 $\pm$ 0.01 & 0.36 $\pm$ 0.21 & 0.33 $\pm$ 0.01 \\
        GuidedBackprop  & 0.69 $\pm$ 0.05 & 0.67 $\pm$ 0.02 & 0.94 $\pm$ 0.01 & 0.85 $\pm$ 0.11 & 1.00 $\pm$ 0.00 & 0.85 $\pm$ 0.03 & 0.78 $\pm$ 0.02 \\
        IntegratedGradients  & 0.36 $\pm$ 0.00 & 0.34 $\pm$ 0.03 & 1.00 $\pm$ 0.00 & 1.00 $\pm$ 0.00 & 1.00 $\pm$ 0.00 & 0.84 $\pm$ 0.06 & 0.76 $\pm$ 0.02 \\
        Saliency  & 0.51 $\pm$ 0.04 & 0.55 $\pm$ 0.03 & 0.92 $\pm$ 0.02 & 1.00 $\pm$ 0.00 & 1.00 $\pm$ 0.00 & 0.87 $\pm$ 0.02 & 0.81 $\pm$ 0.01 \\
        InputXGradient  & 0.49 $\pm$ 0.03 & 0.47 $\pm$ 0.05 & 1.00 $\pm$ 0.00 & 1.00 $\pm$ 0.00 & 1.00 $\pm$ 0.00 & 0.74 $\pm$ 0.05 & 0.54 $\pm$ 0.03 \\
        GNNExplainer  & 0.49 $\pm$ 0.01 & 0.50 $\pm$ 0.00 & 0.50 $\pm$ 0.00 & 0.51 $\pm$ 0.01 & 0.53 $\pm$ 0.05 & 0.53 $\pm$ 0.03 & 0.54 $\pm$ 0.06 \\
        SEAL ($\lambda = 2$)  & 1.00 $\pm$ 0.00 & 0.34 $\pm$ 0.03 & 1.00 $\pm$ 0.00 & 0.99 $\pm$ 0.00 & 0.88 $\pm$ 0.01 & 0.96 $\pm$ 0.00 & 0.77 $\pm$ 0.00 \\
        SEAL ($\lambda = 1$)  & 1.00 $\pm$ 0.00 & 0.35 $\pm$ 0.01 & 1.00 $\pm$ 0.00 & 0.99 $\pm$ 0.00 & 0.88 $\pm$ 0.01 & 0.96 $\pm$ 0.00 & 0.77 $\pm$ 0.01 \\
        SEAL ($\lambda = 0.5$)  & 1.00 $\pm$ 0.00 & 0.35 $\pm$ 0.02 & 1.00 $\pm$ 0.00 & 0.99 $\pm$ 0.00 & 0.88 $\pm$ 0.01 & 0.96 $\pm$ 0.00 & 0.78 $\pm$ 0.01 \\
        SEAL ($\lambda = 10^{-1}$)  & 1.00 $\pm$ 0.00 & 0.33 $\pm$ 0.02 & 1.00 $\pm$ 0.00 & 0.99 $\pm$ 0.00 & 0.88 $\pm$ 0.01 & 0.96 $\pm$ 0.00 & 0.78 $\pm$ 0.01 \\
        SEAL ($\lambda = 10^{-2}$)  & 1.00 $\pm$ 0.00 & 0.34 $\pm$ 0.02 & 1.00 $\pm$ 0.00 & 0.99 $\pm$ 0.00 & 0.88 $\pm$ 0.01 & 0.96 $\pm$ 0.00 & 0.78 $\pm$ 0.01 \\
        SEAL ($\lambda = 10^{-3}$)  & 0.98 $\pm$ 0.01 & 0.29 $\pm$ 0.02 & 1.00 $\pm$ 0.00 & 0.96 $\pm$ 0.04 & 0.88 $\pm$ 0.01 & 0.96 $\pm$ 0.00 & 0.80 $\pm$ 0.01 \\
        SEAL ($\lambda = 10^{-4}$)  & 0.96 $\pm$ 0.01 & 0.25 $\pm$ 0.02 & 1.00 $\pm$ 0.00 & 0.96 $\pm$ 0.04 & 0.88 $\pm$ 0.01 & 0.96 $\pm$ 0.00 & 0.83 $\pm$ 0.02 \\
        SEAL ($\lambda = 0$)  & 0.87 $\pm$ 0.04 & 0.23 $\pm$ 0.02 & 1.00 $\pm$ 0.00 & 0.91 $\pm$ 0.04 & 0.88 $\pm$ 0.01 & 0.96 $\pm$ 0.00 & 0.83 $\pm$ 0.01 \\
                \bottomrule
    \end{tabular}
\end{table*}

\begin{table*}
    \centering
    \small
    \setlength{\tabcolsep}{9pt} 
    \caption{
    Performance of various model explanations on the B-XAIC benchmark. The null explanation (NE) metric is employed for negative examples, checking uniform distribution.
    }
    \label{tab:extended_NE_synth}

    \begin{tabular}{ p{2.4cm} c c c c c c c
    }
        \toprule
        Model & rings-count & rings-max & X & P & B & Indole & PAINS \\
        \midrule
        \multicolumn{8}{c}{NE $\uparrow$} \\
        \midrule
        Deconvolution  & 0.56 $\pm$ 0.06 & 0.46 $\pm$ 0.07 & 0.84 $\pm$ 0.01 & 0.84 $\pm$ 0.01 & 0.82 $\pm$ 0.01 & 0.80 $\pm$ 0.01 & 0.81 $\pm$ 0.01 \\
        GuidedBackprop  & 0.32 $\pm$ 0.08 & 0.10 $\pm$ 0.06 & 0.35 $\pm$ 0.10 & 0.51 $\pm$ 0.04 & 0.45 $\pm$ 0.03 & 0.33 $\pm$ 0.05 & 0.28 $\pm$ 0.02 \\
        IntegratedGradients  & 0.81 $\pm$ 0.08 & 0.55 $\pm$ 0.20 & 0.19 $\pm$ 0.10 & 0.36 $\pm$ 0.37 & 0.22 $\pm$ 0.07 & 0.31 $\pm$ 0.13 & 0.42 $\pm$ 0.25 \\
        Saliency  & 0.48 $\pm$ 0.05 & 0.36 $\pm$ 0.08 & 0.37 $\pm$ 0.05 & 0.55 $\pm$ 0.03 & 0.50 $\pm$ 0.08 & 0.42 $\pm$ 0.03 & 0.35 $\pm$ 0.04 \\
        InputXGradient  & 0.53 $\pm$ 0.05 & 0.49 $\pm$ 0.14 & 0.23 $\pm$ 0.07 & 0.69 $\pm$ 0.16 & 0.39 $\pm$ 0.14 & 0.49 $\pm$ 0.01 & 0.40 $\pm$ 0.04 \\
        GNNExplainer  & 0.80 $\pm$ 0.07 & 0.80 $\pm$ 0.05 & 0.65 $\pm$ 0.02 & 0.67 $\pm$ 0.01 & 0.67 $\pm$ 0.00 & 0.73 $\pm$ 0.09 & 0.55 $\pm$ 0.26 \\
        SEAL ($\lambda = 2$)  & 0.44 $\pm$ 0.05 & 0.47 $\pm$ 0.09 & 0.32 $\pm$ 0.09 & 0.16 $\pm$ 0.06 & 0.44 $\pm$ 0.02 & 0.64 $\pm$ 0.04 & 0.61 $\pm$ 0.01 \\
        SEAL ($\lambda = 1$)  & 0.42 $\pm$ 0.05 & 0.47 $\pm$ 0.06 & 0.36 $\pm$ 0.09 & 0.18 $\pm$ 0.18 & 0.45 $\pm$ 0.03 & 0.63 $\pm$ 0.03 & 0.62 $\pm$ 0.01 \\
        SEAL ($\lambda = 0.5$)  & 0.45 $\pm$ 0.04 & 0.52 $\pm$ 0.03 & 0.38 $\pm$ 0.08 & 0.21 $\pm$ 0.16 & 0.46 $\pm$ 0.05 & 0.63 $\pm$ 0.04 & 0.63 $\pm$ 0.01 \\
        SEAL ($\lambda = 10^{-1}$)  & 0.53 $\pm$ 0.06 & 0.44 $\pm$ 0.10 & 0.32 $\pm$ 0.05 & 0.22 $\pm$ 0.07 & 0.48 $\pm$ 0.05 & 0.68 $\pm$ 0.03 & 0.63 $\pm$ 0.01 \\
        SEAL ($\lambda = 10^{-2}$)  & 0.48 $\pm$ 0.04 & 0.48 $\pm$ 0.08 & 0.51 $\pm$ 0.05 & 0.09 $\pm$ 0.05 & 0.49 $\pm$ 0.04 & 0.63 $\pm$ 0.02 & 0.63 $\pm$ 0.01 \\
        SEAL ($\lambda = 10^{-3}$)  & 0.71 $\pm$ 0.10 & 0.56 $\pm$ 0.06 & 0.36 $\pm$ 0.07 & 0.10 $\pm$ 0.01 & 0.43 $\pm$ 0.07 & 0.69 $\pm$ 0.05 & 0.65 $\pm$ 0.01 \\
        SEAL ($\lambda = 10^{-4}$)  & 0.70 $\pm$ 0.05 & 0.62 $\pm$ 0.07 & 0.33 $\pm$ 0.09 & 0.10 $\pm$ 0.02 & 0.26 $\pm$ 0.06 & 0.65 $\pm$ 0.05 & 0.70 $\pm$ 0.03 \\
        SEAL ($\lambda = 0$)  & 0.59 $\pm$ 0.06 & 0.62 $\pm$ 0.07 & 0.60 $\pm$ 0.07 & 0.16 $\pm$ 0.08 & 0.38 $\pm$ 0.12 & 0.57 $\pm$ 0.05 & 0.68 $\pm$ 0.01 \\
                        \bottomrule
    \end{tabular}
\end{table*}

\begin{table*}
    \centering
    \small
    \setlength{\tabcolsep}{7.8pt} 
    \caption{Comparison of model performance on real-world datasets (hERG and CYP2C9).}
    \label{tab:extended_results_real_class}

    \begin{tabular}{c l c c c}
        \toprule
        & Model & AUROC $\uparrow$ & F1 $\uparrow$ & Accuracy $\uparrow$ \\
        \midrule
        \parbox[t]{2mm}{\multirow{11}{*}{\rotatebox[origin=c]{90}{hERG}}} 
       & GIN  & 0.86 $\pm$ 0.01 & 0.78 $\pm$ 0.01 & 0.78 $\pm$ 0.01 \\
       & GAT  & 0.70 $\pm$ 0.01 & 0.64 $\pm$ 0.01 & 0.65 $\pm$ 0.01 \\
       & GCN  & 0.81 $\pm$ 0.03 & 0.73 $\pm$ 0.02 & 0.73 $\pm$ 0.02 \\
        & SEAL ($\lambda = 2$)  & 0.81 $\pm$ 0.01 & 0.71 $\pm$ 0.03 & 0.74 $\pm$ 0.01 \\
        & SEAL ($\lambda = 1$)  & 0.81 $\pm$ 0.01 & 0.73 $\pm$ 0.02 & 0.75 $\pm$ 0.01 \\
        & SEAL ($\lambda = 0.5$)  & 0.81 $\pm$ 0.01 & 0.73 $\pm$ 0.02 & 0.75 $\pm$ 0.01 \\
        & SEAL ($\lambda = 10^{-1}$)  & 0.79 $\pm$ 0.01 & 0.70 $\pm$ 0.02 & 0.73 $\pm$ 0.01 \\
        & SEAL ($\lambda = 10^{-2}$)  & 0.80 $\pm$ 0.00 & 0.70 $\pm$ 0.02 & 0.73 $\pm$ 0.01 \\
        & SEAL ($\lambda = 10^{-3}$)  & 0.81 $\pm$ 0.03 & 0.71 $\pm$ 0.03 & 0.74 $\pm$ 0.02 \\
        & SEAL ($\lambda = 10^{-4}$)  & 0.85 $\pm$ 0.01 & 0.76 $\pm$ 0.01 & 0.77 $\pm$ 0.00 \\
        & SEAL ($\lambda = 0$)  & 0.85 $\pm$ 0.01 & 0.76 $\pm$ 0.01 & 0.78 $\pm$ 0.01 \\
               \midrule
        \parbox[t]{2mm}{\multirow{11}{*}{\rotatebox[origin=c]{90}{CYP2C9}}}
        & GIN  & 0.86 $\pm$ 0.01 & 0.79 $\pm$ 0.01 & 0.79 $\pm$ 0.01 \\
        & GAT  & 0.68 $\pm$ 0.01 & 0.67 $\pm$ 0.01 & 0.69 $\pm$ 0.01 \\
        & GCN  & 0.84 $\pm$ 0.01 & 0.78 $\pm$ 0.01 & 0.78 $\pm$ 0.01 \\
        & SEAL ($\lambda = 2$)  & 0.81 $\pm$ 0.01 & 0.65 $\pm$ 0.02 & 0.78 $\pm$ 0.01 \\
        & SEAL ($\lambda = 1$)  & 0.81 $\pm$ 0.01 & 0.64 $\pm$ 0.03 & 0.78 $\pm$ 0.01 \\
        & SEAL ($\lambda = 0.5$)  & 0.81 $\pm$ 0.00 & 0.64 $\pm$ 0.02 & 0.78 $\pm$ 0.01 \\
        & SEAL ($\lambda = 10^{-1}$)  & 0.79 $\pm$ 0.01 & 0.59 $\pm$ 0.03 & 0.76 $\pm$ 0.01 \\
        & SEAL ($\lambda = 10^{-2}$)  & 0.79 $\pm$ 0.01 & 0.58 $\pm$ 0.03 & 0.76 $\pm$ 0.01 \\
        & SEAL ($\lambda = 10^{-3}$)  & 0.83 $\pm$ 0.01 & 0.64 $\pm$ 0.03 & 0.78 $\pm$ 0.01 \\
        & SEAL ($\lambda = 10^{-4}$)  & 0.83 $\pm$ 0.00 & 0.64 $\pm$ 0.04 & 0.78 $\pm$ 0.01 \\
        & SEAL ($\lambda = 0$)  & 0.83 $\pm$ 0.01 & 0.64 $\pm$ 0.03 & 0.79 $\pm$ 0.01 \\
                \bottomrule
    \end{tabular}
\end{table*}

\begin{table*}
    \centering
    \small
    \setlength{\tabcolsep}{7.8pt} 
    \caption{Comparison of model performance on real-world datasets (Solubility).}
    \label{tab:extended_results_real_regr}

    \begin{tabular}{c l c c}
        \toprule
        & Model & MAE $\downarrow$ & RMSE $\downarrow$ \\
        \midrule
        \parbox[t]{2mm}{\multirow{11}{*}{\rotatebox[origin=c]{90}{Solubility}}} 
        & GIN  & 0.41 $\pm$ 0.01 & 0.60 $\pm$ 0.02 \\
        & GAT  & 0.57 $\pm$ 0.01 & 0.75 $\pm$ 0.03 \\
        & GCN  & 0.49 $\pm$ 0.02 & 0.67 $\pm$ 0.03 \\
        & SEAL ($\lambda = 2$)  & 0.54 $\pm$ 0.04 & 0.73 $\pm$ 0.05 \\
        & SEAL ($\lambda = 1$)  & 0.53 $\pm$ 0.05 & 0.73 $\pm$ 0.05 \\
        & SEAL ($\lambda = 0.5$)  & 0.54 $\pm$ 0.05 & 0.73 $\pm$ 0.05 \\
        & SEAL ($\lambda = 10^{-1}$)  & 0.54 $\pm$ 0.05 & 0.73 $\pm$ 0.06 \\
        & SEAL ($\lambda = 10^{-2}$)  & 0.53 $\pm$ 0.05 & 0.73 $\pm$ 0.04 \\
        & SEAL ($\lambda = 10^{-3}$)  & 0.48 $\pm$ 0.01 & 0.68 $\pm$ 0.04 \\
        & SEAL ($\lambda = 10^{-4}$)  & 0.47 $\pm$ 0.01 & 0.66 $\pm$ 0.04 \\
        & SEAL ($\lambda = 0$)  & 0.45 $\pm$ 0.01 & 0.66 $\pm$ 0.03 \\
        
            \bottomrule
    \end{tabular}
\end{table*}

\begin{table*}
    \centering
    \small
    \setlength{\tabcolsep}{7.8pt} 
        \caption{Performance of model explanations on real-world datasets (hERG, CYP2C9 and Solubility). Explanations are evaluated using Fidelity metrics at 10\%, 20\%, and 30\% masking thresholds, representing the proportion of most important atoms (nodes) either removed or retained during the evaluation.}
    \label{tab:extended_fidelity_real}

    \begin{tabular}{ c l c  c c c c c
    }
        \toprule
        & Model & $\text{Fidelity}_{10}+ \uparrow$ & $\text{Fidelity}_{10}- \downarrow$ & $\text{Fidelity}_{20}+ \uparrow$ & $\text{Fidelity}_{20}- \downarrow$ & $\text{Fidelity}_{30}+ \uparrow$ & $\text{Fidelity}_{30}- \downarrow$ \\
        \midrule
        \parbox[t]{2mm}{\multirow{15}{*}{\rotatebox[origin=c]{90}{hERG}}} 
       & Deconvolution  & 0.33 $\pm$ 0.06 & 0.49 $\pm$ 0.02 & 0.45 $\pm$ 0.04 & 0.49 $\pm$ 0.02 & 0.48 $\pm$ 0.02 & 0.49 $\pm$ 0.02 \\
       & GuidedBackprop  & 0.39 $\pm$ 0.03 & 0.49 $\pm$ 0.02 & 0.44 $\pm$ 0.04 & 0.49 $\pm$ 0.02 & 0.47 $\pm$ 0.02 & 0.49 $\pm$ 0.02 \\
       & IntegratedGradients  & 0.47 $\pm$ 0.07 & 0.49 $\pm$ 0.02 & 0.54 $\pm$ 0.10 & 0.47 $\pm$ 0.03 & 0.58 $\pm$ 0.19 & 0.41 $\pm$ 0.16 \\
      &  Saliency  & 0.35 $\pm$ 0.03 & 0.49 $\pm$ 0.02 & 0.42 $\pm$ 0.03 & 0.49 $\pm$ 0.02 & 0.46 $\pm$ 0.02 & 0.49 $\pm$ 0.02 \\
      &  InputXGradient  & 0.33 $\pm$ 0.05 & 0.49 $\pm$ 0.02 & 0.40 $\pm$ 0.04 & 0.49 $\pm$ 0.02 & 0.44 $\pm$ 0.04 & 0.50 $\pm$ 0.02 \\
       & GNNExplainer  & 0.42 $\pm$ 0.15 & 0.49 $\pm$ 0.02 & 0.46 $\pm$ 0.03 & 0.49 $\pm$ 0.02 & 0.50 $\pm$ 0.05 & 0.47 $\pm$ 0.04 \\        
        & SEAL ($\lambda = 2$)  & 0.57 $\pm$ 0.01 & 0.00 $\pm$ 0.00 & 0.66 $\pm$ 0.01 & 0.00 $\pm$ 0.00 & 0.76 $\pm$ 0.01 & 0.00 $\pm$ 0.00 \\
        & SEAL ($\lambda = 1$)  & 0.57 $\pm$ 0.02 & 0.00 $\pm$ 0.00 & 0.65 $\pm$ 0.02 & 0.00 $\pm$ 0.00 & 0.75 $\pm$ 0.01 & 0.00 $\pm$ 0.00 \\
        & SEAL ($\lambda = 0.5$)  & 0.57 $\pm$ 0.01 & 0.00 $\pm$ 0.00 & 0.66 $\pm$ 0.02 & 0.00 $\pm$ 0.00 & 0.76 $\pm$ 0.01 & 0.00 $\pm$ 0.00 \\
        & SEAL ($\lambda = 10^{-1}$)  & 0.59 $\pm$ 0.01 & 0.00 $\pm$ 0.00 & 0.68 $\pm$ 0.01 & 0.00 $\pm$ 0.00 & 0.77 $\pm$ 0.01 & 0.00 $\pm$ 0.00 \\
        & SEAL ($\lambda = 10^{-2}$)  & 0.59 $\pm$ 0.01 & 0.00 $\pm$ 0.00 & 0.68 $\pm$ 0.01 & 0.00 $\pm$ 0.00 & 0.78 $\pm$ 0.01 & 0.00 $\pm$ 0.00 \\
        & SEAL ($\lambda = 10^{-3}$)  & 0.63 $\pm$ 0.03 & 0.02 $\pm$ 0.01 & 0.72 $\pm$ 0.03 & 0.01 $\pm$ 0.01 & 0.80 $\pm$ 0.03 & 0.01 $\pm$ 0.01 \\
        & SEAL ($\lambda = 10^{-4}$)  & 0.63 $\pm$ 0.01 & 0.09 $\pm$ 0.03 & 0.71 $\pm$ 0.01 & 0.07 $\pm$ 0.02 & 0.78 $\pm$ 0.01 & 0.05 $\pm$ 0.02 \\
        & SEAL ($\lambda = 0$)  & 0.67 $\pm$ 0.02 & 0.15 $\pm$ 0.02 & 0.74 $\pm$ 0.02 & 0.15 $\pm$ 0.01 & 0.78 $\pm$ 0.03 & 0.14 $\pm$ 0.01 \\
        
        \midrule
        \parbox[t]{2mm}{\multirow{15}{*}{\rotatebox[origin=c]{90}{CYP2C9}}} 
        & Deconvolution  & 0.24 $\pm$ 0.02 & 0.40 $\pm$ 0.09 & 0.31 $\pm$ 0.03 & 0.37 $\pm$ 0.03 & 0.34 $\pm$ 0.03 & 0.37 $\pm$ 0.03 \\
        & GuidedBackprop  & 0.33 $\pm$ 0.03 & 0.40 $\pm$ 0.10 & 0.34 $\pm$ 0.04 & 0.39 $\pm$ 0.07 & 0.35 $\pm$ 0.03 & 0.38 $\pm$ 0.05 \\
        & IntegratedGradients  & 0.45 $\pm$ 0.08 & 0.29 $\pm$ 0.16 & 0.59 $\pm$ 0.14 & 0.29 $\pm$ 0.16 & 0.68 $\pm$ 0.21 & 0.29 $\pm$ 0.16 \\
        & Saliency  & 0.32 $\pm$ 0.04 & 0.38 $\pm$ 0.06 & 0.37 $\pm$ 0.03 & 0.36 $\pm$ 0.03 & 0.38 $\pm$ 0.04 & 0.36 $\pm$ 0.03 \\
        & InputXGradient  & 0.31 $\pm$ 0.02 & 0.41 $\pm$ 0.12 & 0.34 $\pm$ 0.02 & 0.41 $\pm$ 0.11 & 0.35 $\pm$ 0.03 & 0.41 $\pm$ 0.11 \\
        & GNNExplainer  & 0.30 $\pm$ 0.02 & 0.37 $\pm$ 0.03 & 0.43 $\pm$ 0.10 & 0.43 $\pm$ 0.16 & 0.42 $\pm$ 0.08 & 0.33 $\pm$ 0.08 \\
        & SEAL ($\lambda = 2$)  & 0.52 $\pm$ 0.02 & 0.04 $\pm$ 0.05 & 0.57 $\pm$ 0.03 & 0.03 $\pm$ 0.04 & 0.66 $\pm$ 0.03 & 0.01 $\pm$ 0.01 \\
        & SEAL ($\lambda = 1$)  & 0.50 $\pm$ 0.02 & 0.05 $\pm$ 0.06 & 0.55 $\pm$ 0.02 & 0.04 $\pm$ 0.05 & 0.64 $\pm$ 0.03 & 0.01 $\pm$ 0.01 \\
        & SEAL ($\lambda = 0.5$)  & 0.50 $\pm$ 0.02 & 0.05 $\pm$ 0.04 & 0.56 $\pm$ 0.01 & 0.04 $\pm$ 0.03 & 0.64 $\pm$ 0.02 & 0.01 $\pm$ 0.01 \\
        & SEAL ($\lambda = 10^{-1}$)  & 0.57 $\pm$ 0.02 & 0.00 $\pm$ 0.00 & 0.63 $\pm$ 0.02 & 0.00 $\pm$ 0.00 & 0.72 $\pm$ 0.02 & 0.00 $\pm$ 0.00 \\
        & SEAL ($\lambda = 10^{-2}$)  & 0.53 $\pm$ 0.02 & 0.00 $\pm$ 0.00 & 0.60 $\pm$ 0.02 & 0.00 $\pm$ 0.00 & 0.69 $\pm$ 0.02 & 0.00 $\pm$ 0.00 \\
        & SEAL ($\lambda = 10^{-3}$)  & 0.53 $\pm$ 0.02 & 0.06 $\pm$ 0.04 & 0.58 $\pm$ 0.02 & 0.05 $\pm$ 0.04 & 0.65 $\pm$ 0.02 & 0.04 $\pm$ 0.03 \\
        & SEAL ($\lambda = 10^{-4}$)  & 0.52 $\pm$ 0.02 & 0.13 $\pm$ 0.05 & 0.58 $\pm$ 0.02 & 0.11 $\pm$ 0.03 & 0.63 $\pm$ 0.03 & 0.10 $\pm$ 0.04 \\
        & SEAL ($\lambda = 0$)  & 0.52 $\pm$ 0.00 & 0.11 $\pm$ 0.02 & 0.59 $\pm$ 0.01 & 0.09 $\pm$ 0.01 & 0.65 $\pm$ 0.01 & 0.07 $\pm$ 0.01 \\
                \midrule
        \parbox[t]{2mm}{\multirow{15}{*}{\rotatebox[origin=c]{90}{Solubility}}} 

       & Deconvolution  & 0.41 $\pm$ 0.08 & 5.24 $\pm$ 1.60 & 0.95 $\pm$ 0.23 & 5.02 $\pm$ 1.52 & 1.54 $\pm$ 0.42 & 4.78 $\pm$ 1.43 \\
       & GuidedBackprop  & 1.11 $\pm$ 0.17 & 5.08 $\pm$ 1.67 & 2.59 $\pm$ 0.54 & 4.70 $\pm$ 1.65 & 3.37 $\pm$ 0.94 & 4.31 $\pm$ 1.56 \\
       & IntegratedGradients  & 0.42 $\pm$ 0.05 & 5.47 $\pm$ 1.84 & 0.84 $\pm$ 0.20 & 5.45 $\pm$ 1.96 & 1.30 $\pm$ 0.42 & 5.33 $\pm$ 1.98 \\
       & Saliency  & 0.95 $\pm$ 0.15 & 5.18 $\pm$ 1.71 & 1.90 $\pm$ 0.46 & 4.94 $\pm$ 1.74 & 2.56 $\pm$ 0.74 & 4.64 $\pm$ 1.66 \\
        &  InputXGradient  & 0.73 $\pm$ 0.11 & 5.13 $\pm$ 1.62 & 1.71 $\pm$ 0.36 & 4.81 $\pm$ 1.56 & 2.44 $\pm$ 0.61 & 4.45 $\pm$ 1.45 \\
       & GNNExplainer  & 0.71 $\pm$ 0.13 & 5.12 $\pm$ 1.60 & 1.81 $\pm$ 0.49 & 4.80 $\pm$ 1.53 & 2.82 $\pm$ 0.86 & 4.47 $\pm$ 1.43 \\
        & SEAL ($\lambda = 2$)  & 1.12 $\pm$ 0.25 & 1.04 $\pm$ 0.42 & 1.20 $\pm$ 0.28 & 0.96 $\pm$ 0.38 & 1.34 $\pm$ 0.32 & 0.84 $\pm$ 0.33 \\
        & SEAL ($\lambda = 1$)  & 1.04 $\pm$ 0.31 & 0.83 $\pm$ 0.39 & 1.11 $\pm$ 0.34 & 0.78 $\pm$ 0.36 & 1.22 $\pm$ 0.38 & 0.69 $\pm$ 0.31 \\
        & SEAL ($\lambda = 0.5$)  & 1.08 $\pm$ 0.31 & 0.89 $\pm$ 0.41 & 1.15 $\pm$ 0.33 & 0.83 $\pm$ 0.38 & 1.26 $\pm$ 0.37 & 0.74 $\pm$ 0.34 \\
        & SEAL ($\lambda = 10^{-1}$)  & 0.77 $\pm$ 0.20 & 0.67 $\pm$ 0.29 & 0.83 $\pm$ 0.23 & 0.62 $\pm$ 0.26 & 0.92 $\pm$ 0.27 & 0.55 $\pm$ 0.23 \\
        & SEAL ($\lambda = 10^{-2}$)  & 0.64 $\pm$ 0.12 & 0.50 $\pm$ 0.13 & 0.68 $\pm$ 0.13 & 0.49 $\pm$ 0.13 & 0.73 $\pm$ 0.14 & 0.46 $\pm$ 0.13 \\
        & SEAL ($\lambda = 10^{-3}$)  & 1.24 $\pm$ 0.20 & 1.26 $\pm$ 0.33 & 1.37 $\pm$ 0.21 & 1.14 $\pm$ 0.30 & 1.55 $\pm$ 0.25 & 0.98 $\pm$ 0.25 \\
        & SEAL ($\lambda = 10^{-4}$)  & 0.78 $\pm$ 0.14 & 0.64 $\pm$ 0.16 & 0.84 $\pm$ 0.15 & 0.58 $\pm$ 0.13 & 0.91 $\pm$ 0.17 & 0.52 $\pm$ 0.09 \\
        & SEAL ($\lambda = 0$)  & 0.49 $\pm$ 0.03 & 0.55 $\pm$ 0.07 & 0.54 $\pm$ 0.04 & 0.53 $\pm$ 0.08 & 0.58 $\pm$ 0.05 & 0.48 $\pm$ 0.07 \\
                \bottomrule
    \end{tabular}
\end{table*}

\section{C. Ablation study on masking strategy}

In our fidelity evaluation, we analyze how masking different types of contributions affects the model’s interpretability. For each fidelity type (positive, negative), we evaluate the impact of masking the top $10\%$, $20\%$, and $30\%$ of nodes or contributions. This allows us to compare how well explanations identify the most influential substructures without exceeding a predefined threshold.

Unlike standard explainers that only operate on node masks, our model allows for masking specific contribution scores directly at the level of the model’s architecture by setting $c_i=0$ for a given fragment. However, a challenge with this approach is that sometimes, even at the beginning of the ranking, a single large important substructure can surpass the 10\% node threshold. Consequently, the masking process might include more atoms than planned, violating the target percentage and possibly impacting the fidelity assessment. Also, we need to carefully select the masking strategy: whether to focus on absolute contributions or to selectively mask only positive or negative influences. However, the optimal strategy may vary depending on the task and model sensitivity, whether one chooses to use or omit masking of contributions, and whether masking is guided by absolute, positive only, or negative only importance scores. The results comparing these masking strategies are reported in Table \ref{tab:masking_herg} for hERG, in Table \ref{tab:masking_cyp} for CYP2C9, and in Table \ref{tab:masking_sol} for Solubility. These results contain the following naming convention: 
\begin{itemize}
    \item mask-abs: zeroing features, mask contributions - based on maximum absolute value,
    \item mask: zeroing features, mask contributions - based on maximum or minimum value,
    \item abs: zeroing features - based on maximum absolute value,
    \item zero: zeroing features - based on maximum or minimum value.
\end{itemize}

\section{D. User study}

\begin{figure*}[t!]
    \centering
    \includegraphics[width=\linewidth]{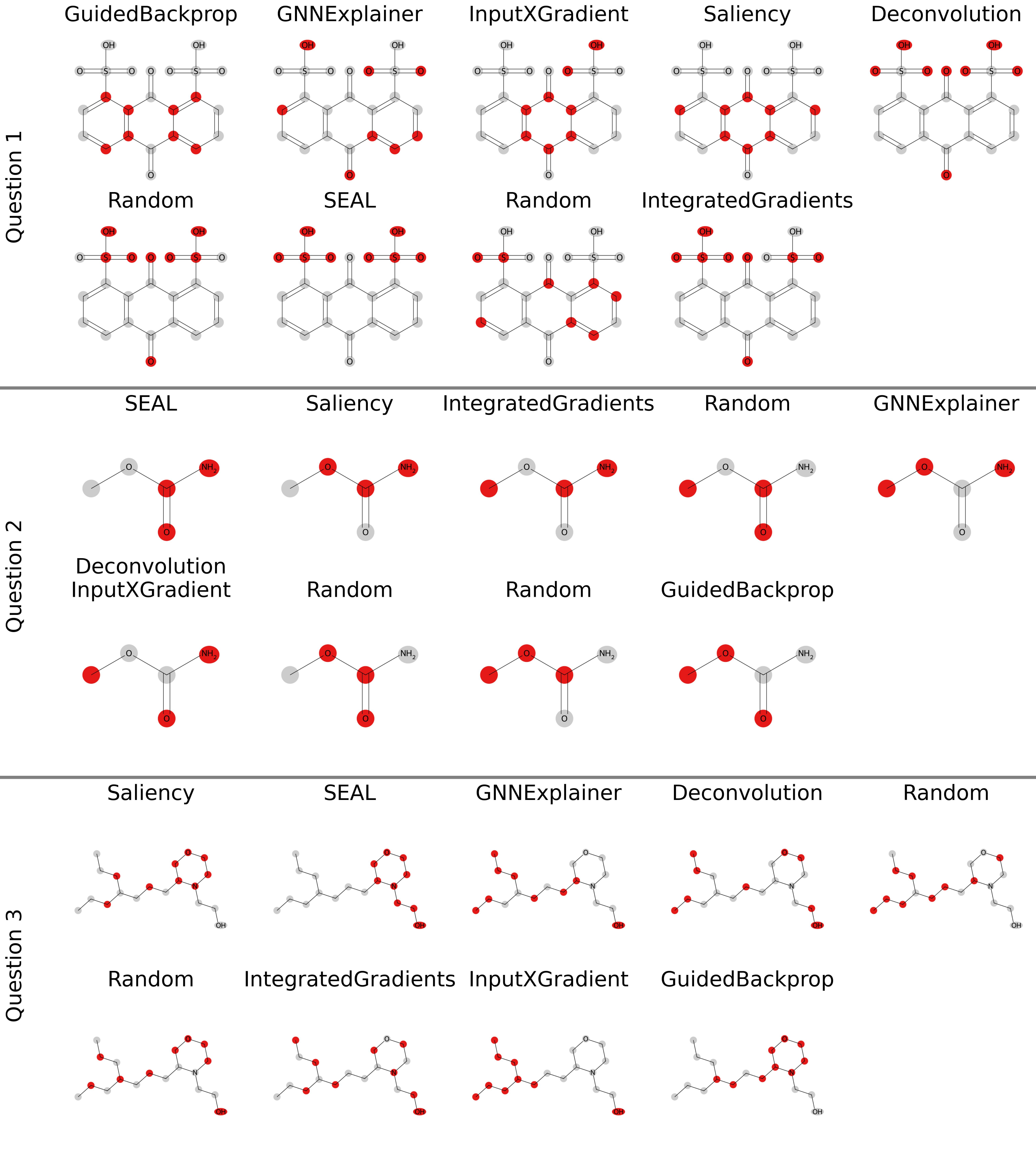}
    \caption{Node-level explanation examples from user study, The red color indicates that the highlighted atoms had a positive contribution to the compound’s solubility. }
    \label{fig:us1}
\end{figure*}

\begin{figure*}
    \centering
    \includegraphics[width=\linewidth]{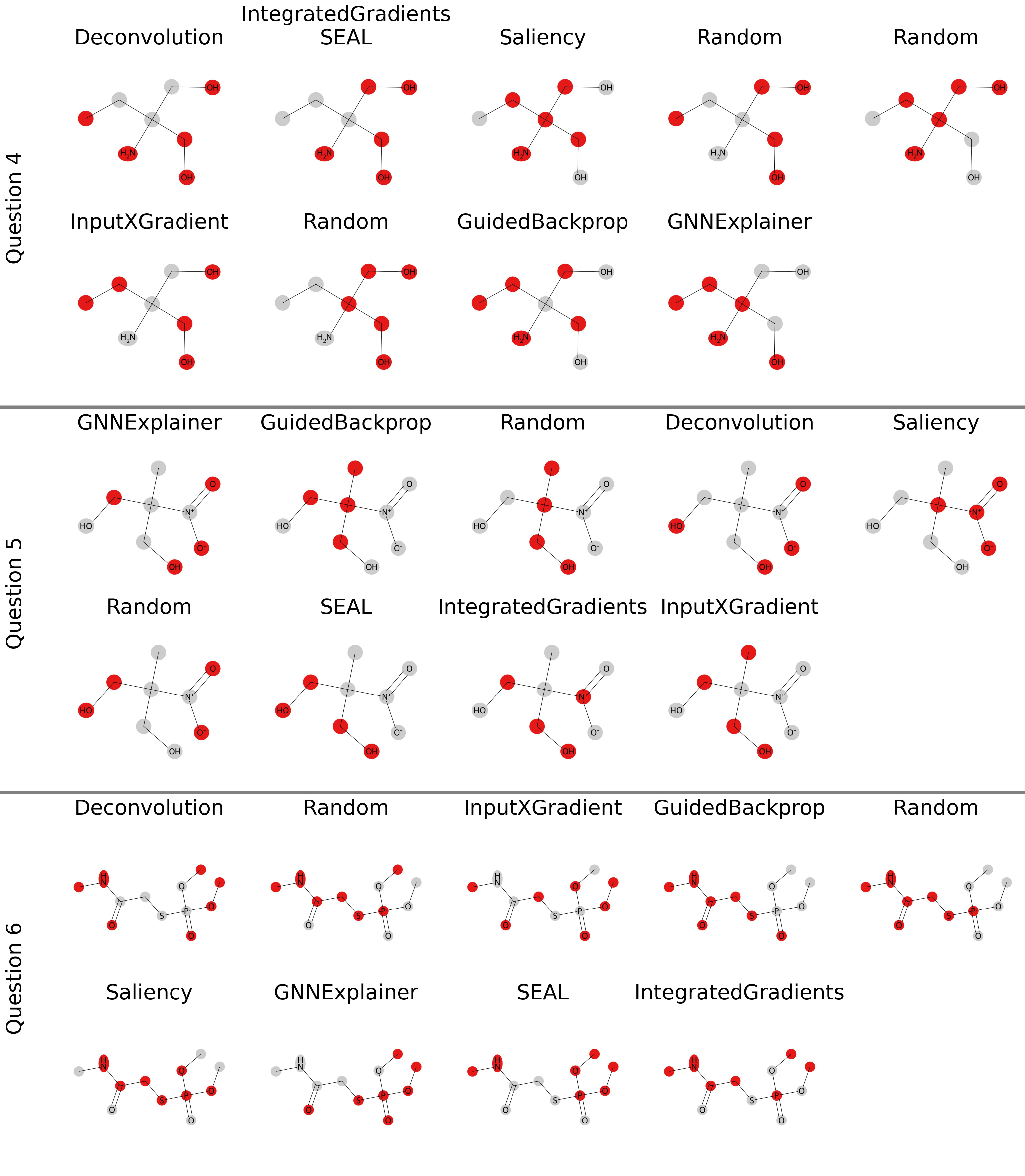}
    \caption{Node-level explanation examples from user study, The red color indicates that the highlighted atoms had a positive contribution to the compound’s solubility. }
    \label{fig:us2}
\end{figure*}

\begin{figure*}
    \centering
    \includegraphics[width=\linewidth]{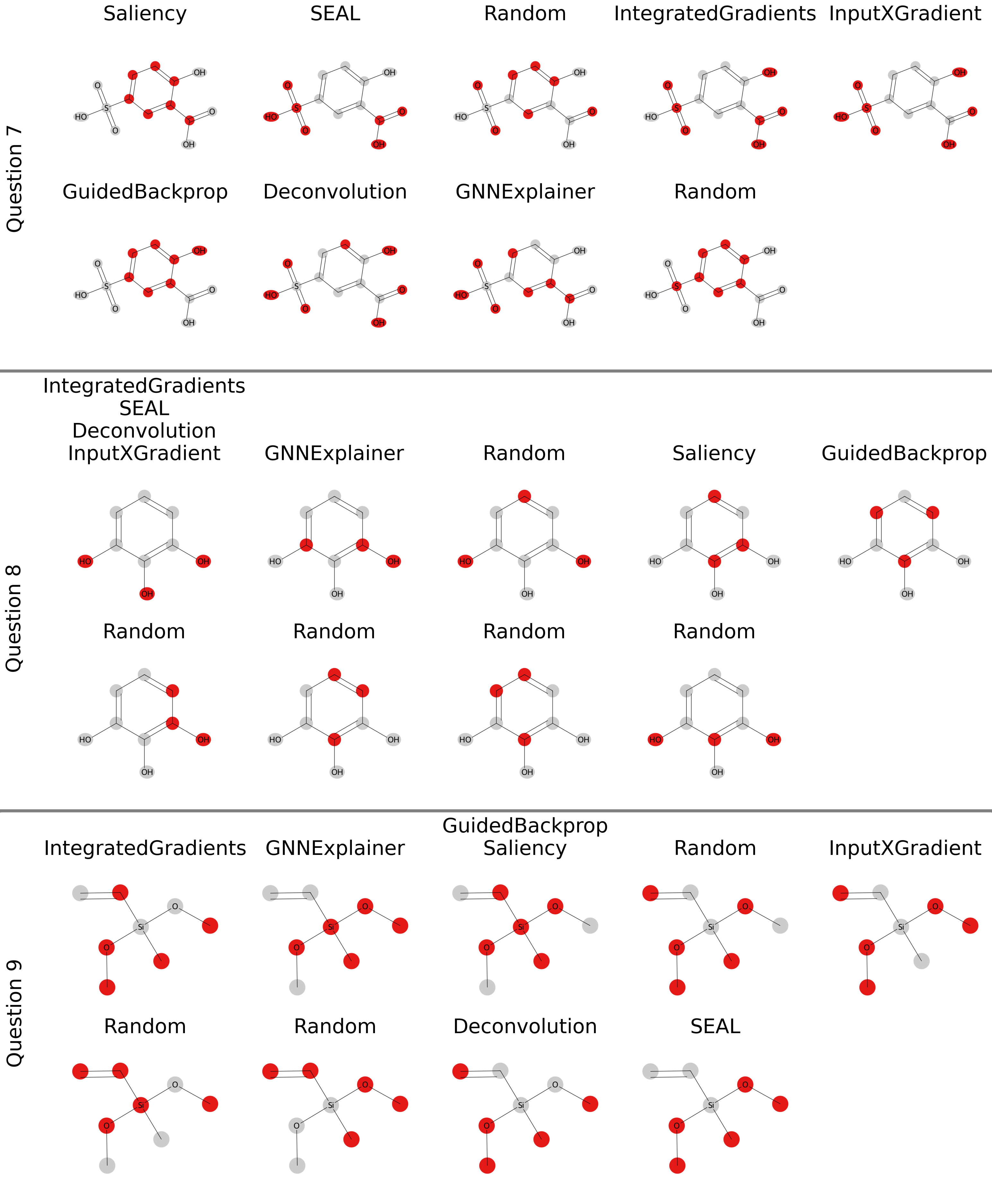}
    \caption{Node-level explanation examples from user study, The red color indicates that the highlighted atoms had a positive contribution to the compound’s solubility. }
    \label{fig:us3}
\end{figure*}

\begin{figure*}
    \centering
    \includegraphics[width=\linewidth]{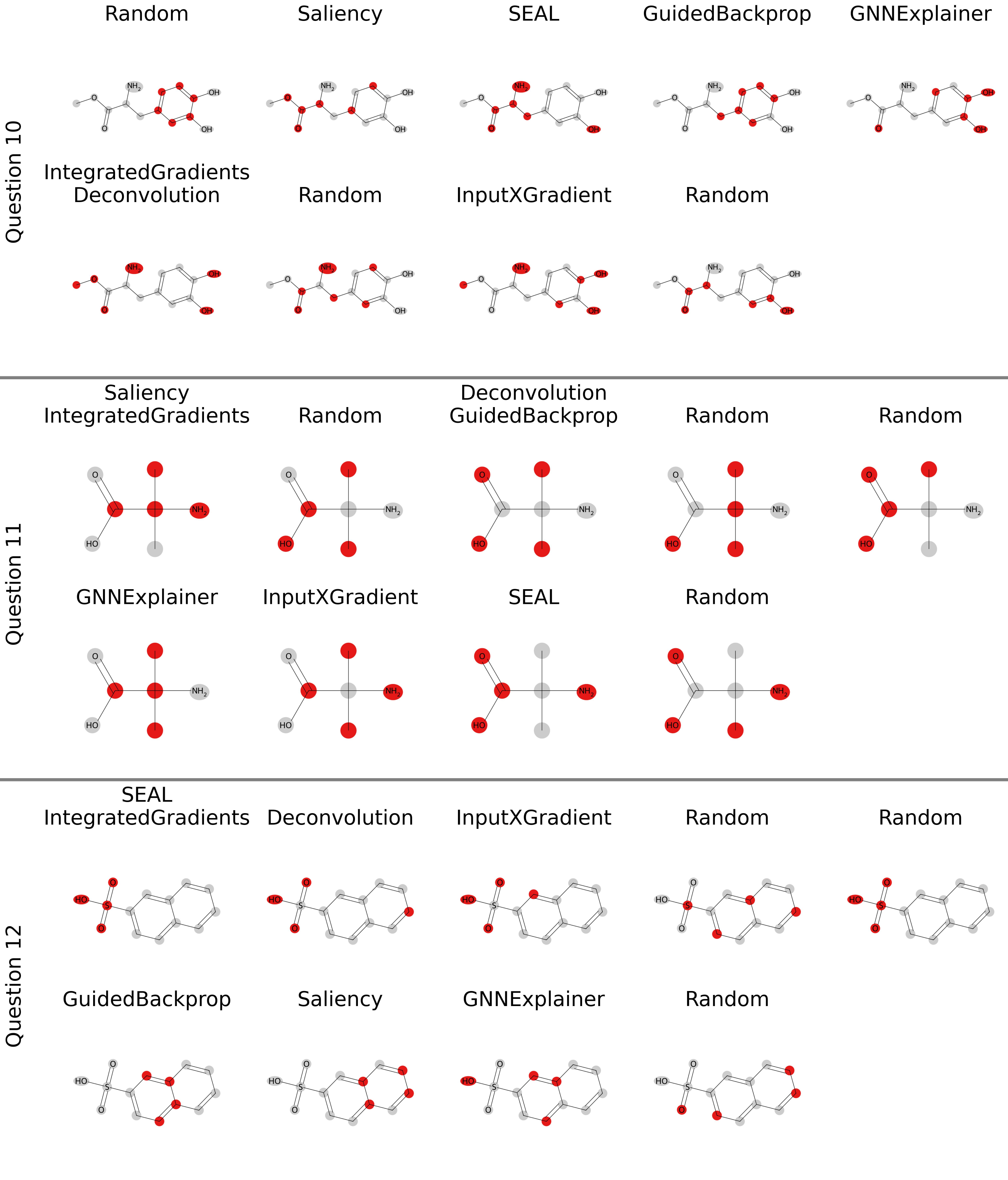}
    \caption{Node-level explanation examples from user study, The red color indicates that the highlighted atoms had a positive contribution to the compound’s solubility. }
    \label{fig:us4}
\end{figure*}

\begin{figure*}
    \centering
    \includegraphics[width=\linewidth]{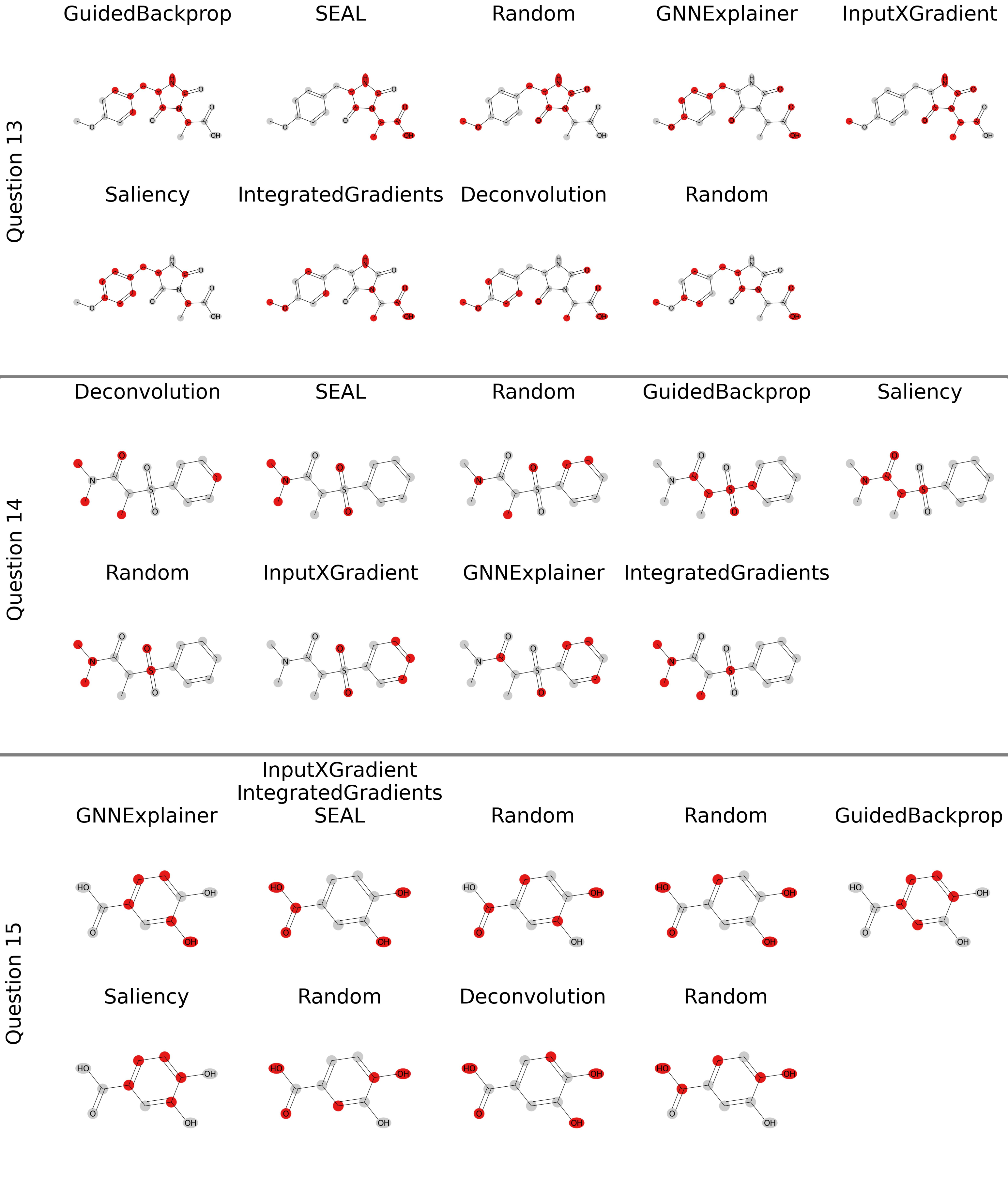}
    \caption{Node-level explanation examples from user study, The red color indicates that the highlighted atoms had a positive contribution to the compound’s solubility. }
    \label{fig:us5}
\end{figure*}

\begin{figure*}
    \centering
    \includegraphics[width=\linewidth]{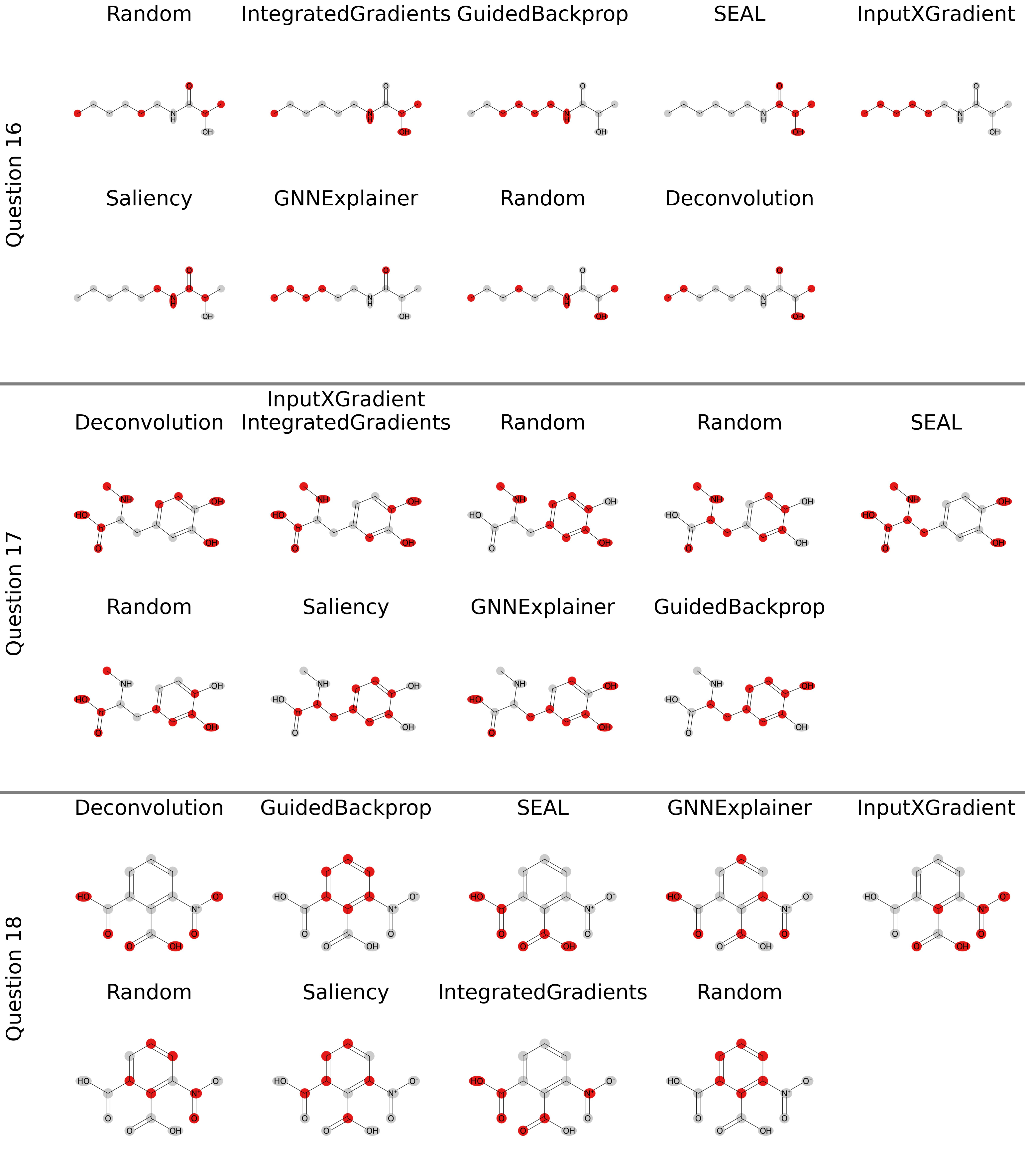}
    \caption{Node-level explanation examples from user study, The red color indicates that the highlighted atoms had a positive contribution to the compound’s solubility. }
    \label{fig:us6}
\end{figure*}

\begin{figure*}
    \centering
    \includegraphics[width=\linewidth]{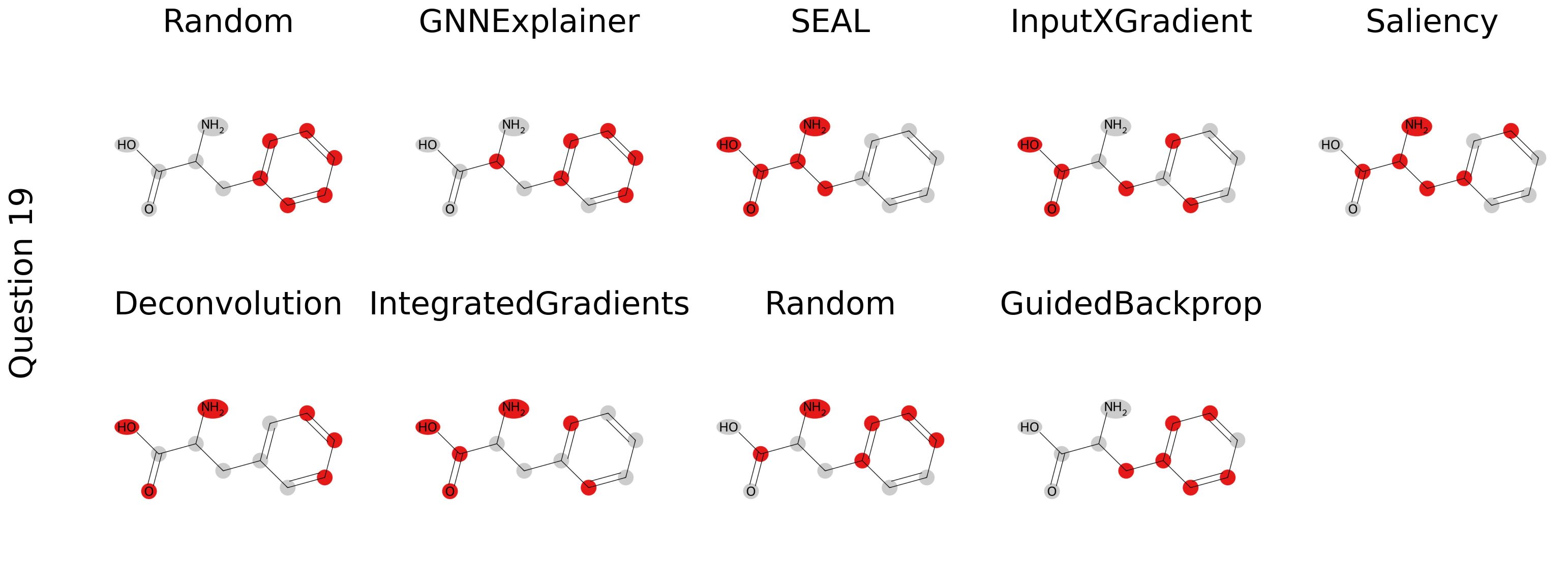}
    \caption{Node-level explanation examples from user study, The red color indicates that the highlighted atoms had a positive contribution to the compound’s solubility. }
    \label{fig:us7}
\end{figure*}

All of the molecules that were included in our user study are presented in Figures \ref{fig:us1}-\ref{fig:us7}. Each explanation is annotated with the name of the method that produced this explanation (the names were not included in the survey, and the order of the explanations was randomized). Some methods resulted in the same explanation, which is why some of the figures have multiple method names. In these situations, we had to generate more random explanations to maintain a consistent number of options across questions. Methods that took part in these experiments: SEAL, GuidedBackprop, GNNExplainer, InputXGradient, Saliency, Deconvolution, IntegratedGradients, two Random methods, the first where we sample from nodes, the second where we sample for substructures generated by BRICS.

\begin{table*}
    \centering
    \small
    \setlength{\tabcolsep}{4.8pt} 
        \caption{Model explanations performance using different type of masking strategy in SEAL architecture for hERG dataset. Evaluating using Fidelity metrics at 10\%, 20\%, and 30\% masking thresholds.}
    \label{tab:masking_herg}

    \begin{tabular}{ c l c  c c c c c
    }
        \toprule
        & Model & $\text{Fidelity}_{10}+ \uparrow$ & $\text{Fidelity}_{10}- \downarrow$ & $\text{Fidelity}_{20}+ \uparrow$ & $\text{Fidelity}_{20}- \downarrow$ & $\text{Fidelity}_{30}+ \uparrow$ & $\text{Fidelity}_{30}- \downarrow$ \\
        \midrule
        \multicolumn{8}{c}{hERG} \\

        \midrule
    \parbox[t]{2mm}{\multirow{4}{*}{\rotatebox[origin=c]{90}{$\lambda = 2$}}} 

& SEAL-mask-abs   & 0.36 $\pm$ 0.01 & 0.18 $\pm$ 0.00 & 0.37 $\pm$ 0.02 & 0.18 $\pm$ 0.01 & 0.38 $\pm$ 0.02 & 0.15 $\pm$ 0.01 \\
& SEAL-mask  & 0.57 $\pm$ 0.01 & 0.00 $\pm$ 0.00 & 0.66 $\pm$ 0.01 & 0.00 $\pm$ 0.00 & 0.76 $\pm$ 0.01 & 0.00 $\pm$ 0.00 \\
& SEAL-abs  & 0.49 $\pm$ 0.04 & 0.46 $\pm$ 0.04 & 0.49 $\pm$ 0.04 & 0.46 $\pm$ 0.04 & 0.49 $\pm$ 0.04 & 0.44 $\pm$ 0.05 \\
& SEAL-zero  & 0.59 $\pm$ 0.05 & 0.45 $\pm$ 0.04 & 0.58 $\pm$ 0.10 & 0.43 $\pm$ 0.06 & 0.58 $\pm$ 0.12 & 0.40 $\pm$ 0.09 \\
        \midrule

    \parbox[t]{2mm}{\multirow{4}{*}{\rotatebox[origin=c]{90}{$\lambda = 1$}}} 

& SEAL-mask-abs  & 0.36 $\pm$ 0.01 & 0.18 $\pm$ 0.01 & 0.37 $\pm$ 0.02 & 0.18 $\pm$ 0.01 & 0.39 $\pm$ 0.02 & 0.15 $\pm$ 0.01 \\
& SEAL-mask  & 0.57 $\pm$ 0.02 & 0.00 $\pm$ 0.00 & 0.65 $\pm$ 0.02 & 0.00 $\pm$ 0.00 & 0.75 $\pm$ 0.01 & 0.00 $\pm$ 0.00 \\
& SEAL-abs  & 0.54 $\pm$ 0.04 & 0.48 $\pm$ 0.10 & 0.55 $\pm$ 0.03 & 0.47 $\pm$ 0.11 & 0.55 $\pm$ 0.03 & 0.45 $\pm$ 0.12 \\
& SEAL-zero  & 0.64 $\pm$ 0.03 & 0.44 $\pm$ 0.15 & 0.66 $\pm$ 0.06 & 0.41 $\pm$ 0.18 & 0.67 $\pm$ 0.10 & 0.38 $\pm$ 0.20 \\
        \midrule
    \parbox[t]{2mm}{\multirow{4}{*}{\rotatebox[origin=c]{90}{$\lambda = 0.5$}}} 

& SEAL-mask-abs  & 0.37 $\pm$ 0.01 & 0.18 $\pm$ 0.01 & 0.37 $\pm$ 0.01 & 0.17 $\pm$ 0.00 & 0.39 $\pm$ 0.02 & 0.15 $\pm$ 0.01 \\
& SEAL-mask  & 0.57 $\pm$ 0.01 & 0.00 $\pm$ 0.00 & 0.66 $\pm$ 0.02 & 0.00 $\pm$ 0.00 & 0.76 $\pm$ 0.01 & 0.00 $\pm$ 0.00 \\
& SEAL-abs  & 0.52 $\pm$ 0.04 & 0.49 $\pm$ 0.05 & 0.52 $\pm$ 0.04 & 0.48 $\pm$ 0.06 & 0.52 $\pm$ 0.04 & 0.47 $\pm$ 0.07 \\
& SEAL-zero  & 0.62 $\pm$ 0.03 & 0.47 $\pm$ 0.08 & 0.62 $\pm$ 0.07 & 0.45 $\pm$ 0.10 & 0.61 $\pm$ 0.09 & 0.42 $\pm$ 0.11 \\
        \midrule
    \parbox[t]{2mm}{\multirow{4}{*}{\rotatebox[origin=c]{90}{$\lambda = 10^{-1}$}}} 

& SEAL-mask-abs  & 0.37 $\pm$ 0.01 & 0.20 $\pm$ 0.02 & 0.38 $\pm$ 0.01 & 0.19 $\pm$ 0.01 & 0.40 $\pm$ 0.01 & 0.16 $\pm$ 0.01 \\
& SEAL-mask  & 0.59 $\pm$ 0.01 & 0.00 $\pm$ 0.00 & 0.68 $\pm$ 0.01 & 0.00 $\pm$ 0.00 & 0.77 $\pm$ 0.01 & 0.00 $\pm$ 0.00 \\
& SEAL-abs  & 0.50 $\pm$ 0.03 & 0.50 $\pm$ 0.03 & 0.51 $\pm$ 0.02 & 0.49 $\pm$ 0.04 & 0.51 $\pm$ 0.02 & 0.48 $\pm$ 0.04 \\
& SEAL-zero  & 0.59 $\pm$ 0.05 & 0.49 $\pm$ 0.04 & 0.60 $\pm$ 0.07 & 0.48 $\pm$ 0.05 & 0.58 $\pm$ 0.07 & 0.45 $\pm$ 0.07 \\
        \midrule
    \parbox[t]{2mm}{\multirow{4}{*}{\rotatebox[origin=c]{90}{$\lambda = 10^{-2}$}}} 

& SEAL-mask-abs  & 0.37 $\pm$ 0.01 & 0.19 $\pm$ 0.02 & 0.38 $\pm$ 0.01 & 0.18 $\pm$ 0.01 & 0.39 $\pm$ 0.01 & 0.15 $\pm$ 0.01 \\
& SEAL-mask  & 0.59 $\pm$ 0.01 & 0.00 $\pm$ 0.00 & 0.68 $\pm$ 0.01 & 0.00 $\pm$ 0.00 & 0.78 $\pm$ 0.01 & 0.00 $\pm$ 0.00 \\
& SEAL-abs  & 0.47 $\pm$ 0.03 & 0.48 $\pm$ 0.03 & 0.48 $\pm$ 0.02 & 0.47 $\pm$ 0.03 & 0.49 $\pm$ 0.02 & 0.46 $\pm$ 0.04 \\
& SEAL-zero  & 0.59 $\pm$ 0.03 & 0.47 $\pm$ 0.05 & 0.58 $\pm$ 0.07 & 0.45 $\pm$ 0.07 & 0.56 $\pm$ 0.09 & 0.43 $\pm$ 0.09 \\
        \midrule
    \parbox[t]{2mm}{\multirow{4}{*}{\rotatebox[origin=c]{90}{$\lambda = 10^{-3}$}}} 

& SEAL-mask-abs  & 0.38 $\pm$ 0.03 & 0.24 $\pm$ 0.01 & 0.39 $\pm$ 0.02 & 0.22 $\pm$ 0.01 & 0.41 $\pm$ 0.02 & 0.20 $\pm$ 0.02 \\
& SEAL-mask  & 0.63 $\pm$ 0.03 & 0.02 $\pm$ 0.01 & 0.72 $\pm$ 0.03 & 0.01 $\pm$ 0.01 & 0.80 $\pm$ 0.03 & 0.01 $\pm$ 0.01 \\
& SEAL-abs  & 0.43 $\pm$ 0.04 & 0.48 $\pm$ 0.01 & 0.45 $\pm$ 0.04 & 0.48 $\pm$ 0.01 & 0.47 $\pm$ 0.03 & 0.47 $\pm$ 0.02 \\
& SEAL-zero  & 0.58 $\pm$ 0.04 & 0.48 $\pm$ 0.01 & 0.59 $\pm$ 0.05 & 0.47 $\pm$ 0.02 & 0.57 $\pm$ 0.05 & 0.44 $\pm$ 0.04 \\
        \midrule
    \parbox[t]{2mm}{\multirow{4}{*}{\rotatebox[origin=c]{90}{$\lambda = 10^{-4}$}}} 

& SEAL-mask-abs  & 0.42 $\pm$ 0.02 & 0.28 $\pm$ 0.04 & 0.44 $\pm$ 0.02 & 0.28 $\pm$ 0.04 & 0.46 $\pm$ 0.02 & 0.27 $\pm$ 0.03 \\
& SEAL-mask  & 0.63 $\pm$ 0.01 & 0.09 $\pm$ 0.03 & 0.71 $\pm$ 0.01 & 0.07 $\pm$ 0.02 & 0.78 $\pm$ 0.01 & 0.05 $\pm$ 0.02 \\
& SEAL-abs  & 0.46 $\pm$ 0.02 & 0.49 $\pm$ 0.01 & 0.49 $\pm$ 0.01 & 0.49 $\pm$ 0.01 & 0.50 $\pm$ 0.02 & 0.49 $\pm$ 0.01 \\
& SEAL-zero  & 0.56 $\pm$ 0.05 & 0.49 $\pm$ 0.01 & 0.57 $\pm$ 0.05 & 0.48 $\pm$ 0.01 & 0.55 $\pm$ 0.06 & 0.47 $\pm$ 0.02 \\
        \midrule
    \parbox[t]{2mm}{\multirow{4}{*}{\rotatebox[origin=c]{90}{$\lambda = 0$}}} 

& SEAL-mask-abs  & 0.46 $\pm$ 0.02 & 0.32 $\pm$ 0.02 & 0.47 $\pm$ 0.01 & 0.32 $\pm$ 0.02 & 0.48 $\pm$ 0.01 & 0.31 $\pm$ 0.02 \\
& SEAL-mask  & 0.67 $\pm$ 0.02 & 0.15 $\pm$ 0.02 & 0.74 $\pm$ 0.02 & 0.15 $\pm$ 0.01 & 0.78 $\pm$ 0.03 & 0.14 $\pm$ 0.01 \\
& SEAL-abs  & 0.43 $\pm$ 0.02 & 0.49 $\pm$ 0.02 & 0.48 $\pm$ 0.02 & 0.49 $\pm$ 0.02 & 0.49 $\pm$ 0.02 & 0.48 $\pm$ 0.02 \\
& SEAL-zero  & 0.52 $\pm$ 0.02 & 0.49 $\pm$ 0.02 & 0.53 $\pm$ 0.02 & 0.48 $\pm$ 0.02 & 0.51 $\pm$ 0.02 & 0.48 $\pm$ 0.02 \\
    \bottomrule
    \end{tabular}
\end{table*}

\begin{table*}
    \centering
    \small
    \setlength{\tabcolsep}{4.8pt} 
        \caption{Model explanations performance using different type of masking strategy in SEAL architecture for CYP2C9 dataset. Evaluating using Fidelity metrics at 10\%, 20\%, and 30\% masking thresholds.}
    \label{tab:masking_cyp}

    \begin{tabular}{ c l c  c c c c c
    }
        \toprule
        & Model & $\text{Fidelity}_{10}+ \uparrow$ & $\text{Fidelity}_{10}- \downarrow$ & $\text{Fidelity}_{20}+ \uparrow$ & $\text{Fidelity}_{20}- \downarrow$ & $\text{Fidelity}_{30}+ \uparrow$ & $\text{Fidelity}_{30}- \downarrow$ \\
        \midrule
        \multicolumn{8}{c}{CYP2C9} \\
        \midrule
    \parbox[t]{2mm}{\multirow{4}{*}{\rotatebox[origin=c]{90}{$\lambda = 2$}}} 

        & SEAL-mask-abs  & 0.36 $\pm$ 0.01 & 0.19 $\pm$ 0.01 & 0.37 $\pm$ 0.01 & 0.18 $\pm$ 0.01 & 0.39 $\pm$ 0.02 & 0.15 $\pm$ 0.02 \\
        & SEAL-mask  & 0.52 $\pm$ 0.02 & 0.04 $\pm$ 0.05 & 0.57 $\pm$ 0.03 & 0.03 $\pm$ 0.04 & 0.66 $\pm$ 0.03 & 0.01 $\pm$ 0.01 \\
        & SEAL-abs  & 0.41 $\pm$ 0.11 & 0.40 $\pm$ 0.11 & 0.42 $\pm$ 0.11 & 0.39 $\pm$ 0.11 & 0.45 $\pm$ 0.10 & 0.38 $\pm$ 0.12 \\
        & SEAL-zero  & 0.49 $\pm$ 0.08 & 0.37 $\pm$ 0.15 & 0.51 $\pm$ 0.08 & 0.35 $\pm$ 0.15 & 0.54 $\pm$ 0.07 & 0.32 $\pm$ 0.16 \\

        \midrule
    \parbox[t]{2mm}{\multirow{4}{*}{\rotatebox[origin=c]{90}{$\lambda = 1$}}} 

& SEAL-mask-abs  & 0.34 $\pm$ 0.02 & 0.20 $\pm$ 0.02 & 0.35 $\pm$ 0.01 & 0.19 $\pm$ 0.02 & 0.36 $\pm$ 0.02 & 0.16 $\pm$ 0.02 \\
& SEAL-mask  & 0.50 $\pm$ 0.02 & 0.05 $\pm$ 0.06 & 0.55 $\pm$ 0.02 & 0.04 $\pm$ 0.05 & 0.64 $\pm$ 0.03 & 0.01 $\pm$ 0.01 \\
& SEAL-abs  & 0.40 $\pm$ 0.13 & 0.40 $\pm$ 0.12 & 0.41 $\pm$ 0.13 & 0.40 $\pm$ 0.12 & 0.42 $\pm$ 0.13 & 0.38 $\pm$ 0.12 \\
& SEAL-zero  & 0.45 $\pm$ 0.10 & 0.39 $\pm$ 0.13 & 0.46 $\pm$ 0.11 & 0.38 $\pm$ 0.13 & 0.47 $\pm$ 0.11 & 0.35 $\pm$ 0.14 \\
        \midrule
    \parbox[t]{2mm}{\multirow{4}{*}{\rotatebox[origin=c]{90}{$\lambda = 0.5$}}} 

& SEAL-mask-abs  & 0.35 $\pm$ 0.01 & 0.20 $\pm$ 0.02 & 0.36 $\pm$ 0.01 & 0.19 $\pm$ 0.02 & 0.37 $\pm$ 0.01 & 0.15 $\pm$ 0.01 \\
& SEAL-mask  & 0.50 $\pm$ 0.02 & 0.05 $\pm$ 0.04 & 0.56 $\pm$ 0.01 & 0.04 $\pm$ 0.03 & 0.64 $\pm$ 0.02 & 0.01 $\pm$ 0.01 \\
& SEAL-abs  & 0.35 $\pm$ 0.02 & 0.36 $\pm$ 0.04 & 0.37 $\pm$ 0.02 & 0.35 $\pm$ 0.04 & 0.40 $\pm$ 0.05 & 0.33 $\pm$ 0.05 \\
& SEAL-zero  & 0.44 $\pm$ 0.05 & 0.34 $\pm$ 0.05 & 0.46 $\pm$ 0.05 & 0.32 $\pm$ 0.05 & 0.48 $\pm$ 0.05 & 0.28 $\pm$ 0.05 \\
        \midrule
    \parbox[t]{2mm}{\multirow{4}{*}{\rotatebox[origin=c]{90}{$\lambda = 10^{-1}$}}} 

        & SEAL-mask-abs  & 0.38 $\pm$ 0.02 & 0.19 $\pm$ 0.02 & 0.38 $\pm$ 0.02 & 0.18 $\pm$ 0.02 & 0.38 $\pm$ 0.02 & 0.16 $\pm$ 0.01 \\
        & SEAL-mask  & 0.57 $\pm$ 0.02 & 0.00 $\pm$ 0.00 & 0.63 $\pm$ 0.02 & 0.00 $\pm$ 0.00 & 0.72 $\pm$ 0.02 & 0.00 $\pm$ 0.00 \\
        & SEAL-abs  & 0.53 $\pm$ 0.12 & 0.52 $\pm$ 0.10 & 0.55 $\pm$ 0.12 & 0.50 $\pm$ 0.10 & 0.57 $\pm$ 0.11 & 0.48 $\pm$ 0.10 \\
        & SEAL-zero  & 0.56 $\pm$ 0.08 & 0.51 $\pm$ 0.11 & 0.59 $\pm$ 0.08 & 0.49 $\pm$ 0.10 & 0.60 $\pm$ 0.08 & 0.45 $\pm$ 0.12 \\
        \midrule
    \parbox[t]{2mm}{\multirow{4}{*}{\rotatebox[origin=c]{90}{$\lambda = 10^{-2}$}}} 

        & SEAL-mask-abs  & 0.35 $\pm$ 0.02 & 0.20 $\pm$ 0.02 & 0.36 $\pm$ 0.02 & 0.20 $\pm$ 0.01 & 0.36 $\pm$ 0.02 & 0.18 $\pm$ 0.01 \\
        & SEAL-mask  & 0.53 $\pm$ 0.02 & 0.00 $\pm$ 0.00 & 0.60 $\pm$ 0.02 & 0.00 $\pm$ 0.00 & 0.69 $\pm$ 0.02 & 0.00 $\pm$ 0.00 \\
        & SEAL-abs  & 0.47 $\pm$ 0.14 & 0.40 $\pm$ 0.08 & 0.48 $\pm$ 0.15 & 0.40 $\pm$ 0.08 & 0.50 $\pm$ 0.15 & 0.38 $\pm$ 0.08 \\
        & SEAL-zero  & 0.50 $\pm$ 0.09 & 0.39 $\pm$ 0.12 & 0.52 $\pm$ 0.10 & 0.38 $\pm$ 0.12 & 0.53 $\pm$ 0.11 & 0.36 $\pm$ 0.14 \\
                \midrule

    \parbox[t]{2mm}{\multirow{4}{*}{\rotatebox[origin=c]{90}{$\lambda = 10^{-3}$}}} 

        & SEAL-mask-abs  & 0.38 $\pm$ 0.02 & 0.22 $\pm$ 0.03 & 0.39 $\pm$ 0.01 & 0.21 $\pm$ 0.02 & 0.39 $\pm$ 0.02 & 0.20 $\pm$ 0.02 \\
        & SEAL-mask  & 0.53 $\pm$ 0.02 & 0.06 $\pm$ 0.04 & 0.58 $\pm$ 0.02 & 0.05 $\pm$ 0.04 & 0.65 $\pm$ 0.02 & 0.04 $\pm$ 0.03 \\
        & SEAL-abs  & 0.41 $\pm$ 0.09 & 0.38 $\pm$ 0.09 & 0.42 $\pm$ 0.10 & 0.37 $\pm$ 0.08 & 0.43 $\pm$ 0.09 & 0.36 $\pm$ 0.08 \\
        & SEAL-zero  & 0.47 $\pm$ 0.06 & 0.36 $\pm$ 0.09 & 0.49 $\pm$ 0.08 & 0.35 $\pm$ 0.10 & 0.51 $\pm$ 0.09 & 0.31 $\pm$ 0.10 \\
        \midrule
    \parbox[t]{2mm}{\multirow{4}{*}{\rotatebox[origin=c]{90}{$\lambda = 10^{-4}$}}} 

        & SEAL-mask-abs  & 0.38 $\pm$ 0.03 & 0.26 $\pm$ 0.04 & 0.41 $\pm$ 0.04 & 0.25 $\pm$ 0.03 & 0.43 $\pm$ 0.04 & 0.24 $\pm$ 0.03 \\
        & SEAL-mask  & 0.52 $\pm$ 0.02 & 0.13 $\pm$ 0.05 & 0.58 $\pm$ 0.02 & 0.11 $\pm$ 0.03 & 0.63 $\pm$ 0.03 & 0.10 $\pm$ 0.04 \\
        & SEAL-abs  & 0.38 $\pm$ 0.07 & 0.41 $\pm$ 0.11 & 0.41 $\pm$ 0.09 & 0.39 $\pm$ 0.09 & 0.44 $\pm$ 0.10 & 0.36 $\pm$ 0.07 \\
        & SEAL-zero  & 0.41 $\pm$ 0.05 & 0.40 $\pm$ 0.13 & 0.45 $\pm$ 0.06 & 0.38 $\pm$ 0.12 & 0.47 $\pm$ 0.07 & 0.34 $\pm$ 0.11 \\
        \midrule
    \parbox[t]{2mm}{\multirow{4}{*}{\rotatebox[origin=c]{90}{$\lambda = 0$}}} 

        & SEAL-mask-abs  & 0.37 $\pm$ 0.01 & 0.25 $\pm$ 0.02 & 0.41 $\pm$ 0.02 & 0.23 $\pm$ 0.02 & 0.43 $\pm$ 0.03 & 0.22 $\pm$ 0.01 \\
        & SEAL-mask  & 0.52 $\pm$ 0.00 & 0.11 $\pm$ 0.02 & 0.59 $\pm$ 0.01 & 0.09 $\pm$ 0.01 & 0.65 $\pm$ 0.01 & 0.07 $\pm$ 0.01 \\
        & SEAL-abs  & 0.30 $\pm$ 0.01 & 0.32 $\pm$ 0.02 & 0.33 $\pm$ 0.01 & 0.31 $\pm$ 0.02 & 0.35 $\pm$ 0.02 & 0.29 $\pm$ 0.02 \\
        & SEAL-zero  & 0.39 $\pm$ 0.02 & 0.28 $\pm$ 0.02 & 0.43 $\pm$ 0.02 & 0.26 $\pm$ 0.02 & 0.47 $\pm$ 0.03 & 0.21 $\pm$ 0.02 \\

    \bottomrule
    \end{tabular}
\end{table*}

\begin{table*}
    \centering
    \small
    \setlength{\tabcolsep}{4.8pt} 
        \caption{Model explanations performance using different type of masking strategy in SEAL architecture for Solubility dataset. Evaluating using Fidelity metrics at 10\%, 20\%, and 30\% masking thresholds.}
    \label{tab:masking_sol}

    \begin{tabular}{ c l c  c c c c c
    }
        \toprule
        & Model & $\text{Fidelity}_{10}+ \uparrow$ & $\text{Fidelity}_{10}- \downarrow$ & $\text{Fidelity}_{20}+ \uparrow$ & $\text{Fidelity}_{20}- \downarrow$ & $\text{Fidelity}_{30}+ \uparrow$ & $\text{Fidelity}_{30}- \downarrow$ \\
        \midrule
        \multicolumn{8}{c}{Solubility} \\
        \midrule
    \parbox[t]{2mm}{\multirow{4}{*}{\rotatebox[origin=c]{90}{$\lambda = 2$}}} 
    & SEAL-mask-abs  & 0.45 $\pm$ 0.03 & 0.26 $\pm$ 0.10 & 0.46 $\pm$ 0.03 & 0.25 $\pm$ 0.10 & 0.49 $\pm$ 0.04 & 0.22 $\pm$ 0.09 \\
    & SEAL-mask  & 0.42 $\pm$ 0.03 & 0.30 $\pm$ 0.10 & 0.45 $\pm$ 0.03 & 0.29 $\pm$ 0.09 & 0.49 $\pm$ 0.03 & 0.27 $\pm$ 0.08 \\
    & SEAL-abs  & 1.12 $\pm$ 0.25 & 1.04 $\pm$ 0.42 & 1.20 $\pm$ 0.28 & 0.96 $\pm$ 0.38 & 1.34 $\pm$ 0.32 & 0.84 $\pm$ 0.33 \\
    & SEAL-zero  & 0.98 $\pm$ 0.23 & 1.20 $\pm$ 0.47 & 1.07 $\pm$ 0.26 & 1.11 $\pm$ 0.42 & 1.21 $\pm$ 0.30 & 0.98 $\pm$ 0.37 \\
            \midrule

    \parbox[t]{2mm}{\multirow{4}{*}{\rotatebox[origin=c]{90}{$\lambda = 1$}}} 

    & SEAL-mask-abs  & 0.44 $\pm$ 0.04 & 0.22 $\pm$ 0.04 & 0.46 $\pm$ 0.04 & 0.21 $\pm$ 0.04 & 0.48 $\pm$ 0.05 & 0.19 $\pm$ 0.04 \\
    & SEAL-mask  & 0.41 $\pm$ 0.03 & 0.27 $\pm$ 0.04 & 0.44 $\pm$ 0.03 & 0.27 $\pm$ 0.04 & 0.47 $\pm$ 0.04 & 0.24 $\pm$ 0.04 \\
    & SEAL-abs  & 1.04 $\pm$ 0.31 & 0.83 $\pm$ 0.39 & 1.11 $\pm$ 0.34 & 0.78 $\pm$ 0.36 & 1.22 $\pm$ 0.38 & 0.69 $\pm$ 0.31 \\
    & SEAL-zero  & 0.90 $\pm$ 0.30 & 0.99 $\pm$ 0.42 & 0.98 $\pm$ 0.33 & 0.93 $\pm$ 0.38 & 1.10 $\pm$ 0.37 & 0.82 $\pm$ 0.33 \\
        \midrule
    \parbox[t]{2mm}{\multirow{4}{*}{\rotatebox[origin=c]{90}{$\lambda = 0.5$}}} 

    & SEAL-mask-abs  & 0.44 $\pm$ 0.02 & 0.22 $\pm$ 0.03 & 0.46 $\pm$ 0.02 & 0.21 $\pm$ 0.03 & 0.48 $\pm$ 0.02 & 0.19 $\pm$ 0.03 \\
    & SEAL-mask  & 0.41 $\pm$ 0.03 & 0.26 $\pm$ 0.03 & 0.44 $\pm$ 0.02 & 0.26 $\pm$ 0.03 & 0.48 $\pm$ 0.03 & 0.24 $\pm$ 0.03 \\
    & SEAL-abs  & 1.08 $\pm$ 0.31 & 0.89 $\pm$ 0.41 & 1.15 $\pm$ 0.33 & 0.83 $\pm$ 0.38 & 1.26 $\pm$ 0.37 & 0.74 $\pm$ 0.34 \\
    & SEAL-zero  & 0.91 $\pm$ 0.26 & 1.08 $\pm$ 0.50 & 1.01 $\pm$ 0.28 & 1.00 $\pm$ 0.46 & 1.13 $\pm$ 0.33 & 0.89 $\pm$ 0.40 \\
        \midrule
    \parbox[t]{2mm}{\multirow{4}{*}{\rotatebox[origin=c]{90}{$\lambda = 10^{-1}$}}} 

    & SEAL-mask-abs  & 0.44 $\pm$ 0.05 & 0.27 $\pm$ 0.03 & 0.45 $\pm$ 0.06 & 0.25 $\pm$ 0.03 & 0.47 $\pm$ 0.06 & 0.23 $\pm$ 0.02 \\
    & SEAL-mask  & 0.45 $\pm$ 0.04 & 0.24 $\pm$ 0.04 & 0.48 $\pm$ 0.04 & 0.24 $\pm$ 0.03 & 0.52 $\pm$ 0.04 & 0.23 $\pm$ 0.02 \\
    & SEAL-abs  & 0.77 $\pm$ 0.20 & 0.67 $\pm$ 0.29 & 0.83 $\pm$ 0.23 & 0.62 $\pm$ 0.26 & 0.92 $\pm$ 0.27 & 0.55 $\pm$ 0.23 \\
    & SEAL-zero  & 0.69 $\pm$ 0.20 & 0.78 $\pm$ 0.33 & 0.75 $\pm$ 0.23 & 0.73 $\pm$ 0.30 & 0.83 $\pm$ 0.25 & 0.65 $\pm$ 0.26 \\
        \midrule
    \parbox[t]{2mm}{\multirow{4}{*}{\rotatebox[origin=c]{90}{$\lambda = 10^{-2}$}}} 

    & SEAL-mask-abs  & 0.42 $\pm$ 0.02 & 0.23 $\pm$ 0.03 & 0.43 $\pm$ 0.02 & 0.22 $\pm$ 0.03 & 0.44 $\pm$ 0.02 & 0.20 $\pm$ 0.03 \\
    & SEAL-mask  & 0.42 $\pm$ 0.01 & 0.21 $\pm$ 0.02 & 0.45 $\pm$ 0.02 & 0.21 $\pm$ 0.02 & 0.48 $\pm$ 0.01 & 0.21 $\pm$ 0.02 \\
    & SEAL-abs  & 0.64 $\pm$ 0.12 & 0.50 $\pm$ 0.13 & 0.68 $\pm$ 0.13 & 0.49 $\pm$ 0.13 & 0.73 $\pm$ 0.14 & 0.46 $\pm$ 0.13 \\
    & SEAL-zero  & 0.59 $\pm$ 0.11 & 0.48 $\pm$ 0.11 & 0.64 $\pm$ 0.12 & 0.47 $\pm$ 0.11 & 0.72 $\pm$ 0.14 & 0.44 $\pm$ 0.11 \\
        \midrule
    \parbox[t]{2mm}{\multirow{4}{*}{\rotatebox[origin=c]{90}{$\lambda = 10^{-3}$}}} 

    & SEAL-mask-abs  & 0.48 $\pm$ 0.04 & 0.29 $\pm$ 0.03 & 0.51 $\pm$ 0.04 & 0.25 $\pm$ 0.03 & 0.55 $\pm$ 0.05 & 0.22 $\pm$ 0.03 \\
    & SEAL-mask  & 0.45 $\pm$ 0.04 & 0.33 $\pm$ 0.03 & 0.50 $\pm$ 0.05 & 0.31 $\pm$ 0.03 & 0.54 $\pm$ 0.05 & 0.27 $\pm$ 0.03 \\
    & SEAL-abs  & 1.24 $\pm$ 0.20 & 1.26 $\pm$ 0.33 & 1.37 $\pm$ 0.21 & 1.14 $\pm$ 0.30 & 1.55 $\pm$ 0.25 & 0.98 $\pm$ 0.25 \\
    & SEAL-zero  & 1.10 $\pm$ 0.23 & 1.43 $\pm$ 0.32 & 1.23 $\pm$ 0.25 & 1.29 $\pm$ 0.29 & 1.40 $\pm$ 0.27 & 1.11 $\pm$ 0.24 \\
        \midrule
    \parbox[t]{2mm}{\multirow{4}{*}{\rotatebox[origin=c]{90}{$\lambda = 10^{-4}$}}} 

    & SEAL-mask-abs  & 0.59 $\pm$ 0.06 & 0.41 $\pm$ 0.04 & 0.62 $\pm$ 0.06 & 0.39 $\pm$ 0.04 & 0.66 $\pm$ 0.06 & 0.36 $\pm$ 0.03 \\
    & SEAL-mask  & 0.56 $\pm$ 0.06 & 0.47 $\pm$ 0.04 & 0.60 $\pm$ 0.06 & 0.45 $\pm$ 0.04 & 0.64 $\pm$ 0.07 & 0.42 $\pm$ 0.03 \\
    & SEAL-abs  & 0.78 $\pm$ 0.14 & 0.64 $\pm$ 0.16 & 0.84 $\pm$ 0.15 & 0.58 $\pm$ 0.13 & 0.91 $\pm$ 0.17 & 0.52 $\pm$ 0.09 \\
    & SEAL-zero  & 0.72 $\pm$ 0.10 & 0.73 $\pm$ 0.22 & 0.79 $\pm$ 0.10 & 0.66 $\pm$ 0.19 & 0.88 $\pm$ 0.12 & 0.58 $\pm$ 0.15 \\
        \midrule
    \parbox[t]{2mm}{\multirow{4}{*}{\rotatebox[origin=c]{90}{$\lambda = 0$}}} 

    & SEAL-mask-abs  & 0.53 $\pm$ 0.06 & 0.46 $\pm$ 0.04 & 0.59 $\pm$ 0.06 & 0.42 $\pm$ 0.04 & 0.63 $\pm$ 0.07 & 0.37 $\pm$ 0.04 \\
    & SEAL-mask  & 0.54 $\pm$ 0.05 & 0.47 $\pm$ 0.04 & 0.61 $\pm$ 0.05 & 0.43 $\pm$ 0.04 & 0.65 $\pm$ 0.06 & 0.38 $\pm$ 0.04 \\
    & SEAL-abs  & 0.49 $\pm$ 0.03 & 0.55 $\pm$ 0.07 & 0.54 $\pm$ 0.04 & 0.53 $\pm$ 0.08 & 0.58 $\pm$ 0.05 & 0.48 $\pm$ 0.07 \\
    & SEAL-zero  & 0.53 $\pm$ 0.03 & 0.49 $\pm$ 0.07 & 0.60 $\pm$ 0.03 & 0.46 $\pm$ 0.06 & 0.65 $\pm$ 0.04 & 0.41 $\pm$ 0.05 \\
        \bottomrule
    \end{tabular}
\end{table*}


\section{E. Visualizations}
\begin{figure*}[t!]
    \centering
    \includegraphics[width=\linewidth]{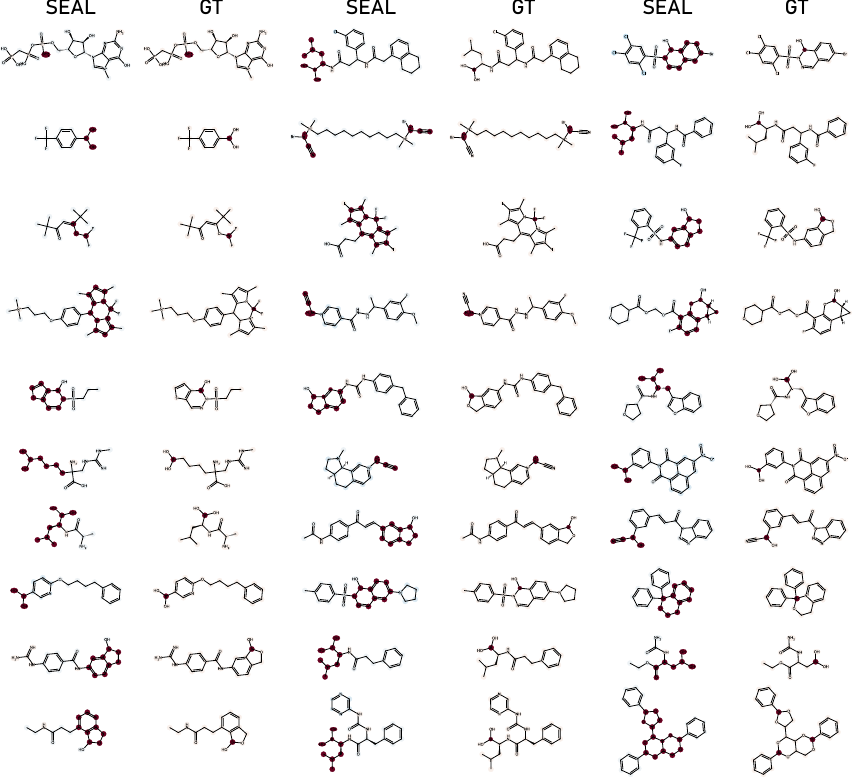}
    \caption{Node-level explanation examples of the SEAL method and Ground-Truth evaluated on the Boron (B) task for the positive target class. The red color indicates that the highlighted atoms had a positive contribution to the compound’s positive prediction. Blue as a negative contribution.}
    \label{fig:synthBpos}
\end{figure*}

\begin{figure*}[t!]
    \centering
    \includegraphics[width=\linewidth]{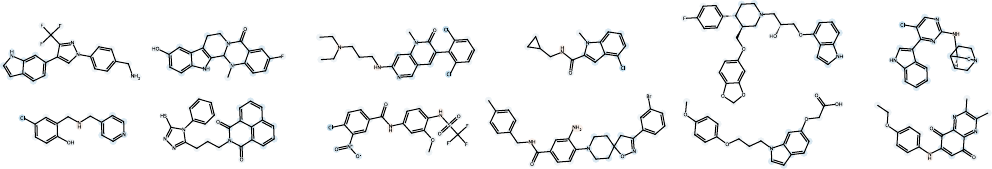}
    \caption{Node-level explanation examples of the SEAL method evaluated on the Boron (B) task for the negative target class. The red color indicates that the highlighted atoms had a positive contribution to the compound’s positive prediction. Blue as a negative contribution. }
    \label{fig:synthBneg}
\end{figure*}

\begin{figure*}[t!]
    \centering
    \includegraphics[width=\linewidth]{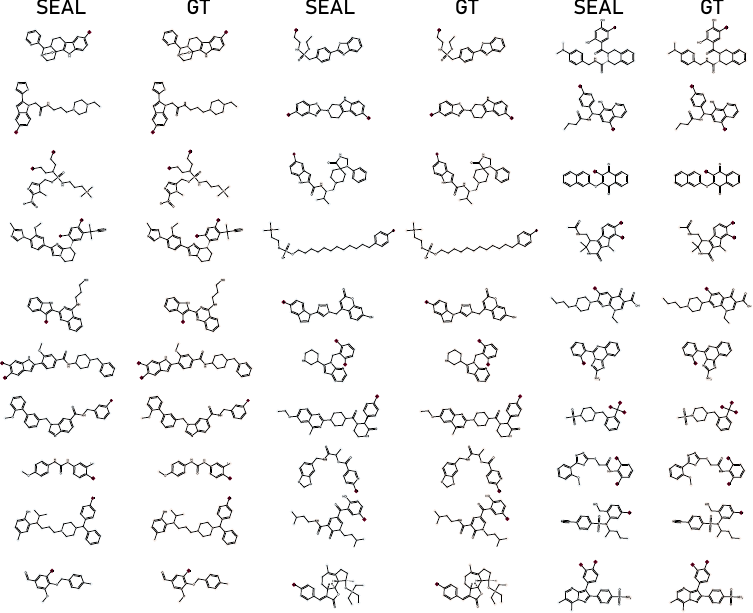}
    \caption{Node-level explanation examples of the SEAL method and Ground-Truth evaluated on Halogens (X) task for the positive target class. The red color indicates that the highlighted atoms had a positive contribution to the compound’s positive prediction. Blue as a negative contribution.}
    \label{fig:synthXpos}
\end{figure*}

\begin{figure*}[t!]
    \centering
    \includegraphics[width=\linewidth]{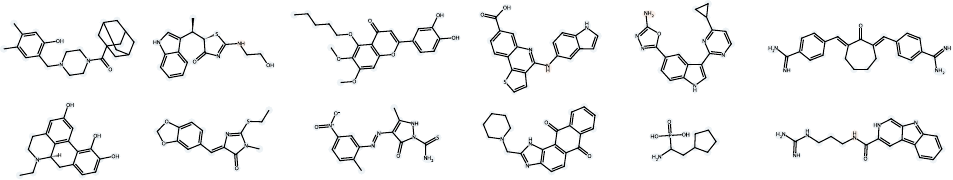}
    \caption{Node-level explanation examples of the SEAL method evaluated on the Halogens (X) task for the negative target class. The red color indicates that the highlighted atoms had a positive contribution to the compound’s positive prediction. Blue as a negative contribution. }
    \label{fig:synthXneg}
\end{figure*}

\begin{figure*}[t!]
    \centering
    \includegraphics[width=\linewidth]{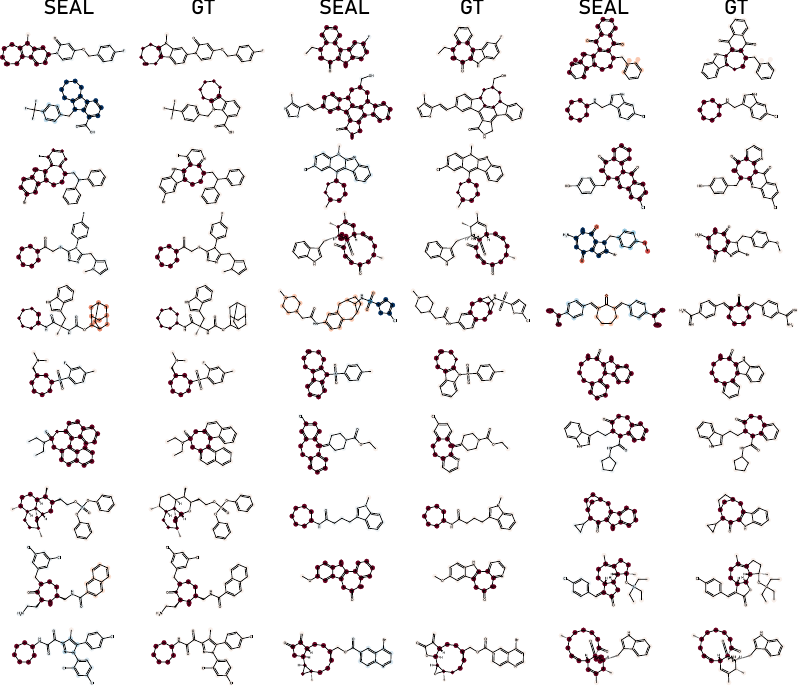}
    \caption{Node-level explanation examples of the SEAL method and Ground-Truth evaluated on the rings-max task for the positive target class. The red color indicates that the highlighted atoms had a positive contribution to the compound’s positive prediction. Blue as a negative contribution.}
    \label{fig:synthmaxpos}
\end{figure*}

\begin{figure*}[t!]
    \centering
    \includegraphics[width=\linewidth]{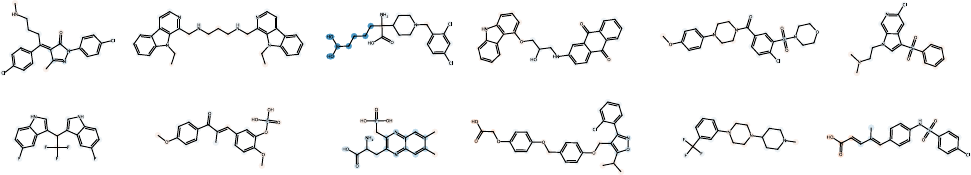}
    \caption{Node-level explanation examples of the SEAL method evaluated on the rings-max task for the negative target class. The red color indicates that the highlighted atoms had a positive contribution to the compound’s positive prediction. Blue as a negative contribution. }
    \label{fig:synthmaxneg}
\end{figure*}

\begin{figure*}[t!]
    \centering
    \includegraphics[width=\linewidth]{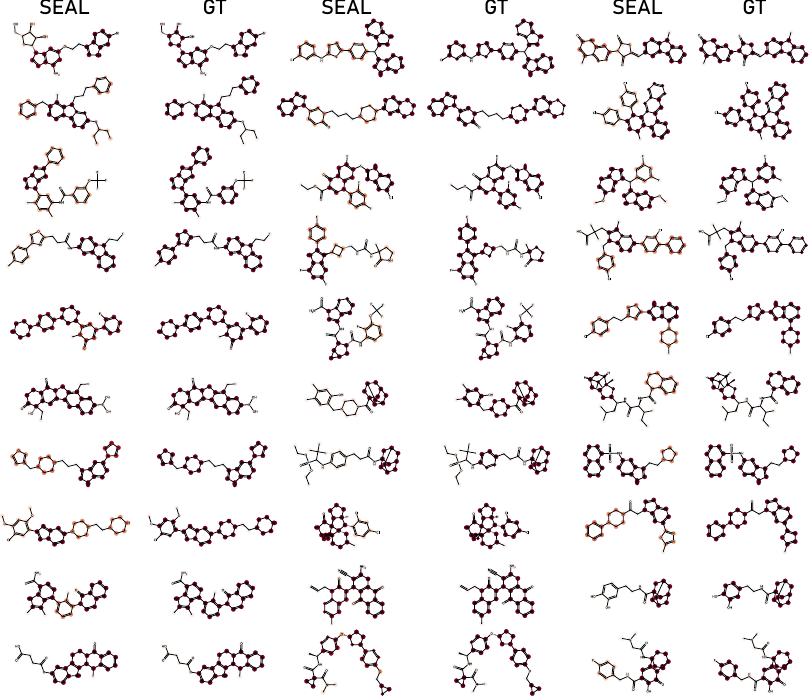}
    \caption{Node-level explanation examples of the SEAL method and Ground-Truth evaluated on the rings-count task for the positive target class. The red color indicates that the highlighted atoms had a positive contribution to the compound’s positive prediction. Blue as a negative contribution.}
    \label{fig:synthcountpos}
\end{figure*}

\begin{figure*}[t!]
    \centering
    \includegraphics[width=\linewidth]{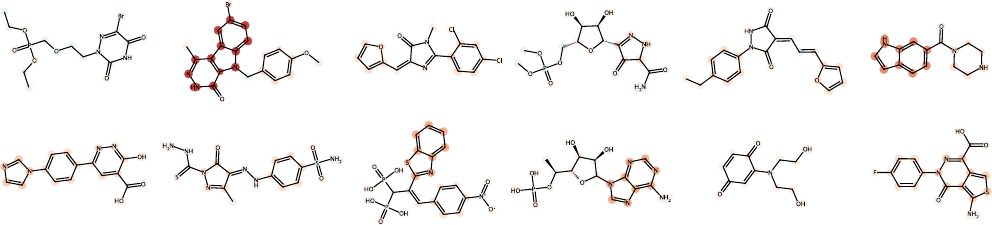}
    \caption{Node-level explanation examples of the SEAL method evaluated on the rings-count task for the negative target class. The red color indicates that the highlighted atoms had a positive contribution to the compound’s positive prediction. Blue as a negative contribution. }
    \label{fig:synthcountneg}
\end{figure*}

\begin{figure*}[t!]
    \centering
    \includegraphics[width=\linewidth]{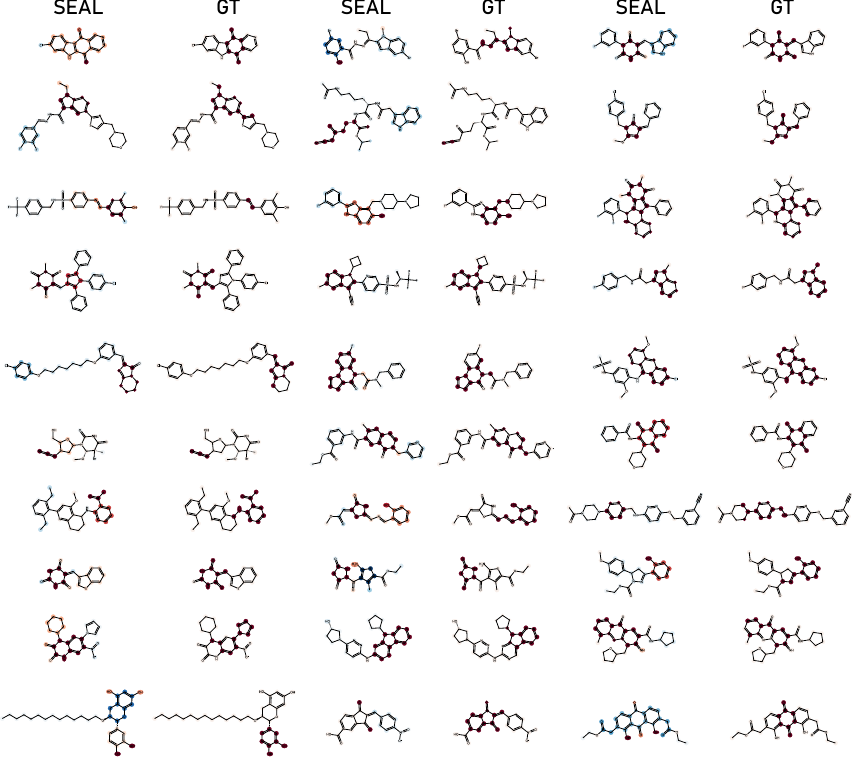}
    \caption{Node-level explanation examples of the SEAL method and Ground-Truth evaluated on the PAINS task for the positive target class. The red color indicates that the highlighted atoms had a positive contribution to the compound’s positive prediction. Blue as a negative contribution.}
    \label{fig:synthpainspos}
\end{figure*}

\begin{figure*}[t!]
    \centering
    \includegraphics[width=\linewidth]{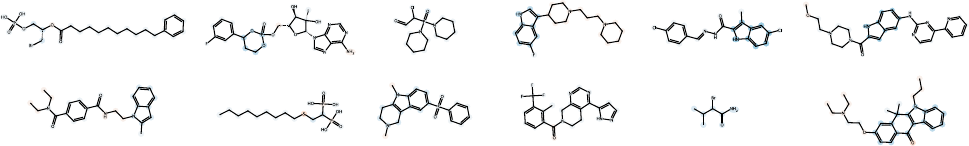}
    \caption{Node-level explanation examples of the SEAL method evaluated on the PAINS task for the negative target class. The red color indicates that the highlighted atoms had a positive contribution to the compound’s positive prediction. Blue as a negative contribution. }
    \label{fig:synthpainsneg}
\end{figure*}

\begin{figure*}[t!]
    \centering
    \includegraphics[width=\linewidth]{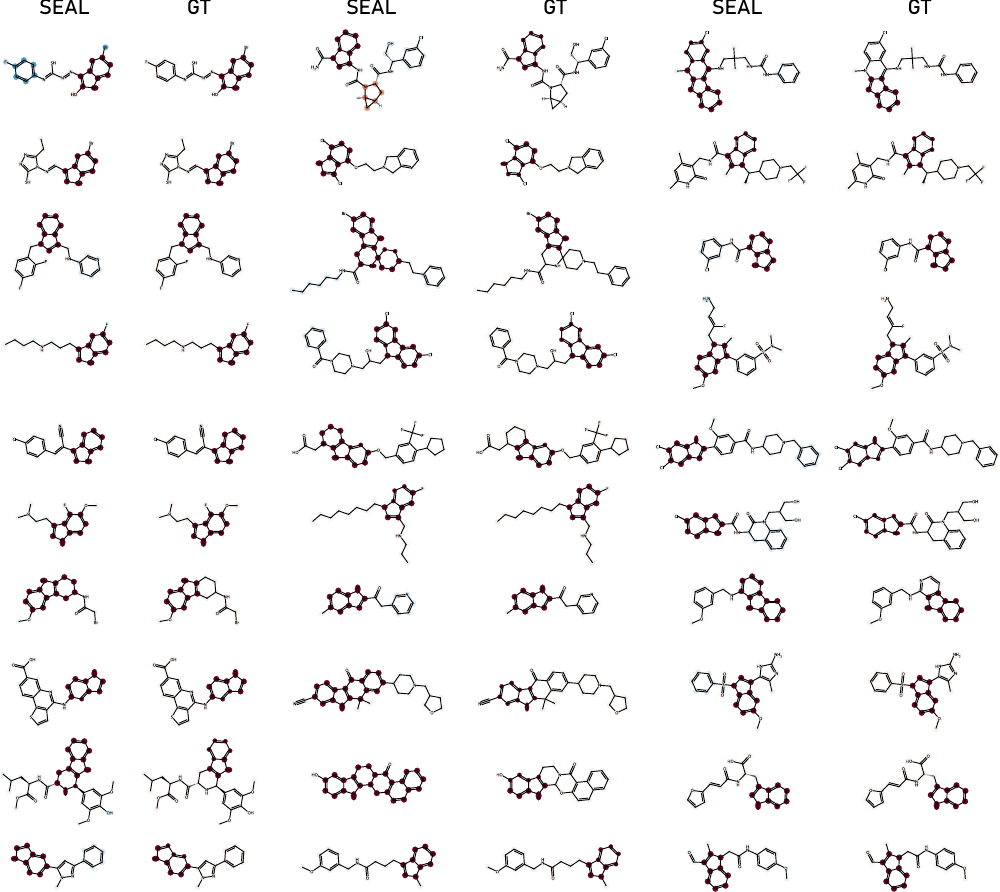}
    \caption{Node-level explanation examples of the SEAL method and Ground-Truth evaluated on the indole task for the positive target class. The red color indicates that the highlighted atoms had a positive contribution to the compound’s positive prediction. Blue as a negative contribution.}
    \label{fig:synthindolepos}
\end{figure*}

\begin{figure*}[t!]
    \centering
    \includegraphics[width=\linewidth]{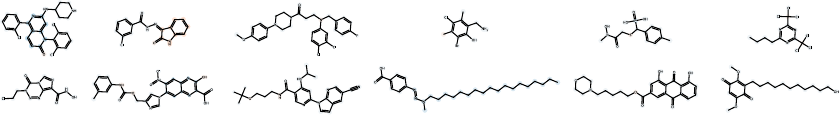}
    \caption{Node-level explanation examples of the SEAL method evaluated on the indole task for the negative target class. The red color indicates that the highlighted atoms had a positive contribution to the compound’s positive prediction. Blue as a negative contribution. }
    \label{fig:synthindoleneg}
\end{figure*}

\begin{figure*}[t!]
    \centering
    \includegraphics[width=\linewidth]{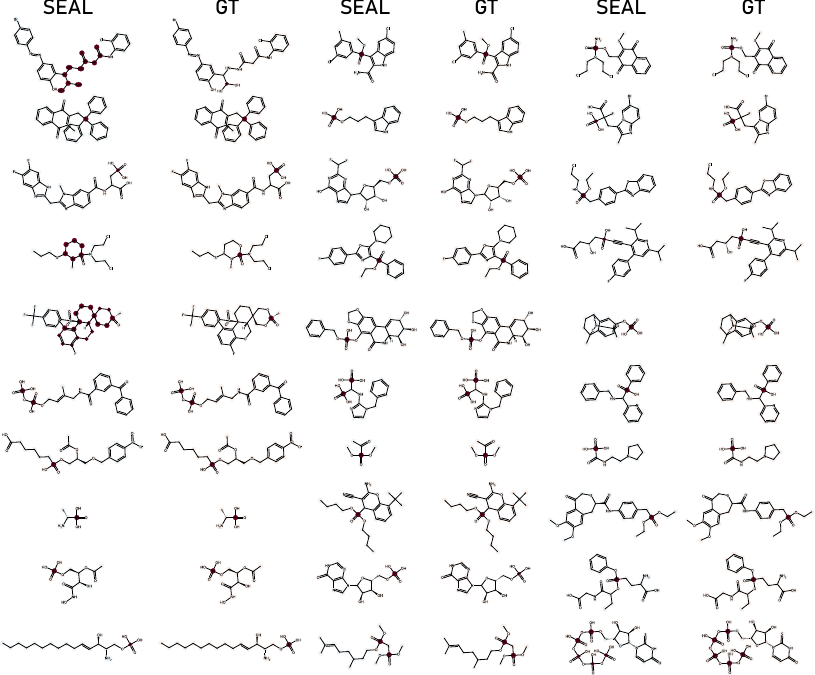}
    \caption{Node-level explanation examples of the SEAL method and Ground-Truth evaluated on the Phosphorus (P) task for the positive target class. The red color indicates that the highlighted atoms had a positive contribution to the compound’s positive prediction. Blue as a negative contribution.}
    \label{fig:synthPpos}
\end{figure*}

\begin{figure*}[t!]
    \centering
    \includegraphics[width=\linewidth]{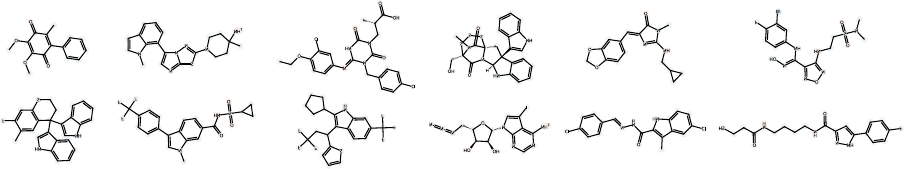}
    \caption{Node-level explanation examples of the SEAL method evaluated on the Phosphorus (P) task for the negative target class. The red color indicates that the highlighted atoms had a positive contribution to the compound’s positive prediction. Blue as a negative contribution. }
    \label{fig:synthPneg}
\end{figure*}

\begin{figure*}[t!]
    \centering
    \includegraphics[width=\linewidth]{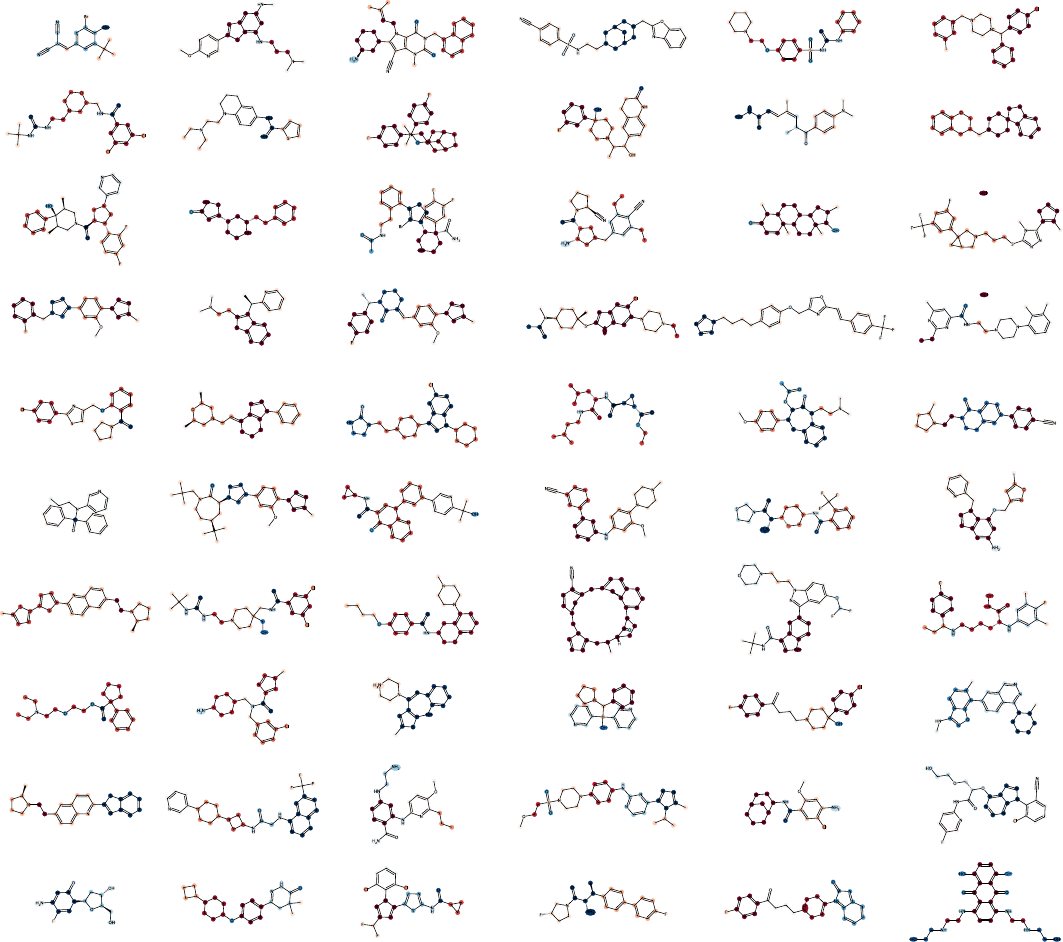}
    \caption{Node-level explanation examples of the SEAL method evaluated on the hERG dataset. The red color indicates that the highlighted atoms had a positive contribution to the positive prediction. Blue as a negative contribution. }
    \label{fig:realherg}
\end{figure*}

\begin{figure*}[t!]
    \centering
    \includegraphics[width=\linewidth]{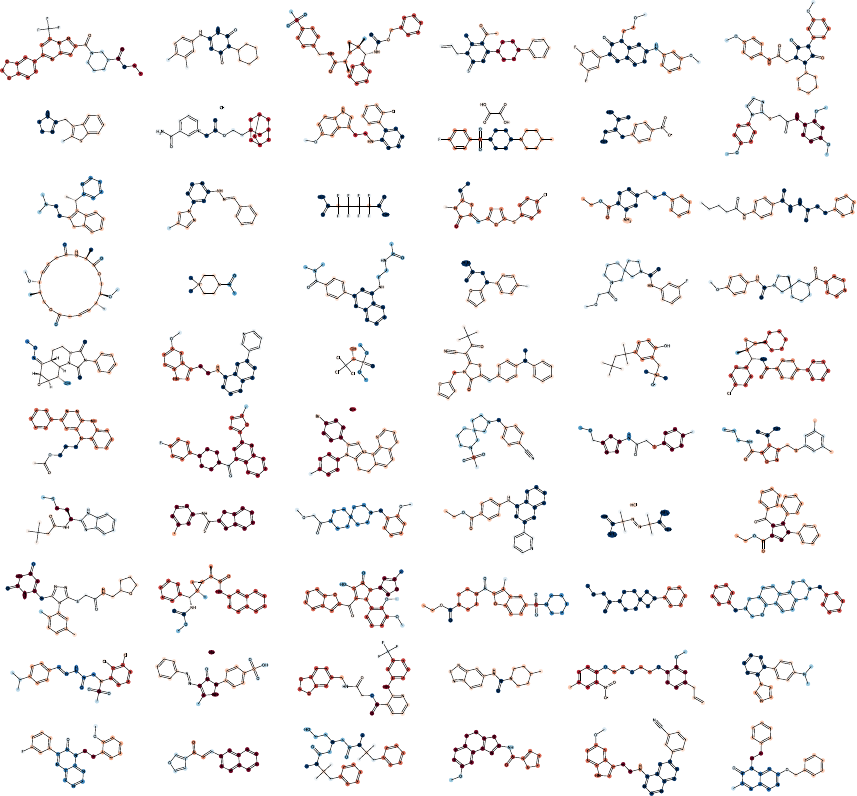}
    \caption{Node-level explanation examples of the SEAL method evaluated on the CYP2C9 dataset. The red color indicates that the highlighted atoms had a positive contribution to the positive prediction. Blue as a negative contribution. }
    \label{fig:realcyp}
\end{figure*}

\begin{figure*}[t!]
    \centering
    \includegraphics[width=\linewidth]{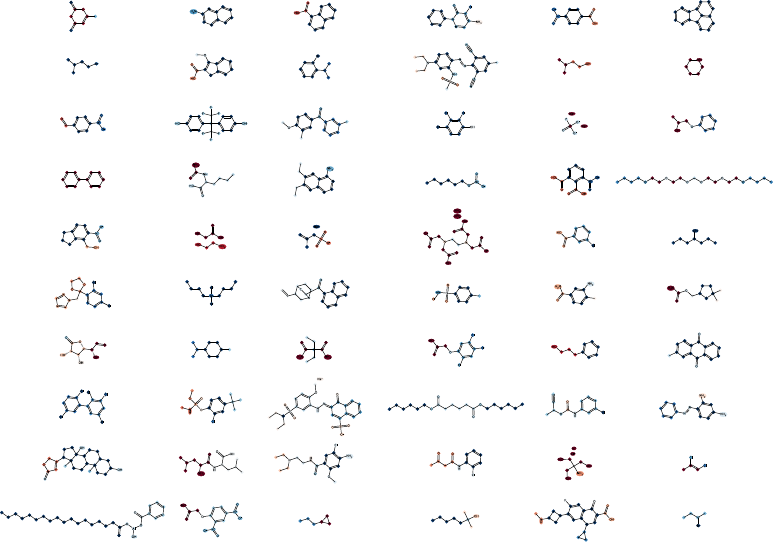}
    \caption{Node-level explanation examples of the SEAL method evaluated on the aqueous solubility dataset. The red color indicates that the highlighted atoms had a positive contribution to the positive prediction. Blue as a negative contribution. }
    \label{fig:realsol}
\end{figure*}

Figures \ref{fig:synthBpos}, \ref{fig:synthXpos}, \ref{fig:synthmaxpos}, \ref{fig:synthcountpos}, \ref{fig:synthpainspos}, \ref{fig:synthindolepos}, and \ref{fig:synthPpos} display examples of explanations generated by the SEAL model for the tasks in the synthetic dataset for the positive target class. The explanations for the negative class, where the substructure is not present in the compound, are illustrated in Figures \ref{fig:synthBneg}, \ref{fig:synthXneg}, \ref{fig:synthmaxneg}, \ref{fig:synthcountneg}, \ref{fig:synthpainsneg}, \ref{fig:synthindoleneg}, and \ref{fig:synthPneg}. The explanations for the real-world datasets are available in Figures \ref{fig:realherg}-\ref{fig:realsol}.


\end{document}